\documentclass[11pt,letterpaper]{iffyuan}

\usepackage[all]{hypcap}
\usepackage{algorithm}
\usepackage{algorithmicx}
\usepackage{algpseudocode}
\usepackage{float}
\usepackage{mathtools}
\usepackage{nicefrac}
\usepackage{subcaption}
\usepackage{wrapfig}
\usepackage{xspace}
\usepackage{tikz}
\usetikzlibrary{positioning, arrows.meta}
\usepackage{array}
\usepackage{makecell}
\usepackage{makecell}

\title{Embodied-R1.5: Evolving Physical Intelligence via Embodied Foundation Models}

\author[1]{Yifu Yuan\textsuperscript{\faGem,\faEnvelope \,}}
\author[1]{Yaoting Huang}
\author[1]{Xianze Yao}
\author[1]{Yutong Li}
\author[1]{Shuoheng Zhang}
\author[1]{Linqi Han}
\author[1]{Pengyi Li}
\author[1]{Jiangeng Sun}
\author[1]{Wenting Jia}
\author[1]{Zhao Zhang}
\author[1]{Yuhao Liu}
\author[1]{Ruihao Liao}
\author[1]{Yucheng Hu}
\author[1]{Qiyu Wu}
\author[1]{Yuxiao Li}
\author[1]{Zibin Dong}
\author[1]{Fei Ni}
\author[1]{Yan Zheng}
\author[2]{Shuyang Gu\textsuperscript{\faEnvelope \,}}
\author[1]{Yi Ma\textsuperscript{\faGem,\faEnvelope \,}}
\author[1]{Hongyao Tang\textsuperscript{\faGem,\faEnvelope \,}}
\author[2]{Han Hu}
\author[1]{Jianye Hao\textsuperscript{\faEnvelope \,}}

\affiliation[1]{Tianjin University}
\affiliation[2]{Tencent Hunyuan}

\contribution[\faGem]{Project Leader}
\contribution[\faEnvelope]{Corresponding Author (Contact: \href{mailto:yuanyf@tju.edu.cn}{yuanyf@tju.edu.cn})}

\setteamlogo{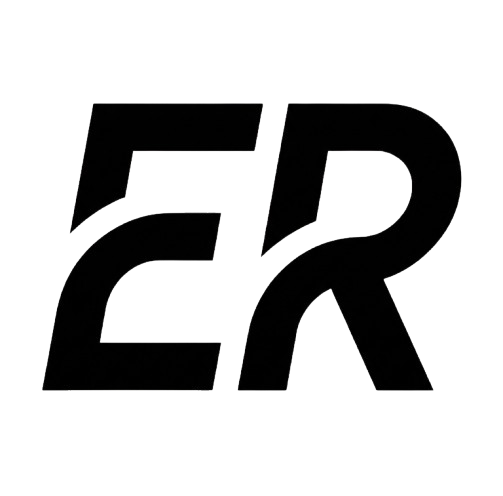}

\abstract{%
We introduce Embodied-R1.5, a unified Embodied Foundation Model (EFM) that integrates comprehensive embodied reasoning capabilities, spanning embodied cognition, task planning, correction, and pointing, within a single architecture toward general physical intelligence. Leveraging three automated data construction pipelines to significantly expand the data coverage of critical capabilities, we build a large-scale data system of over 15B tokens, and design a multi-task balanced RL recipe to alleviate heterogeneous task conflicts. We further introduce a Planner-Grounder-Corrector (PGC) closed-loop framework that enables a single model to autonomously execute and self-correct over long-horizon tasks. With only 8B parameters, Embodied-R1.5 achieves SOTA on 16 out of 24 embodied VLM benchmarks, surpassing leading models like Gemini-Robotics-ER-1.5 and GPT-5.4. Benefiting from the internalized embodied capabilities, Embodied-R1.5 can be fine-tuned into a VLA with only a small amount of data, outperforming leading VLA models like $\pi_{0.5}$ across 4 popular manipulation benchmark suites. We further conduct extensive zero-shot real-robot experiments, validating performance in instruction following, affordance grounding, articulated object manipulation, and long-horizon complex tasks, demonstrating strong generalization to the physical world. We open-source model weights, datasets, training code, and EmbodiedEvalKit, an evaluation framework tailored for embodied tasks, to facilitate future research in EFMs.%
}

\metadata[\faGlobe\ Project]{\href{https://embodied-r.github.io/}{https://embodied-r.github.io/}}
\metadata[\raisebox{-1.5pt}{\includegraphics[height=1.03em]{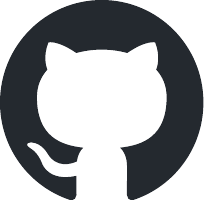}}\ Code]{\href{https://github.com/pickxiguapi/Embodied-R1.5}{https://github.com/pickxiguapi/Embodied-R1.5}}
\metadata[\raisebox{-1.5pt}{\includegraphics[height=1.03em]{figures/github-logo.pdf}}\ EmbodiedEvalKit]{\href{https://github.com/pickxiguapi/EmbodiedEvalKit}{https://github.com/pickxiguapi/EmbodiedEvalKit}}
\metadata[\raisebox{-1.5pt}{\includegraphics[height=0.96em]{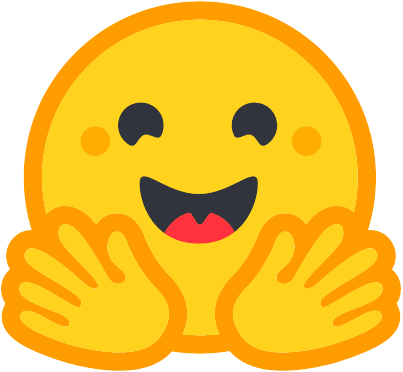}}\ Models \& Datasets]{\href{https://huggingface.co/collections/IffYuan/embodied-r15}{https://huggingface.co/collections/IffYuan/embodied-r15}}

\begin{document}

\vspace*{-1.5cm}

\maketitle

\begin{figure}[h]
\centering
\vspace{-5pt}
\includegraphics[width=1.01\linewidth]{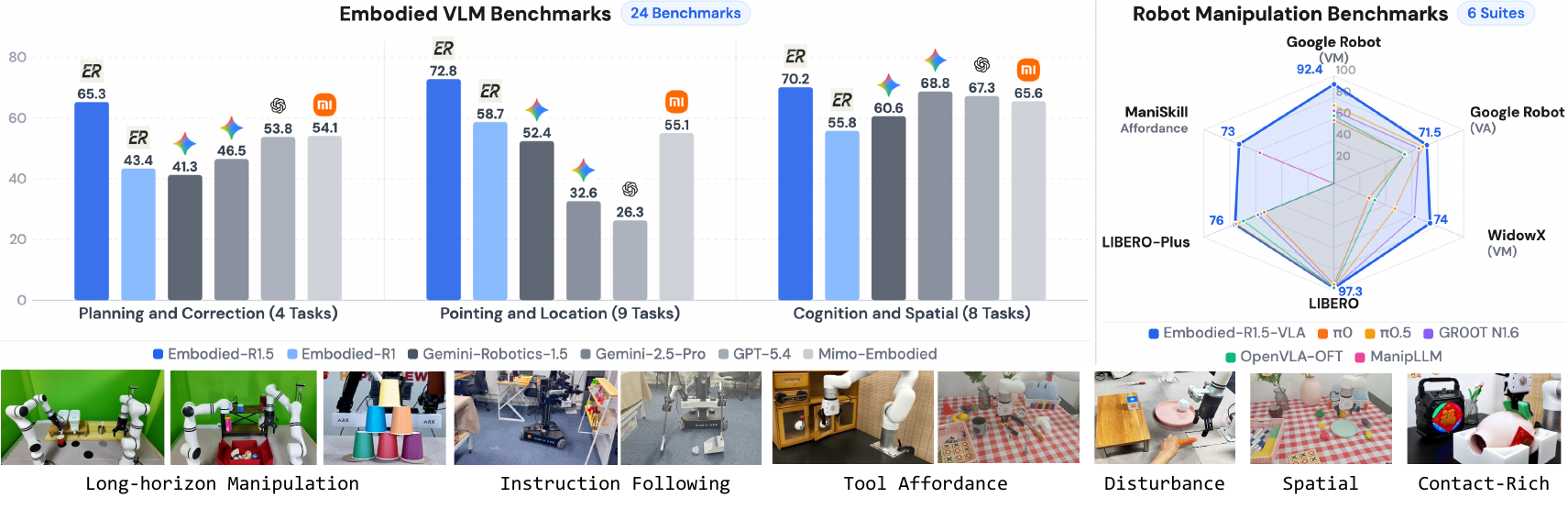}
\vspace{-20pt}
\caption{\textbf{Performance overview of Embodied-R1.5.} \textbf{Top:} Performance across 24 embodied VLM benchmarks (21 main benchmarks + 3 visual trace benchmarks) and 4 robotic manipulation benchmark suites, compared with leading general and embodied models. \textbf{Bottom:} Zero-shot real-robot experiments validating diverse embodied capabilities, including long-horizon manipulation, instruction following, tool affordance, and contact-rich tasks.}
\label{fig:teaser}
\end{figure}

\newpage
\tableofcontents
\newpage

\section{Introduction}

\begin{quote}
\textit{``Reasoning initiates the action; Action fulfills the reasoning.''}

\hfill --- Wang Yangming (1509), Pioneer of the Philosophy.
\end{quote}

Large language models have achieved remarkable success in the digital world, yet grounding intelligence in the physical world to realize general-purpose physical intelligence remains a central challenge~\citep{team2025gemini,yuan2025embodiedr,ni2025embodied}. Embodied reasoning is emerging as a critical pathway to bridge the seeing-to-doing gap~\citep{yuan2025seeing}: it requires models not only to perceive the physical world but also to reason about spatial geometry and task arrangement within it. To this end, we argue that achieving general physical intelligence calls for an Embodied Foundation Model (EFM)~\citep{team2025gemini}. An EFM requires all capabilities to be grounded in physics, unifying perception, reasoning and execution within a single architecture. We organize the capabilities required by an EFM into three core dimensions: \textit{spatial cognition and reasoning}, which endows the model with the ability to comprehend the semantic and spatial structure of the physical world; \textit{task planning and correction}, which organizes execution logic while monitoring progress and correcting errors; and \textit{embodied pointing and location}, which grounds high-level reasoning in coordinates and trajectories. Together, they enable a single model to perceive, act, and self-correct in a unified loop.

However, existing work faces three fundamental bottlenecks in realizing such a unified EFM. 1) \textit{Fragmented capabilities.} Current embodied models each cover only part of the spectrum: some focus on cognition~\citep{sermanet2024robovqa}, some on planning and correction~\citep{ji2025robobrain,team2025robobrain,pan2026thinker}, and others on grounding~\citep{yuan2024robopoint,yuan2025seeing,yuan2025embodiedr}; efforts like RynnBrain~\citep{dang2026rynnbrain} attempt to span multiple dimensions but rely on separate models of different scales for different tasks rather than truly unifying them within a single model. 2) \textit{Multi-task conflict}~\citep{feng2025onethinker}. Long text reasoning, trajectory prediction, and diverse grounding tasks differ drastically in output format; multi-task joint learning suffers from severe convergence difficulties, where different capabilities erode one another. 3) \textit{Lack of closed-loop autonomy validation on long-horizon tasks.} Most existing EFMs remain at the level of Embodied QA, without verifying whether reasoning capabilities are truly grounded in physics under long-horizon complex decision-making.

Building on the paradigm of our prior work Embodied-R1~\citep{yuan2025embodiedr}, Embodied-R1.5 leaps from a pointing specialist to a comprehensive EFM that unifies all three capability dimensions, addressing the three bottlenecks above with a systematic solution. Specifically, Embodied-R1.5 jointly internalizes all three capability dimensions within a single model, breaking free from the fragmented landscape. To support this capability unification, we employ three automated data production pipelines to construct a large-scale data corpus of over 15B tokens, and design a two-stage training paradigm together with a multi-task balanced RL recipe that resolves the interference inherent to heterogeneous joint training. Building on this, we further propose the Planner-Grounder-Corrector~(PGC) closed-loop framework, where a single model simultaneously drives the full autonomy stack, enabling long-horizon real-world tasks (e.g., making milk tea, sweeping garbage, stacking cups) to be fully completed without human intervention, with a single 8B model serving as planner, grounder, and corrector simultaneously.

Ultimately, with only 8B parameters, Embodied-R1.5 achieves SOTA on 16 out of 24 Embodied VLM benchmarks, with an average score of 70.4\% across the 21 main accuracy-based benchmarks, surpassing the embodied model Gemini-Robotics-ER-1.5~\citep{team2025gemini} and the general-purpose model GPT-5.4 by 17.0\% and 21.7\%, respectively. More critically, because the EFM has already internalized comprehensive embodied reasoning capabilities upstream, the downstream action head only needs to learn a simple mapping from understood intent to continuous actions. Consequently, Embodied-R1.5 requires no large-scale action pretraining: with only a small amount of action data, it can be adapted into Embodied-R1.5-VLA, which comprehensively outperforms strong VLA baselines that rely on large-scale action pretraining across 4 popular robotic manipulation benchmark suites: 92.4\% on SimplerEnv Google Robot Visual Matching (surpassing $\pi_{0.5}$~\citep{intelligence2025visionlanguageaction} by over 20\%) and substantially outperforming the specialized method ManipLLM~\citep{li2024manipllm} by 11\% on PartNet-Mobility, demonstrating that the internalization of embodied reasoning can effectively substitute for action data scaling. Zero-shot real-robot experiments further cover instruction following, affordance grounding, articulated object manipulation, and long-horizon complex tasks, demonstrating strong generalization to diverse real-world scenarios.

Our contributions are as follows:
(1) \textbf{Unified embodied capability system with closed-loop autonomy.} We unify all capability dimensions within a single model, and propose the PGC closed-loop execution framework that enables autonomous planning, execution, and self-correction on long-horizon complex tasks without human intervention.
(2) \textbf{A complete EFM recipe.} We provide an end-to-end recipe covering data mixture, training, and EmbodiedEvalKit, a comprehensive unified EFM evaluation framework supporting 25+ embodied benchmarks.
(3) \textbf{Comprehensive SOTA performance.} With only 8B parameters, we achieve SOTA on 16 out of 24 Embodied VLM benchmarks, surpassing Gemini-Robotics-ER-1.5; light action-data fine-tuning outperforms $\pi_{0.5}$ and other strong VLA baselines across 4 robotic manipulation benchmark suites; zero-shot real-robot experiments cover diverse scenarios and demonstrate strong generalization.
(4) \textbf{Open-source ecosystem.} We open-source all model weights, training data, and training code, providing a fully reproducible infrastructure for community research on EFMs.

\section{Unified Embodied Capabilities \& Architecture}
\label{sec:capabilities}

\begin{figure}[t]
\centering
\includegraphics[width=\linewidth]{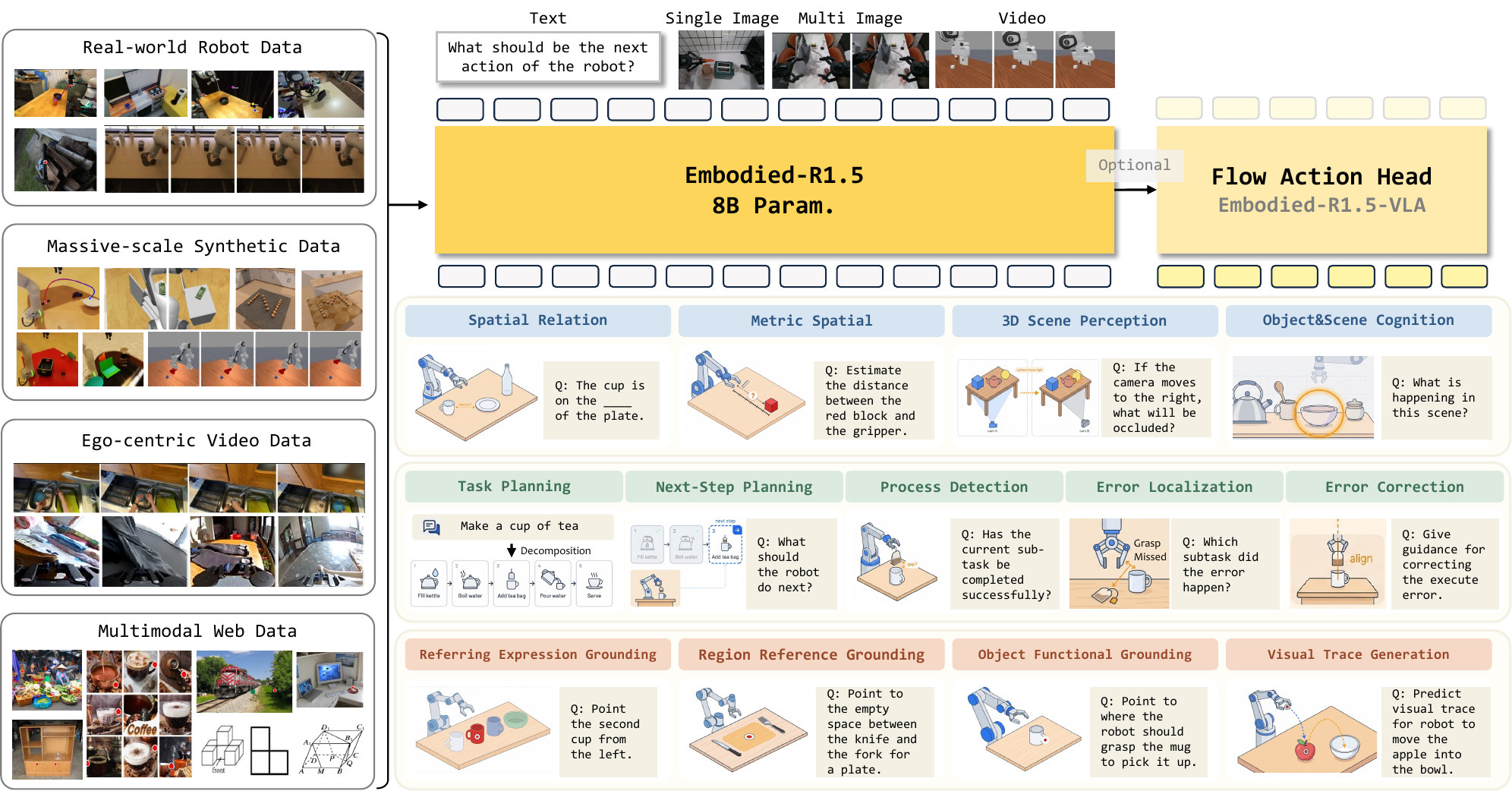}
\caption{\textbf{Capability taxonomy and architecture of Embodied-R1.5.} \textbf{Top:} Embodied-R1.5 is an EFM that can be further extended into a VLA by attaching a lightweight flow-matching action expert. \textbf{Bottom:} The three unified embodied capability dimensions, color-coded as \textcolor[HTML]{4A9B8E}{Cognition \& Spatial Reasoning}, \textcolor[HTML]{3B6FA0}{Planning \& Correction}, and \textcolor[HTML]{D4853B}{Pointing \& Location}, each supported by dedicated automated data pipelines totaling over 15B tokens.}
\label{fig:architecture}
\end{figure}

As illustrated in Figure~\ref{fig:architecture}, we organize the capabilities required by an EFM into three complementary dimensions that form a progressive reasoning chain from perception to decision to execution. Unifying them within a single model allows information to flow freely across dimensions without external communication. We first formalize these three dimensions, then describe the model architecture.

\subsection{Unified Embodied Capabilities}
\label{sec:cap:unified}

These three capabilities are not an arbitrary collection of independent tasks, but rather complementary expressions of embodied reasoning at different levels of abstraction: cognition and spatial reasoning endow the model with the foundational ability to comprehend the semantic and spatial structure of the physical world; planning and correction handle high-level task scheduling while monitoring progress and errors in closed loop during execution; and pointing and location ground high-level reasoning results into point and trajectory level actionable interaction information.

\textbf{Embodied Cognition \& Spatial Reasoning.} This dimension requires the model to see and understand the structure of the physical world, encompassing both static geometric relations and dynamic interaction possibilities. We decompose this capability into four sub-dimensions.
(1) \textit{Spatial Relation Understanding} Recognizing relative orientations between objects, support relations, containment, as well as occlusion and visibility from the current viewpoint. This is the foundation for downstream planning and localization.
(2) \textit{Metric Spatial Reasoning} Reasoning about metric information in the physical world, including estimating distances between objects, judging object sizes, and determining whether a free space can accommodate a target object. This provides quantitative grounding for precise manipulation.
(3) \textit{3D Scene Perception} Performing monocular depth estimation, cross-view geometric reasoning, and maintaining spatial consistency across video frames, providing geometric priors for collision avoidance and precise manipulation.
(4) \textit{Object \& Scene Cognition} Understanding scenes from a robot-centric or first-person perspective, distinguishing manipulation targets from obstacles and inferring the task context.

\textbf{Embodied Planning \& Correction.} This dimension covers the entire life cycle of task execution, from pre-execution planning through runtime monitoring to post-failure recovery. In terms of \textit{Task Planning}: (1) \textit{Long-horizon Task Decomposition} takes the overall language instruction and current scene observation to output a structured sub-task sequence; (2) \textit{Next-step Planning} rolls out the next atomic action instruction during execution, conditioned on current progress and observation, directly serving closed-loop control. Regarding \textit{Process Monitoring \& Correction}: (3) \textit{Process Detection} judges whether the current sub-task has been completed, determining when to advance to the next sub-task; (4) \textit{Error Localization} identifies the specific cause of failure from both execution and planning perspectives once a failure is detected; (5) \textit{Error Correction} generates concrete corrective suggestions (replanning sub-tasks or rolling back to retry), which feed directly as input context for the next round of planning.

\label{sec:cap:pointing}
\textbf{Embodied Pointing \& Location.} Pointing is the signature capability of the Embodied-R series~\citep{yuan2025embodiedr} and the critical interface connecting embodied reasoning to physical execution. Compared to Embodied-R1, Embodied-R1.5 substantially strengthens open-vocabulary pointing, with particularly significant advances in physical affordance understanding. Beyond this, Embodied-R1.5 treats coordinates as symbols that can be repeatedly referenced and reasoned about by the model, internalizing pointing as an element of embodied reasoning and maintaining visual anchoring throughout multi-step reasoning. The pointing capability is organized into four types:
(1) \textit{Referring Expression Grounding (REG).} Given a natural-language expression, the model outputs the corresponding object's point coordinate.
(2) \textit{Region Referring Grounding (RRG).} RRG handles regions rather than objects, for example identifying \textit{``an empty region that can hold a plate''}.
(3) \textit{Object Functional Grounding (OFG).} Beyond object recognition, OFG further localizes an object's functional parts (spout, handle, button, etc.).
(4) \textit{Visual Trace Generation (VTG).} VTG generates ordered point sequences that describe complete manipulation trajectories. Embodied-R1.5 extends this to simultaneously support \textit{object flow} and \textit{end-effector flow}: the former describes the expected motion path of the target object from its starting position to the goal position, and the latter describes the expected motion path of the robot end-effector. Beyond 2D traces, Embodied-R1.5 additionally supports 3D visual trace generation, producing spatially grounded trajectories in three-dimensional space.

\subsection{Architecture}
\label{sec:cap:architecture}

As shown in Figure~\ref{fig:architecture}, Embodied-R1.5 is an 8B-parameter VLM whose scale balances strong reasoning capability with practical deployment cost. All outputs are expressed as plain-text token sequences, with coordinates normalized to $[0,1000]$, trajectories as ordered coordinate sequences, and reasoning as free-form text, enabling the model to freely interleave reasoning chains and planning steps. Compared to approaches that rely on additional special tokens for coordinate output, directly generating numeric coordinates as plain text yields more stable predictions while incurring negligible vocabulary overhead. Coordinates persist as referenceable symbols across reasoning steps, maintaining visual anchoring throughout multi-step inference. Embodied-R1.5 can be further extended into a VLA that directly outputs continuous actions, termed \textbf{Embodied-R1.5-VLA}. Since the VLM has already internalized rich embodied reasoning capabilities, the mapping from understood intent to actions is comparatively simple; we therefore attach a lightweight flow-matching action expert based on DiT~\citep{peebles2023scalable,lipman2023flow} to the VLM backbone, forming a dual-system architecture (System~2 for reasoning, System~1 for action generation)~\citep{bjorck2025gr00t}. The action expert extracts vision-language features from intermediate VLM layers and generates continuous action sequences via action chunking~\citep{zhao2023learning}. Our hypothesis is that when the VLM has sufficiently internalized embodied reasoning capabilities, large-scale action pretraining becomes less critical; a stronger embodied backbone should substantially reduce the data requirement for downstream action learning~\citep{zhang2026vlmvla, li2026robointer}.

\section{Training Data Construction}
\label{sec:data}

Building a unified Embodied Foundation Model (EFM) demands a training corpus that is both large in scale and balanced across capability dimensions. We perform large-scale integration and restructuring of existing open-source resources, and design three automated data construction pipelines that generate proprietary data targeting critical capability gaps not covered by existing datasets. The resulting embodied data system encompasses 34 datasets with a total scale exceeding 15B tokens. Full data composition details and pipeline technical specifications are provided in the Appendix.

\textbf{Pipeline 1: 3D Scene Annotation for Spatial Reasoning.}
We first integrate multi-view spatiotemporal reasoning data (VLM-3R~\citep{vlm3r}, Cambrian-S~\citep{cambrians}, SAT~\citep{ray2024sat}), depth estimation data (metric depth data curated following DepthLM~\citep{cai2025depthlm}), and robot-view cognition data (Robo2VLM~\citep{wang2025robo2vlm}), covering spatial relations, distance metrics, and scene understanding. However, these datasets primarily correspond to room-level navigation scenarios and leave a significant gap in fine-grained tabletop manipulation spatial reasoning. To address this, we construct the \textit{ER1.5-Spatial} dataset ($\sim$20K samples): a fully automated pipeline reconstructs 3D semantic scene graphs from RGB images of real robot scenes (Fractal~\citep{brohan2022rt}, BridgeData~V2~\citep{walke2023bridgedata}, DROID~\citep{khazatsky2024droid}), chaining dual-backend semantic understanding, metric-scale monocular geometry estimation from MoGe-2~\citep{wang2025moge}, open-vocabulary instance segmentation (Grounded-SAM~\citep{ren2024grounded}), and RANSAC-based horizontal plane alignment, with multi-layer quality control embedded at each stage. Spatial reasoning QA pairs are then programmatically generated from the scene graphs to form ER1.5-Spatial.

\textbf{Pipeline 2: Failure-Aware Annotation for Planning and Correction.}
Planning data integrates RoboVQA~\citep{sermanet2024robovqa}, EgoPlan-IT~\citep{chen2024egoplan}, etc., covering long-horizon task decomposition and next-step planning. However, existing datasets only provide successful demonstrations and lack structured failure annotations. We therefore construct the \textit{ER1.5-Correction} dataset ($\sim$800K samples), motivated by the failure taxonomy frameworks of RoboFAC~\citep{ye2025robofac} and Guardian~\citep{pacaud2025guardian}. We organize the data along two orthogonal dimensions: by stage into planning failures and execution failures, and by cognitive level into failure detection, localization, and correction, yielding six QA types. The three cognitive levels directly mirror the complete error correction chain in closed-loop autonomous execution: detection discovers anomalies, localization diagnoses what went wrong and at which step, and correction generates a repair plan that transitions from the erroneous state back to the correct one. This layered design ensures the model can not only judge whether an error exists, but also precisely pinpoint its source and produce executable corrective instructions. For planning failures, five structured perturbation operators (step omission, redundancy, swap, object error, action replacement) are applied to correct sub-task plans, with each perturbation simultaneously generating three levels of QA. For execution failures, we combine video truncation to simulate interruption, description replacement to simulate object/action errors, and physics-engine perturbation injection in simulation to simulate manipulation failures. Data sources span real-world (BridgeData~V2~\citep{walke2023bridgedata}, RoboFail~\citep{liu2023reflect}, RoboFAC~\citep{ye2025robofac}) and simulated (ManiSkill~\citep{tao2024maniskill3}, GEMBench~\citep{garcia2024gembench}) scenarios. All types undergo perturbation validity verification, positive/negative balancing, and human sampling review.

\textbf{Pipeline 3: Affordance and Trajectory Data for Pointing.}
Pointing is the capability dimension where Embodied-R1.5 exhibits the largest advantage over existing models. We construct the \textit{ER1.5-Pointing} dataset, substantially expanding data coverage in both functional affordance and trace generation through automated pipelines. For affordance, since part-level annotation in the real world is extremely expensive, we overcome this bottleneck through large-scale data augmentation and restructuring: on one hand, we synthesize object functional part grounding data from simulation environments (ManiSkill-PartNet~\citep{xiang2020sapien} and PRISM~\citep{deshpande2025graspmolmo}); on the other hand, we systematically reorganize and rephrase multiple existing data sources, unifying their heterogeneous affordance annotations into the OFG training format to substantially scale up data volume and diversity. Additionally, inspired by CAPTURE~\citep{pothiraj2025capture}, we construct RegularRearrangement data that presents scenes with regular arrangement patterns (stars, squares, etc.) but with 1-2 elements missing, requiring the model to reason about where to place objects to complete the pattern, combining spatial reasoning with pointing capability. For trajectories, Embodied-R1.5 simultaneously supports both end-effector traces and object traces. We design a comprehensive automated extraction pipeline: for datasets with 3D end-effector poses (e.g., DROID~\citep{khazatsky2024droid}), poses are directly projected onto the 2D image plane; for datasets without metadata, we fine-tune a Detectron2~\citep{wu2019detectron2} based end-effector detector to track gripper motion; object traces leverage models such as Co-Tracker3~\citep{karaev2025cotracker3} to track manipulated object motion across frames. Additionally, we utilize the rich assets provided by RoboCasa~\citep{nasiriany2024robocasa} to generate 3D trace data for articulated object manipulation. We also incorporate large-scale manipulation trajectories from GenManip~\citep{gao2025genmanip} and InternData-A1~\citep{tian2026interndata}, which provide diverse simulated manipulation demonstrations for instruction-following tasks across a wide range of scenes and embodiments. Notably, the interaction motion semantics required by trace data do not demand high-fidelity simulation; low-fidelity simulated data still achieves strong generalization (we term this \textit{semantic generalization}). This is because VTG focuses on the topological structure and directional semantics of motion rather than rendering fidelity, so even low-fidelity simulation environments provide effective motion semantic supervision. After multiple rounds of quality filtering and coordinate normalization, the final ER1.5-Pointing dataset comprises $\sim$400K samples covering diverse affordance and trajectory annotations.

Finally, we balance the overall data mixture across all capability categories and incorporate substantial general visual cognition, logical reasoning, and instruction-following data~\citep{liu2024llava15,zhang2025bee,lian2025euclid,zhao2025mmifengine} as regularization to prevent catastrophic forgetting of general visual understanding.

\section{Training Strategy}
\label{sec:training}

Embodied-R1.5 adopts a two-stage paradigm: Supervised Fine-Tuning (SFT) builds foundational capability, and Reinforced Fine-Tuning (RFT) refines it via RL with verifiable rewards, particularly boosting pointing. We fine-tune on the full data system for one epoch, reserving low-resource but high-value tasks for the RL stage where verifiable rewards provide more efficient signals.

\subsection{Stage 1: Supervised Fine-Tuning}
\label{sec:training:sft}

Starting from the Qwen3-VL-8B-Instruct backbone, we perform full-parameter fine-tuning on the SFT data system described in Section~\ref{sec:data} with the standard causal language modeling objective. All heterogeneous capability outputs (natural language reasoning, point coordinates, trajectory sequences, etc.) are uniformly represented as token sequences and jointly optimized under the same objective for one epoch. We adopt the AdamW optimizer with bfloat16 precision, a peak learning rate of $2{\times}10^{-6}$ with cosine decay and 10\% warmup, and an effective global batch size of 512. The maximum context length is 8192 tokens. The vision encoder is jointly trained with the LLM backbone without freezing, as the visual distribution of embodied scenarios differs significantly from general pretraining data and requires deep adaptation of visual features. The goal of the SFT stage is to establish a multi-task foundation covering all capability dimensions while providing a strong initialization for the RFT stage, ensuring that rollouts across different tasks exhibit sufficient discriminability to produce effective RL signals.

\subsection{Stage 2: Reinforced Fine-Tuning}
\label{sec:training:rft}

\subsubsection{Multi-Task Balanced RL Recipe}
\label{sec:training:balance}

While GRPO~\citep{shao2024deepseekmath} excels at single-task RFT, heterogeneous multi-task training introduces two complementary imbalances~\citep{feng2025onethinker}: (1)~\textit{Intra-task}: group-level std normalization biases updates toward low-variance samples, under-optimizing medium-difficulty samples where learning signal is strongest; (2)~\textit{Inter-task}: vastly different reward scales across capabilities cause high-density tasks to dominate gradients. We address both at the data and normalization levels:

\textit{(1) Difficulty-aware data filtering.} We filter the SFT corpus via rollout pass rates from the SFT checkpoint, retaining $\sim$200K medium-difficulty samples that maximize learning signal.

\textit{(2) Dynamic filtering.} When all rollouts in a group receive identical rewards, no effective gradient signal can be produced. We automatically detect and mask such degenerate groups during training, ensuring gradient updates come only from samples with discriminative reward variation.

\textit{(3) Global batch reward normalization.} Standard GRPO normalizes by group-level std, which cannot eliminate cross-task reward scale differences since each group contains rollouts from only a single query. EMA-GRPO~\citep{feng2025onethinker} introduces per-task moving averages, but low-resource task statistics become unstable when data volumes differ substantially. We instead compute advantages as $\hat{A}_i = (R_i - \mu_{\text{group}})/(\sigma_{\text{batch}} + \epsilon)$, where $\mu_{\text{group}}$ is the within-group mean and $\sigma_{\text{batch}}$ is the std over the entire mixed batch. The group-level mean preserves intra-group relative ordering (which rollout is better), while the batch-level std unifies gradient magnitudes across tasks, without requiring task labels or historical state.

\textit{(4) Multi-task reward design.} We design five families of reward functions matched to output structure: exact matching (0/1), IoU (spatial grounding), point distance (decay), trajectory RMSE (continuous), and semantic similarity (RM scoring). All continuous rewards use piecewise-linear decay to provide partial credit. Details are presented in Section~\ref{sec:training:reward}.

\textit{(5) Adaptive thinking.} We only constrain the output format (requiring answers within \texttt{<answer></answer>} tags) without forcing explicit reasoning chains. Since the RL reward evaluates only the final answer quality without rewarding intermediate reasoning, the model naturally learns to allocate computation on demand during optimization: for perception-oriented simple tasks such as pointing, additional reasoning tokens yield no reward gain, so the model learns to output coordinates with near-zero reasoning overhead; for complex planning tasks requiring multi-step reasoning, thorough thinking significantly improves answer quality and thus receives higher reward, so the model generates structured thinking processes. This emergent adaptive computation allocation naturally aligns with embodied scenario requirements: high-frequency localization tasks demand real-time responsiveness, while lower-frequency planning tasks benefit from more thorough reasoning.

\textbf{Training setup.} The RL stage is initialized from the SFT-stage checkpoint. We use an improved EasyR1 framework\footnote{\url{https://github.com/hiyouga/EasyR1}} extended to support mixed video-and-image inputs as well as external reward model scoring. We set the clip ratio lower/upper thresholds to 0.2/0.28; since embodied reasoning responses are relatively short, the actual clipping rate is extremely low. KL regularization adopts the KL-in-reward formulation with K1 unbiased KL estimation directly as a reward penalty term ($\beta = 0.01$), constraining the policy from drifting too far from the SFT distribution. Each prompt is sampled with $n{=}8$ parallel rollouts. Training uses AdamW with a learning rate of $3{\times}10^{-6}$ and gradient clipping at 1.0. The vision encoder is jointly trained with the LLM backbone, and the model is trained for 2 epochs.

\subsubsection{Multi-Task Reward Design}
\label{sec:training:reward}

We design five families of reward functions for the heterogeneous output types in embodied reasoning. For continuous-valued rewards, we employ a unified piecewise-linear decay framework: given a distance metric~$d$ and threshold pair~$(\tau_p,\tau_z)$,
\begin{equation}
    \phi(d;\,\tau_p,\tau_z) = \mathrm{clip}\!\left(\frac{\tau_z - d}{\tau_z - \tau_p},\;0,\;1\right),
\end{equation}
which assigns full credit when $d < \tau_p$, zero credit when $d \geq \tau_z$, and linearly decays in between. This partial-credit design provides dense gradient signals for RL training, avoiding the sparsity inherent in binary rewards.

\textbf{(1) Exact-match reward (deterministic-answer tasks).}
For multiple-choice questions, numerical comparisons, and mathematical derivations that admit a single correct answer, we use a binary reward $R = \mathbb{1}[\mathrm{match}(\hat{y},\,y^*)]$, outputting $\{0,1\}$. Multiple-choice grading employs a flexible matching mechanism that accepts answer variants such as ``A'', ``A.dog'', while ordering questions are treated as strict exact matches. Numerical answers are rounded to one decimal place before comparison. Math problems use symbolic equivalence verification to determine whether two mathematical expressions are equivalent.

\textbf{(2) IoU reward (spatial grounding tasks).}
For tasks whose output is a bounding box, we compute the standard 2D Intersection-over-Union between the predicted box $b$ and the ground-truth box $b^*$: $R_{\mathrm{IoU}} = {|b \cap b^*|}/{|b \cup b^*|}$, yielding a reward in $[0,1]$. The continuous IoU naturally provides partial credit.

\textbf{(3) Point-distance reward (point localization tasks).}
For point localization tasks, we compute the \textit{average nearest-neighbor distance} $d_{\mathrm{nn}}$ between the predicted and ground-truth point sets, then apply the piecewise-linear decay: $R = \phi(d_{\mathrm{nn}};\,40,\,150)$. When the ground truth is a segmentation polygon or bounding box, we switch to a \textit{point-in-region} check and reward the fraction of predicted points that fall within the target region. An additional count penalty $\delta_c = 0.3$ is applied when the number of predicted points does not match the ground truth.

\textbf{(4) Trajectory-RMSE reward (trajectory prediction tasks).}
For 2D and 3D visual trace tasks, we first align predicted and ground-truth trajectories to an equal number of sampling points via linear interpolation, then use per-point RMSE as the distance metric: $R_{2\mathrm{D}} = \phi(\mathrm{RMSE}_{2\mathrm{D}};\,50,\,120)$. For 3D trajectories, the depth dimension is independently evaluated via MAE: $R_{\mathrm{depth}} = \phi(\mathrm{MAE}_d;\,0.1,\,0.4)$, and the final reward is an equal-weight average $R = 0.5\,R_{2\mathrm{D}} + 0.5\,R_{\mathrm{depth}}$. A length-mismatch penalty $\delta_l = 0.35$ is imposed when the predicted and ground-truth trajectory lengths differ, and single-point outputs are assigned zero reward to prevent reward hacking.

\textbf{(5) Semantic-similarity reward (open-ended text tasks).}
For open-ended outputs such as error correction suggestions and long-horizon plans, we use Skywork-Reward-V2-Qwen3-4B~\citep{liu2025skywork} as the default reward model for semantic quality scoring; the raw reward-model score is mapped to $[0,1]$ via sigmoid temperature normalization. When reward-model inference fails, we fall back to BLEU scores to measure semantic consistency.

\textbf{Format reward.}
Across all tasks, we include a unified format-checking reward that jointly verifies two conditions: (1)~the output contains the required \texttt{<answer>...</answer>} tag structure, and (2)~the content inside the tags conforms to the task-specific structural format (e.g., point tasks require \texttt{point\_2d} as a two-element numeric list; trajectory tasks require a \texttt{depth} field). The format reward is binary and combined with the accuracy reward via a weighted sum:
\begin{equation}
    R = (1 - \lambda)\,R_{\mathrm{acc}} + \lambda\,R_{\mathrm{fmt}}, \quad \lambda = 0.1.
\end{equation}

\section{Closed-Loop PGC Autonomy Framework}
\label{sec:framework}

\begin{figure}[t]
\centering
\includegraphics[width=\linewidth]{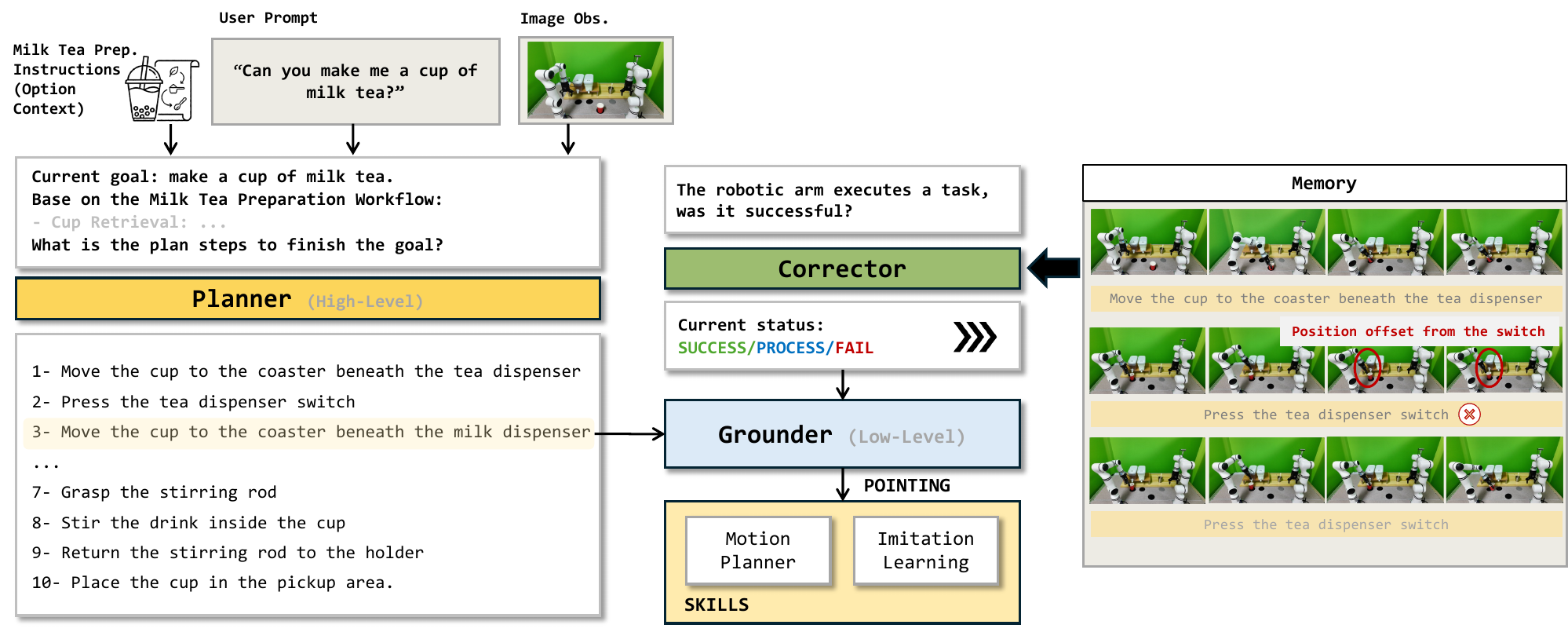}
\vspace{-15pt}
\caption{\textbf{Planner-Grounder-Corrector (PGC) closed-loop framework.} A single Embodied-R1.5 instance asynchronously serves all three roles, maintaining a minimal FIFO memory buffer. The example shows a milk tea preparation task: the Planner decomposes it into sub-tasks, the Grounder provides spatial grounding for execution, and the Corrector continuously monitors progress, triggering replanning upon failure.}
\vspace{-15pt}
\label{fig:framework}
\end{figure}

To validate whether the three capability dimensions can truly ground in physics under long-horizon execution, we design the minimalist \textbf{Planner-Grounder-Corrector (PGC)} closed-loop framework. The core design principle is to deploy a single Embodied-R1.5 model as a unified inference service and asynchronously invoke its different capability interfaces to drive the full autonomy stack, requiring no multi-model cascading or multi-agent orchestration.

\Cref{fig:framework} illustrates the complete execution flow using ``making a cup of milk tea'' as an example. The system receives a user instruction, the current image observation, and optional external context (e.g., a task SOP document) as input. The \textbf{Planner} (high-level) first invokes the planning capability to decompose the long-horizon task into a structured sub-task sequence; as execution progresses, the Planner performs next-step planning after each sub-task completion based on the latest observation, dynamically adjusting subsequent task arrangements. Execution then proceeds step by step: for each sub-task, the \textbf{Grounder} (low-level) adaptively orchestrates a combination of pointing capabilities, autonomously deciding which capability interfaces to invoke based on requirements (e.g., ``press the tea dispenser switch'' requires first localizing the switch as a functional part via OFG, then planning a pressing trajectory via VTG), producing precise spatial grounding commands that are passed to a low-level skill executor. During execution, the \textbf{Corrector} runs as an asynchronous query, continuously assessing the execution state based on the current observation and historical information in Memory: SUCCESS, PROCESS (still executing, continue waiting), or FAIL. The entire task maintains a \textbf{Memory} module with a simple first-in-first-out (FIFO) buffer design, performing fixed-rate image sampling during execution and recording status annotations for each sub-task. While more sophisticated memory strategies~\citep{torne2026mem,shi2025memoryvla} could further enhance this design, the current minimalist buffer already suffices for effective closed-loop correction. When the Corrector detects a failure, the error information is written into Memory and fed back to the Planner, triggering a retry or adaptive replanning for that sub-task. As visible in the figure, the same sub-task ``press the tea dispenser switch'' successfully completes on the second attempt after an initial failure, demonstrating the autonomous recovery capability of correction.

Since the Planner, Grounder, and Corrector roles are all served by the same Embodied-R1.5 instance, the Corrector can directly leverage the reasoning context established during the Planner phase and the spatial understanding from the Grounder, avoiding the information loss inherent in cascade systems and enabling more accurate error attribution. Notably, the PGC framework itself is merely a lightweight stateless harness responsible for control flow scheduling and Memory management, containing no reasoning logic or heuristic rules; all intelligent decisions including task understanding and error attribution are entirely driven by the internalized capabilities of the Embodied-R1.5 model. The framework's ceiling is determined solely by the model's capability rather than system engineering complexity, endowing it with the property of naturally scaling as model capability improves. This framework enables long-chain real-world tasks to be completed autonomously without human intervention.

\section{EmbodiedEvalKit}
\label{sec:evalkit}

\begin{figure*}[t]
\centering
\includegraphics[width=\textwidth]{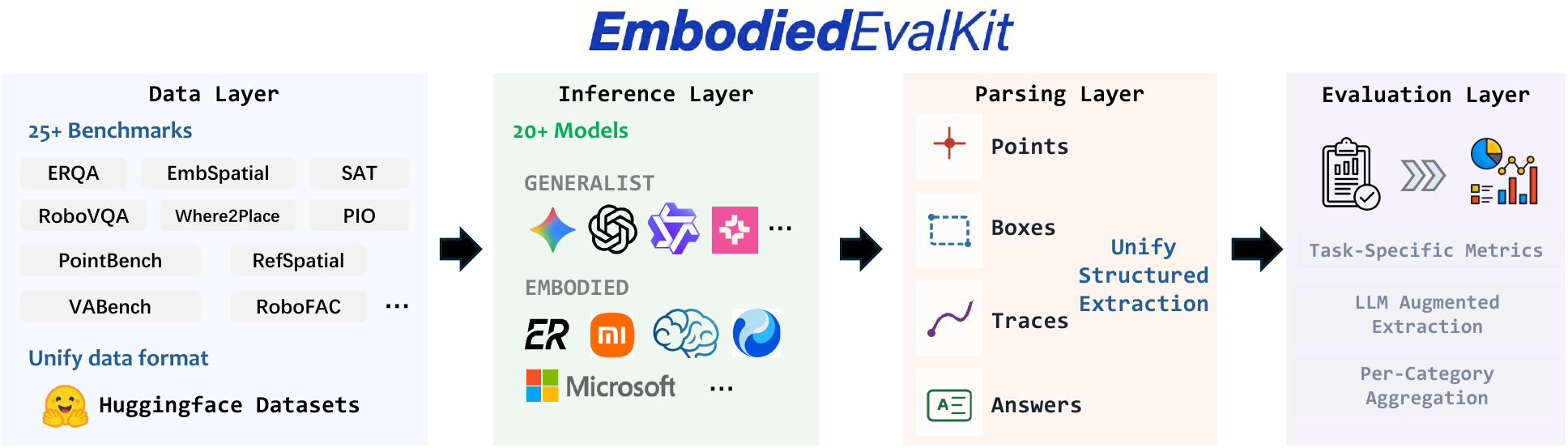}
\caption{\textbf{Architecture of EmbodiedEvalKit.} A four-layer modular framework supporting 25+ embodied benchmarks and 20+ models in a single reproducible evaluation pipeline.}
\label{fig:evalkit}
\end{figure*}

The general-purpose VLM community benefits from mature evaluation frameworks such as VLMEvalKit~\citep{duan2024vlmevalkit} and lmms-eval~\citep{zhang2024lmmseval}, yet these tools are designed for standard VQA and captioning tasks and cannot handle the unique requirements of embodied evaluation: parsing point coordinates, bounding boxes, and trajectory sequences from model outputs, unifying different coordinate output formats across models (e.g., normalized to 1000, absolute pixels, etc.), and computing embodied-specific metrics. As a result, existing works rely on ad-hoc evaluation scripts with different grounding format conventions and often evaluate on non-overlapping benchmark subsets, making cross-paper comparison unreliable. To bridge this gap, we develop and open-source \textbf{EmbodiedEvalKit}, a unified evaluation framework designed specifically for embodied VLMs.

As illustrated in Figure~\ref{fig:evalkit}, EmbodiedEvalKit adopts a four-layer modular design. The \emph{data layer} reorganizes diverse embodied benchmarks spanning all three capability dimensions into a standardized HuggingFace Parquet format, enabling a single data loading pipeline for all tasks. The \emph{inference layer} provides a model-agnostic interface supporting vLLM~\citep{kwon2023efficient}, HuggingFace Transformers, and API backends, with a \emph{backbone} abstraction that automatically handles model-specific differences in coordinate systems and prompt templates, allowing both open-source and proprietary models to be evaluated within the same framework. The \emph{parsing layer} detects and normalizes heterogeneous grounding outputs into a unified structured representation. Finally, the \emph{evaluation layer} computes official metrics for each benchmark and aggregates results by capability category. EmbodiedEvalKit is fully open-sourced and extensible: adding a new benchmark requires only defining a data adapter and an evaluator following standardized templates. We release pre-processed evaluation data for all supported benchmarks to ensure exact reproducibility. The framework currently supports 25+ embodied benchmarks and 20+ models; the experiments in this paper report results on 24 benchmarks and 13 baselines, all evaluated through this unified pipeline.

\section{Experiments}
\label{sec:experiments}
\vspace{-10pt}

\begin{figure}[h]
\centering
\includegraphics[width=\linewidth]{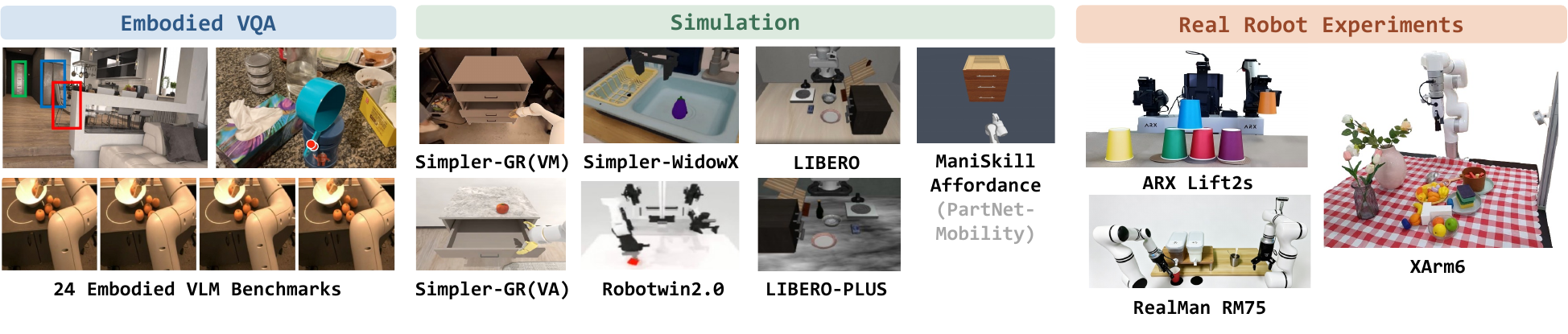}
\caption{\textbf{Overview of the experimental evaluation of Embodied-R1.5.} We organize the evaluation into five parts spanning Embodied VLM benchmarks, manipulation benchmarks, zero-shot real-world manipulation, long-horizon closed-loop demonstrations, and ablation analysis.}
\label{fig:exp-overview}
\end{figure}

\begin{table*}[t]
\centering
\footnotesize
\caption{\textbf{Embodied Planning \& Correction benchmark results.} \textbf{Bold}: best; \underline{underline}: second best.}
\label{tab:vlm-planning}
\vspace{2pt}
\renewcommand{\arraystretch}{1.1}
\setlength{\tabcolsep}{5pt}
\begin{tabular}{l|cccc|c}
\toprule
\textbf{Model} & RoboVQA & EgoPlan-2 & Cosmos & RoboFAC & \textbf{Avg} \\
\midrule
\multicolumn{6}{l}{\cellcolor{gray!8}\textit{General-Purpose VLMs}} \\
GPT-4o & 34.5 & 41.8 & 53.3 & 48.9 & 44.6 \\
GPT-5.4 & 31.8 & \underline{44.1} & \underline{69.2} & \underline{70.1} & 53.8 \\
Gemini-2.5-Pro & 33.9 & 42.9 & 62.8 & 46.4 & 46.5 \\
Qwen3-VL-8B & 55.0 & 32.3 & 64.0 & 65.4 & \underline{54.2} \\
InternVL3.5-8B & 28.6 & 31.6 & 48.2 & 48.6 & 39.2 \\
Molmo-D-7B & 23.0 & 26.9 & 22.2 & 11.5 & 20.9 \\
\midrule
\multicolumn{6}{l}{\cellcolor{gray!8}\textit{Embodied VLMs}} \\
Gemini-Robotics-ER-1.5 & 32.6 & 32.0 & 53.7 & 47.0 & 41.3 \\
RoboBrain-2.0-7B & 57.5 & 33.2 & 33.8 & 54.2 & 44.7 \\
VeBrain & 42.4 & 27.3 & 58.8 & 58.8 & 46.8 \\
Mimo-Embodied & \textbf{62.0} & 43.0 & 56.8 & 54.6 & 54.1 \\
Pelican-VL-1.0 & 58.5 & 32.0 & 62.6 & 60.2 & 53.3 \\
Magma & 14.3 & 4.1 & 32.8 & 18.1 & 17.3 \\
Embodied-R1 & 51.8 & 26.5 & 55.2 & 40.2 & 43.4 \\
\midrule
\multicolumn{6}{l}{\cellcolor{gray!8}\textit{Ours}} \\
\rowcolor{blue!6}
\textbf{Embodied-R1.5} & \underline{61.0} & \textbf{53.8} & \textbf{69.3} & \textbf{77.2} & \textbf{65.3} \\
\bottomrule
\end{tabular}
\end{table*}

We organize the evaluation into five parts: (1)~Embodied VLM benchmarks comprehensively validate the breadth and depth of foundational capabilities (\S\ref{sec:exp:vlm}); (2)~Manipulation benchmarks examine whether these embodied capabilities are truly internalized and transfer to downstream action execution (\S\ref{sec:exp:vla}); (3)~zero-shot robot manipulation experiments verify generalization from benchmarks to the physical world (\S\ref{sec:exp-realworld}); (4)~long-horizon closed-loop demonstrations validate the full system's autonomous capability (\S\ref{sec:exp:closedloop}); (5)~ablation analysis answers the effectiveness of key design choices (\S\ref{sec:exp:ablation}).

\subsection{Embodied VLM Benchmarks}
\label{sec:exp:vlm}

\textbf{Setup.} All evaluations are conducted through EmbodiedEvalKit, ensuring full reproducibility under a unified pipeline. We evaluate Embodied-R1.5 on a comprehensive suite of 24 embodied VLM benchmarks organized along three capability dimensions. \textit{Embodied Planning \& Correction} (4 benchmarks, \cref{tab:vlm-planning}): RoboVQA~\citep{sermanet2024robovqa}, EgoPlan-2~\citep{egoplan2bench2024}, Cosmos-Reason~\citep{azzolini2025cosmos}, and RoboFAC~\citep{ye2025robofac}. \textit{Embodied Pointing \& Location} (9 pointing benchmarks + 3 visual trace benchmarks, \cref{tab:vlm-pointing,tab:vlm-trace}): VAbench-P~\citep{yuan2025seeing}, Where2Place~\citep{yuan2024robopoint}, RefSpatial~\citep{refspatial2025}, Part-Afford~\citep{yuan2025embodiedr}, RoboRefit~\citep{roborefit2023}, RoboAfford~\citep{roboafford2025}, PointBench~\citep{pointbench2025}, PIO~\citep{piobench2025}, Pixmo-Point~\citep{deitke2025molmo}, ShareRobot-V, VABench-V~\citep{yuan2025seeing}, and PIO-S3-Verified~\citep{piobench2025}. \textit{Embodied Cognition \& Spatial Reasoning} (8 benchmarks, \cref{tab:vlm-cognition}): ERQA~\citep{team2025gemini}, OpenEQA~\citep{majumdar2024openeqa}, CV-Bench~\citep{cvbench2024}, EmbSpatial~\citep{embspatial2024}, SAT~\citep{ray2024sat}, RoboSpatial~\citep{robospatial2024}, BLINK~\citep{blink2024}, and VSIBench~\citep{vsibench2024}. We additionally evaluate on 7 standard general vision benchmarks (\cref{tab:general-bench}) to verify capability retention. To prevent data contamination, we perform rigorous deduplication between our training corpus and all evaluation benchmarks. All reported results use official held-out test splits.

We compare against two categories of baselines: (a)~\textit{General-purpose VLMs}: GPT-4o, GPT-5.4, Gemini-2.5-Pro, Qwen3-VL-8B~\citep{bai2025qwen3vl}, InternVL3.5-8B~\citep{wang2025internvl3}, and Molmo-D-7B~\citep{deitke2025molmo}; (b)~\textit{Embodied VLMs}: Gemini-Robotics-ER-1.5~\citep{team2025gemini}, RoboBrain-2.0-7B~\citep{ji2025robobrain}, VeBrain~\citep{luo2025visual}, Mimo-Embodied~\citep{hao2025mimo}, Pelican-VL-1.0~\citep{zhang2025pelican}, Magma~\citep{yang2025magma}, and Embodied-R1~\citep{yuan2025embodiedr}.

\begin{table*}[t]
\centering
\footnotesize
\caption{\textbf{Embodied Pointing \& Location benchmark results.} $\ddagger$\,S1\&S2 Subset; $\S$\,All Split.}
\label{tab:vlm-pointing}
\vspace{2pt}
\renewcommand{\arraystretch}{1.1}
\setlength{\tabcolsep}{3pt}
\resizebox{\textwidth}{!}{%
\begin{tabular}{l|ccccccccc|c}
\toprule
\textbf{Model} & VAbench-P & Where2Place & RefSpatial\textsuperscript{$\S$} & Part-Afford & RoboRefit & RoboAfford & PointBench & PIO\textsuperscript{$\ddagger$} & Pixmo-Pt & \textbf{Avg} \\
\midrule
\multicolumn{11}{l}{\cellcolor{gray!8}\textit{General-Purpose VLMs}} \\
GPT-4o & 13.7 & 20.4 & 8.8 & 13.3 & 14.2 & 20.5 & 29.5 & 13.4 & 13.5 & 16.4 \\
GPT-5.4 & 25.7 & 34.7 & 15.7 & 17.2 & 29.2 & 29.4 & 37.9 & 19.3 & 27.5 & 26.3 \\
Gemini-2.5-Pro & 21.7 & 49.6 & 36.5 & 25.5 & 38.4 & 23.4 & 62.8 & 23.2 & 12.7 & 32.6 \\
Qwen3-VL-8B & 42.3 & 64.0 & 42.2 & 48.7 & 83.6 & \underline{69.9} & 65.6 & \underline{56.8} & \underline{59.9} & \underline{59.2} \\
InternVL3.5-8B & 23.0 & 34.8 & 16.8 & 22.7 & 30.8 & 31.5 & 32.4 & 23.1 & 44.1 & 28.8 \\
Molmo-D-7B & 38.0 & 39.3 & 45.4 & \underline{73.2} & 81.1 & 61.7 & 61.6 & 44.3 & 51.9 & 55.2 \\
\midrule
\multicolumn{11}{l}{\cellcolor{gray!8}\textit{Embodied VLMs}} \\
Gemini-Robotics-ER-1.5 & 41.6 & 48.3 & 39.7 & 27.0 & 80.0 & 57.6 & \underline{70.7} & 54.0 & 52.5 & 52.4 \\
RoboBrain-2.0-7B & 41.0 & 63.6 & 32.5 & 45.6 & 70.4 & 51.5 & 60.4 & 54.0 & 54.7 & 52.6 \\
VeBrain & 1.7 & 12.3 & 0.3 & 32.1 & 32.2 & 2.1 & 20.2 & 24.1 & 16.8 & 15.8 \\
Mimo-Embodied & 46.9 & 63.6 & \underline{48.0} & 65.5 & 82.3 & 69.8 & 50.2 & 27.5 & 42.4 & 55.1 \\
Pelican-VL-1.0 & 14.5 & 57.8 & 37.5 & 28.8 & 74.9 & 63.4 & 36.4 & 32.9 & 26.1 & 41.4 \\
Magma & 6.6 & 10.9 & 4.5 & 0.3 & 5.1 & 13.4 & 23.4 & 4.6 & 12.8 & 9.1 \\
Embodied-R1 & \underline{66.0} & \underline{69.5} & 39.7 & 56.6 & \underline{85.6} & 67.2 & 49.7 & 44.4 & 49.4 & 58.7 \\
\midrule
\multicolumn{11}{l}{\cellcolor{gray!8}\textit{Ours}} \\
\rowcolor{blue!6}
\textbf{Embodied-R1.5} & \textbf{73.7} & \textbf{74.0} & \textbf{54.2} & \textbf{82.9} & \textbf{88.4} & \textbf{80.0} & \textbf{74.6} & \textbf{62.6} & \textbf{64.8} & \textbf{72.8} \\
\bottomrule
\end{tabular}%
}
\end{table*}

\begin{table*}[t]
\centering
\footnotesize
\caption{\textbf{Embodied Cognition \& Spatial Reasoning benchmark results.} $\dagger$\,Rel.~Depth Subset.}
\label{tab:vlm-cognition}
\vspace{2pt}
\renewcommand{\arraystretch}{1.1}
\setlength{\tabcolsep}{3pt}
\resizebox{\textwidth}{!}{%
\begin{tabular}{l|cccccccc|c}
\toprule
\textbf{Model} & ERQA & OpenEQA & CV-Bench & EmbSpatial & SAT & RoboSpatial & BLINK\textsuperscript{$\dagger$} & VSIBench & \textbf{Avg} \\
\midrule
\multicolumn{10}{l}{\cellcolor{gray!8}\textit{General-Purpose VLMs}} \\
GPT-4o & 32.5 & 51.1 & 78.6 & 71.9 & 66.7 & 44.4 & 64.5 & 43.6 & 56.7 \\
GPT-5.4 & \underline{50.5} & \textbf{69.3} & 76.7 & 73.2 & 74.7 & 53.1 & \underline{87.9} & 52.6 & 67.3 \\
Gemini-2.5-Pro & \textbf{55.7} & \underline{65.8} & 84.6 & \textbf{78.7} & \textbf{76.7} & 59.9 & 85.5 & 43.4 & \underline{68.8} \\
Qwen3-VL-8B & 41.8 & 65.1 & 86.3 & \underline{78.5} & 65.3 & \underline{65.4} & 78.2 & 55.1 & 67.0 \\
InternVL3.5-8B & 41.0 & 42.9 & 81.5 & 70.3 & 55.3 & 51.1 & 75.0 & \underline{55.7} & 59.1 \\
Molmo-D-7B & 40.6 & 35.1 & 67.9 & 58.7 & 54.0 & 36.0 & 75.0 & 6.7 & 46.8 \\
\midrule
\multicolumn{10}{l}{\cellcolor{gray!8}\textit{Embodied VLMs}} \\
Gemini-Robotics-ER-1.5 & 48.5 & 50.5 & 83.6 & 73.4 & 62.7 & 40.1 & 85.9 & 39.9 & 60.6 \\
RoboBrain-2.0-7B & 38.5 & 24.2 & 85.8 & 76.3 & 75.3 & 54.2 & 79.0 & 36.1 & 58.7 \\
VeBrain & 37.3 & 27.8 & 79.7 & 70.5 & 58.0 & 42.5 & 71.8 & 36.2 & 53.0 \\
Mimo-Embodied & 46.8 & 53.6 & \textbf{88.8} & 76.2 & 54.6 & 61.8 & \textbf{94.4} & 48.5 & 65.6 \\
Pelican-VL-1.0 & 39.8 & 54.7 & 78.9 & 73.2 & 54.6 & 57.5 & 78.2 & 52.8 & 61.2 \\
Magma & 25.4 & 32.7 & 65.9 & 64.6 & 71.3 & 33.7 & 58.9 & 17.1 & 46.2 \\
Embodied-R1 & 35.2 & 34.5 & 82.7 & 67.4 & \underline{76.3} & 47.4 & 76.6 & 26.6 & 55.8 \\
\midrule
\multicolumn{10}{l}{\cellcolor{gray!8}\textit{Ours}} \\
\rowcolor{blue!6}
\textbf{Embodied-R1.5} & 46.0 & 62.6 & \underline{86.9} & 78.1 & 74.7 & \textbf{69.7} & 87.1 & \textbf{56.1} & \textbf{70.2} \\
\bottomrule
\end{tabular}%
}
\end{table*}

\begin{table*}[t]
\centering
\footnotesize
\caption{\textbf{Visual Trace results.} $\downarrow$: lower is better. PIO-S3-Verified uses improved LLM evaluation for higher scoring accuracy.}
\label{tab:vlm-trace}
\vspace{2pt}
\renewcommand{\arraystretch}{1.1}
\setlength{\tabcolsep}{4pt}
\begin{tabular}{l|cc|cc|c}
\toprule
& \multicolumn{2}{c|}{\textbf{ShareRobot-V}} & \multicolumn{2}{c|}{\textbf{VABench-V}} & \textbf{PIO-S3-Verified} \\
\cmidrule(lr){2-3} \cmidrule(lr){4-5} \cmidrule(lr){6-6}
\textbf{Model} & RMSE$\downarrow$ & DFD$\downarrow$ & RMSE$\downarrow$ & DFD$\downarrow$ & Score$\uparrow$ \\
\midrule
GPT-5.4 & 23.81 & 33.55 & 18.20 & 26.50 & 44.90 \\
Qwen3-VL-8B & 22.58 & 32.47 & 15.30 & 21.80 & 46.60 \\
RoboBrain 2.0 & \textbf{13.23} & \textbf{19.63} & 18.09 & 26.68 & \underline{46.90} \\
Embodied-R1 & 23.20 & 32.90 & \underline{9.04} & \underline{12.21} & \underline{46.90} \\
\rowcolor{blue!6}
\textbf{Embodied-R1.5} & \underline{15.79} & \underline{22.64} & \textbf{7.00} & \textbf{9.83} & \textbf{49.56} \\
\bottomrule
\end{tabular}
\end{table*}

\textbf{Results.} Embodied-R1.5 achieves state-of-the-art on 16 out of 24 embodied benchmarks across all three capability dimensions.

\textit{Planning \& Correction} (\cref{tab:vlm-planning}): Embodied-R1.5 achieves the highest average (65.3), with RoboFAC (77.2\%) directly validating our correction data pipeline and EgoPlan-2 (53.8\%) confirming egocentric planning. The +24.0pp margin over Gemini-Robotics-ER-1.5 demonstrates a decisive advantage in planning and fault awareness.

\textit{Pointing \& Location} (\cref{tab:vlm-pointing,tab:vlm-trace}): This dimension exhibits the most dominant advantage, with Embodied-R1.5 achieving SOTA on all 9 pointing benchmarks (avg 72.8). Notably, Part-Afford (82.9\%) and RoboAfford (80.0\%) validate the substantial improvement brought by our OFG functional part grounding pipeline, while RefSpatial (54.2\%) and Where2Place (74.0\%) demonstrate clear leadership on RRG region grounding tasks. On visual trace (\cref{tab:vlm-trace}), Embodied-R1.5 performs strongly on both object-centric flow (VABench-V, RMSE 7.00) and robot-centric flow (ShareRobot-V, RMSE 15.79).

\textit{Cognition \& Spatial Reasoning} (\cref{tab:vlm-cognition}): Embodied-R1.5 achieves the highest dimension average (70.2) among all models, with SOTA on RoboSpatial (69.7\%). The margins on this dimension are relatively smaller compared to Pointing, as strong general-purpose VLMs like Gemini-2.5-Pro and GPT-5.4 already exhibit excellent spatial reasoning on individual benchmarks. Nevertheless, Embodied-R1.5 maintains competitive performance across all 8 benchmarks without significant weakness on any single task, confirming that embodied training does not sacrifice general spatial understanding.

\begin{table*}[t]
\centering
\footnotesize
\caption{\textbf{General vision benchmark results.} Embodied-R1.5 retains general capability relative to the base model.}
\label{tab:general-bench}
\vspace{2pt}
\renewcommand{\arraystretch}{1.1}
\setlength{\tabcolsep}{4pt}
\begin{tabular}{l|ccccccc}
\toprule
\textbf{Model} & AI2D & SEED & POPE & MMMU & RealWQA & ChartQA & SciQA \\
\midrule
Qwen3-VL-8B & \textbf{85.7} & \textbf{77.6} & 88.1 & 52.0 & \textbf{71.3} & \textbf{83.0} & 92.3 \\
\rowcolor{blue!6}
\textbf{Embodied-R1.5} & 83.3 & 76.5 & \textbf{88.2} & \textbf{54.8} & 69.4 & \textbf{83.0} & \textbf{93.1} \\
\bottomrule
\end{tabular}
\end{table*}

\textit{General capability retention} (\cref{tab:general-bench}): Evaluation on 7 standard vision benchmarks confirms minimal degradation from the base model Qwen3-VL-8B, with MMMU and ScienceQA actually improving. This demonstrates that our embodied training pipeline preserves general visual understanding while substantially enhancing embodied capabilities.

\subsection{Robotic Manipulation in Simulation}
\label{sec:exp:vla}

\textbf{Setup.} To examine whether embodied reasoning capabilities are truly internalized and can transfer to downstream action execution, we build \textit{Embodied-R1.5-VLA} by attaching a lightweight flow-matching action expert to the Embodied-R1.5 backbone and fine-tuning on benchmark data directly, without any large-scale action pretraining. Embodied-R1.5-VLA adopts a dual-system architecture: the VLM backbone produces hidden representations of dimension 2048, and a DiT-B action head uses VLM hidden states as cross-attention context, combined with 32 learnable future query tokens, to generate 7-dimensional continuous action trajectories via flow matching. Training uses AdamW ($\beta_1{=}0.9$, $\beta_2{=}0.95$) with action model learning rate $10^{-4}$ and VLM learning rate $10^{-5}$, total batch size 128 on 8 NVIDIA H20-96GB GPUs. During inference, actions are generated from Gaussian noise via 4-step Euler integration. All VLA experiments are conducted using the starVLA~\citep{community2026starvla} framework.

We evaluate on three benchmark suites: SimplerEnv~\citep{li2024evaluating}, LIBERO~\citep{liu2023libero}, and LIBERO-Plus~\citep{fei2025libero}. For SimplerEnv, the action prediction horizon is 16 steps with delta end-effector actions, single camera view at $224{\times}224$, trained on a mixture of BridgeData~\citep{walke2023bridgedata} and Fractal~\citep{brohan2022rt} datasets for 80K steps. For LIBERO, the action prediction horizon is 8 steps with delta joint position actions, dual-view observations at $224{\times}224$, trained on all four suites for 80K steps; we run 50 episodes per task for evaluation. LIBERO-Plus directly uses the LIBERO checkpoint for testing, following the official evaluation protocol.

We compare against a comprehensive set of baselines: RT-1-X and RT-2-X~\citep{brohan2022rt,brohan2023rt2,openxembodiment2023}, OpenVLA~\citep{openvla2024}, OpenVLA-OFT~\citep{kim2025openvlaoft}, $\pi_0$~\citep{pi0_2024}, $\pi_0$-FAST~\citep{pertsch2025fast}, $\pi_{0.5}$~\citep{intelligence2025visionlanguageaction}, GR00T-N1.5/N1.6~\citep{groot_n15}, SpatialVLA~\citep{spatialvla2025}, and Diffusion Policy~\citep{chi2023diffusionpolicy}.

\begin{table*}[t]
\centering
\footnotesize
\caption{\textbf{SimplerEnv: Google Robot (Visual Matching).} Success rate (\%). Results marked with ``$-$'' are sourced from starVLA~\citep{community2026starvla}.}
\label{tab:simpler-vm}
\vspace{2pt}
\renewcommand{\arraystretch}{1.1}
\setlength{\tabcolsep}{4pt}
\begin{tabular}{l|cccc|c}
\toprule
\textbf{Method} & \textbf{Pick Coke} & \textbf{Move Near} & \textbf{Open/Close Drawer} & \textbf{Open Top Drawer
and Place Apple} & \textbf{Overall} \\
\midrule
RT-1-X & 56.7 & 31.7 & 59.7 & 21.3 & 42.4 \\
RT-2-X & 78.7 & 77.9 & 25.0 & 3.7 & 46.3 \\
OpenVLA & 16.3 & 46.2 & 35.6 & 0.0 & 24.5 \\
SpatialVLA & 86.0 & 77.9 & 57.4 & 0.0 & 55.3 \\
OpenVLA-OFT & 72.3 & 69.6 & 47.2 & -- & 63.0 \\
$\pi_0$ & \textbf{97.9} & \underline{78.7} & \underline{62.2} & \underline{46.6} & 71.4 \\
$\pi_0$-FAST & 75.3 & 67.5 & 42.9 & 0.0 & 46.4 \\
$\pi_{0.5}$ & -- & -- & -- & -- & \underline{72.7} \\
GR00T-N1.5 & 51.7 & 54.0 & 27.8 & 7.4 & 35.2 \\
GR00T-N1.6 & -- & -- & -- & -- & 67.7 \\
\rowcolor{blue!6}
\textbf{Embodied-R1.5-VLA} & \underline{92.3} & \textbf{93.8} & \textbf{86.1} & \textbf{97.2} & \textbf{92.4} \\
\bottomrule
\end{tabular}
\end{table*}

\begin{table*}[t]
\centering
\footnotesize
\caption{\textbf{SimplerEnv: Google Robot (Variant Aggregation).} Success rate (\%). Results marked with ``$-$'' are sourced from starVLA~\citep{community2026starvla}.}
\label{tab:simpler-va}
\vspace{2pt}
\renewcommand{\arraystretch}{1.1}
\setlength{\tabcolsep}{4pt}
\begin{tabular}{l|cccc|c}
\toprule
\textbf{Method} & \textbf{Pick Coke} & \textbf{Move Near} & \textbf{Open/Close Drawer} & \textbf{Place Apple} & \textbf{Overall} \\
\midrule
RT-1-X & 49.0 & 32.3 & 29.4 & 10.1 & 30.2 \\
RT-2-X & 82.3 & \underline{79.2} & 35.3 & \underline{20.6} & 54.4 \\
OpenVLA & 54.5 & 47.7 & 17.7 & 0.0 & 30.0 \\
SpatialVLA & \underline{88.0} & 72.7 & \underline{41.8} & 6.3 & 52.2 \\
OpenVLA-OFT & 65.3 & 59.0 & 12.2 & -- & 45.5 \\
$\pi_0$ & \textbf{90.1} & \textbf{80.7} & 27.6 & 20.5 & 54.7 \\
$\pi_0$-FAST & 77.6 & 68.2 & 31.3 & 0.0 & 44.3 \\
$\pi_{0.5}$ & -- & -- & -- & -- & \underline{68.4} \\
GR00T-N1.5 & 69.3 & 68.7 & 35.8 & 4.0 & 44.5 \\
GR00T-N1.6 & -- & -- & -- & -- & 65.3 \\
\rowcolor{blue!6}
\textbf{Embodied-R1.5-VLA} & 80.6 & 72.2 & \textbf{58.3} & \textbf{75.0} & \textbf{71.5} \\
\bottomrule
\end{tabular}
\end{table*}

\begin{table*}[t]
\centering
\footnotesize
\caption{\textbf{SimplerEnv: WidowX (Visual Matching).} Success rate (\%).}
\label{tab:simpler-widowx}
\vspace{2pt}
\renewcommand{\arraystretch}{1.1}
\setlength{\tabcolsep}{4pt}
\begin{tabular}{l|cccc|c}
\toprule
\textbf{Method} & \textbf{Spoon on Towel} & \textbf{Carrot on Plate} & \textbf{Stack Blocks} & \textbf{Eggplant in Basket} & \textbf{Overall} \\
\midrule
RT-1-X & 0.0 & 4.2 & 0.0 & 0.0 & 1.1 \\
CogACT & 71.7 & 50.8 & 15.0 & 67.5 & 51.2 \\
OpenVLA & 4.2 & 0.0 & 0.0 & 12.5 & 4.2 \\
SpatialVLA & 16.7 & 25.0 & 29.2 & \textbf{100.0} & 42.7 \\
OpenVLA-OFT & 34.2 & 30.0 & 30.0 & 72.5 & 41.8 \\
$\pi_0$ & 29.1 & 0.0 & 16.6 & 62.5 & 27.1 \\
$\pi_0$-FAST & 29.1 & 21.9 & 10.8 & 66.6 & 32.1 \\
$\pi_{0.5}$ & 49.3 & 64.7 & \underline{44.7} & 69.7 & 57.1 \\
GR00T-N1.5 & \underline{75.3} & 54.3 & \textbf{57.0} & 61.3 & \underline{62.0} \\
GR00T-N1.6 & 64.5 & \underline{65.5} & 5.5 & \underline{93.0} & 57.1 \\
\rowcolor{blue!6}
\textbf{Embodied-R1.5-VLA} & \textbf{83.3} & \textbf{75.0} & 37.5 & \textbf{100.0} & \textbf{74.0} \\
\bottomrule
\end{tabular}
\end{table*}

\begin{table*}[t]
\centering
\footnotesize
\caption{\textbf{LIBERO results.} Success rate (\%). Pt.: whether action pretraining is used.}
\label{tab:libero}
\vspace{2pt}
\renewcommand{\arraystretch}{1.1}
\setlength{\tabcolsep}{4pt}
\begin{tabular}{l|c|cccc|c}
\toprule
\textbf{Method} & \textbf{Pt.} & \textbf{Goal} & \textbf{Spatial} & \textbf{Object} & \textbf{Long} & \textbf{Overall} \\
\midrule
\multicolumn{7}{l}{\cellcolor{gray!8}\textit{With Action Pretraining}} \\
OpenVLA & Y & 79.2 & 84.7 & 88.4 & 53.7 & 76.5 \\
$\pi_0$ & Y & 95.8 & 96.8 & 98.8 & 85.2 & 94.2 \\
$\pi_0$-FAST & Y & 88.6 & 96.4 & 96.8 & 60.2 & 85.5 \\
$\pi_{0.5}$ & Y & \textbf{98.0} & \textbf{98.8} & 98.2 & 92.4 & 96.9 \\
GR00T-N1 & Y & 93.0 & 94.4 & 97.6 & 90.6 & 93.9 \\
GR00T-N1.6 & Y & 97.5 & 97.7 & 98.5 & 94.4 & 97.0 \\
OpenVLA-OFT & Y & 97.9 & 97.6 & 98.4 & \textbf{94.5} & 97.1 \\
\midrule
\multicolumn{7}{l}{\cellcolor{gray!8}\textit{Without Action Pretraining}} \\
Diffusion Policy & N & 68.3 & 78.3 & 92.5 & 50.5 & 72.4 \\
OpenVLA-OFT & N & 91.7 & 94.3 & 95.2 & 86.5 & 91.9 \\
$\pi_0$-FAST & N & 89.0 & 87.0 & 63.0 & 48.0 & 71.8 \\
$\pi_{0.5}$ & N & 94.6 & 96.6 & 97.2 & 85.8 & 93.6 \\
\rowcolor{blue!6}
\textbf{Embodied-R1.5-VLA} & N & 97.6 & 98.3 & \textbf{99.2} & 93.9 & \textbf{97.3} \\
\bottomrule
\end{tabular}
\end{table*}

\begin{table*}[t]
\vspace{-10pt}
\centering
\footnotesize
\caption{\textbf{LIBERO-Plus results.} Success rate (\%) under distribution shifts.}
\label{tab:libero-plus}
\vspace{2pt}
\setlength{\tabcolsep}{3pt}
\renewcommand{\arraystretch}{1.1}
\begin{tabular}{l|ccccccc|c}
\toprule
\textbf{Method} & \textbf{Camera} & \textbf{Robot} & \textbf{Language} & \textbf{Lighting} & \textbf{Background} & \textbf{Noise} & \textbf{Layout} & \textbf{Total} \\
\midrule
OpenVLA & 0.8 & 3.5 & 23.0 & 8.1 & 34.8 & 15.2 & 28.5 & 15.6 \\
$\pi_0$ & 13.8 & 6.0 & 58.8 & 85.0 & 81.4 & 79.0 & 68.9 & 53.6 \\
$\pi_0$-FAST & \textbf{65.1} & 21.6 & 61.0 & 73.2 & 73.2 & 74.4 & 68.8 & 61.6 \\
OpenVLA-OFT & 56.4 & 31.9 & 79.5 & 88.7 & 93.3 & 75.8 & 74.2 & 69.6 \\
\rowcolor{blue!6}
\textbf{Embodied-R1.5-VLA} & 53.2 & \textbf{54.1} & \textbf{88.0} & \textbf{94.9} & \textbf{93.9} & \textbf{80.6} & \textbf{78.4} & \textbf{76.0} \\
\bottomrule
\end{tabular}
\vspace{-10pt}
\end{table*}

\textbf{SimplerEnv.} SimplerEnv~\citep{li2024evaluating} evaluates cross-embodiment generalization across three robot environments (\cref{tab:simpler-vm,tab:simpler-va,tab:simpler-widowx}). Embodied-R1.5-VLA achieves dominant performance across all three settings. On Google Robot Visual Matching (\cref{tab:simpler-vm}), Embodied-R1.5-VLA reaches 92.4\% overall, surpassing the second-best $\pi_{0.5}$ (72.7\%) by over 20pp. Most notably, on the hardest task Open Top Drawer and Place Apple, the model achieves 97.2\% (vs.\ $\pi_0$'s 46.6\%), demonstrating the substantial advantage of internalized planning and spatial reasoning for complex multi-step manipulation. On Google Robot Variant Aggregation (\cref{tab:simpler-va}), the model achieves 71.5\% overall, with significant leads on the harder tasks Open/Close Drawer (58.3\% vs.\ 41.8\%) and Place Apple (75.0\% vs.\ 20.6\%). On WidowX (\cref{tab:simpler-widowx}), the model achieves 74.0\%, exceeding GR00T-N1.5 (62.0\%) by +12pp.

\textbf{LIBERO.} As shown in \cref{tab:libero}, Embodied-R1.5-VLA achieves 97.3\% overall without action pretraining. This is particularly significant because other methods suffer substantial performance drops without pretraining ($\pi_{0.5}$: 96.9\% $\rightarrow$ 93.6\%, OpenVLA-OFT: 97.1\% $\rightarrow$ 91.9\%). In contrast, Embodied-R1.5-VLA matches and even exceeds the best pretrained model (OpenVLA-OFT, 97.1\%) without any action pretraining, demonstrating that internalized embodied reasoning can effectively substitute for large-scale action data.

\textbf{LIBERO-Plus.} LIBERO-Plus~\citep{fei2025libero} (\cref{tab:libero-plus}) evaluates robustness under seven distribution shifts. Embodied-R1.5-VLA achieves 76.0\%, outperforming OpenVLA-OFT (69.6\%) by +6.4pp. The model achieves best performance on 6 out of 7 shift types, with Robot shift (+22.2pp over second-best) and Language shift (+8.5pp) showing the most prominent advantages.

\subsection{Zero-Shot Manipulation Transfer}
\label{sec:exp-realworld}

\begin{table*}[t]
\centering
\footnotesize
\caption{\textbf{ManiSkill-Affordance: Full per-category results.} Success rate on 20 seen (train) and 10 unseen (test) articulated object categories. \textbf{Bold}: best result per category.}
\label{tab:maniskill}
\vspace{2pt}
\renewcommand{\arraystretch}{1.12}
\setlength{\tabcolsep}{1.5mm}
\resizebox{\textwidth}{!}{%
\begin{tabular}{l cccccccccccccccc}
\toprule
& \multicolumn{16}{c}{\textbf{Seen Categories (Train)}} \\
\cmidrule(lr){2-17}
\textbf{Method}
& \includegraphics[width=0.04\linewidth]{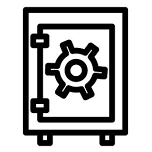}
& \includegraphics[width=0.04\linewidth]{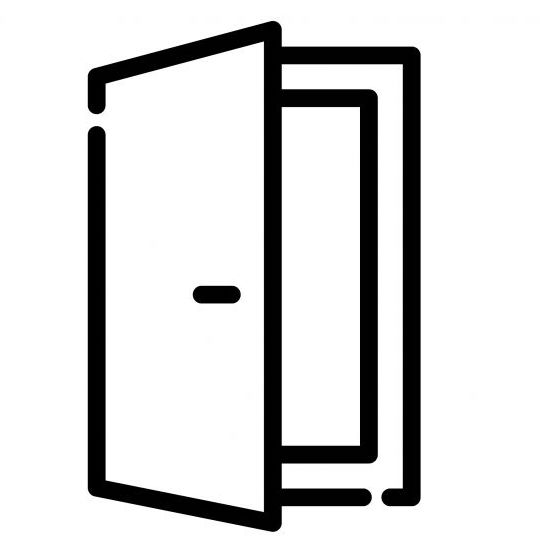}
& \includegraphics[width=0.04\linewidth]{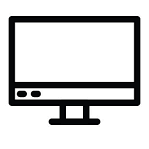}
& \includegraphics[width=0.04\linewidth]{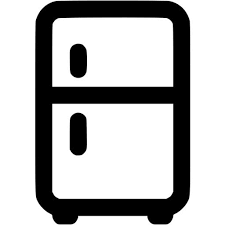}
& \includegraphics[width=0.04\linewidth]{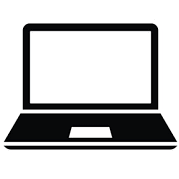}
& \includegraphics[width=0.04\linewidth]{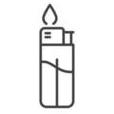}
& \includegraphics[width=0.04\linewidth]{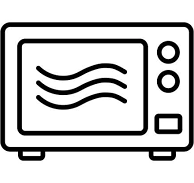}
& \includegraphics[width=0.04\linewidth]{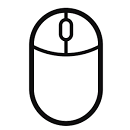}
& \includegraphics[width=0.04\linewidth]{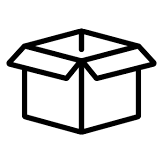}
& \includegraphics[width=0.04\linewidth]{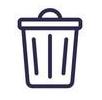}
& \includegraphics[width=0.04\linewidth]{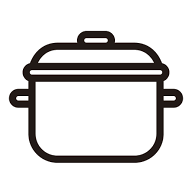}
& \includegraphics[width=0.04\linewidth]{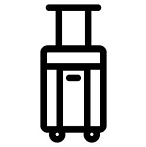}
& \includegraphics[width=0.04\linewidth]{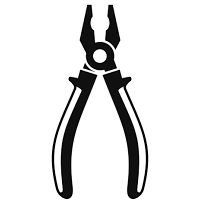}
& \includegraphics[width=0.04\linewidth]{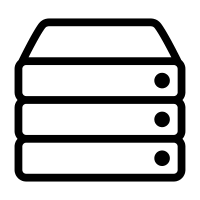}
& \includegraphics[width=0.04\linewidth]{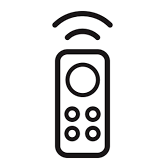}
& \includegraphics[width=0.04\linewidth]{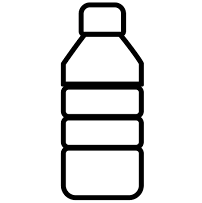} \\
\midrule
Where2Act~\citep{mo2021where2act} & 0.26 & 0.36 & 0.19 & 0.27 & 0.23 & 0.11 & 0.15 & 0.47 & 0.14 & 0.24 & 0.13 & 0.12 & 0.56 & 0.68 & 0.07 & 0.40 \\
Implicit3D~\citep{zhong2023implicit3d} & 0.53 & 0.58 & 0.35 & 0.55 & 0.28 & 0.66 & 0.58 & 0.51 & 0.52 & \textbf{0.57} & 0.45 & 0.34 & 0.41 & 0.54 & 0.39 & 0.43 \\
ManipLLM~\citep{li2024manipllm} & 0.68 & 0.64 & 0.36 & 0.77 & 0.43 & 0.62 & 0.65 & \textbf{0.61} & 0.65 & 0.52 & 0.53 & 0.40 & \textbf{0.64} & 0.71 & \textbf{0.60} & \textbf{0.64} \\
\rowcolor{blue!6}
\textbf{Embodied-R1.5} & \textbf{0.96} & \textbf{0.92} & \textbf{0.92} & \textbf{0.88} & \textbf{0.90} & \textbf{0.80} & \textbf{0.96} & 0.59 & \textbf{0.94} & 0.33 & \textbf{1.00} & \textbf{0.50} & 0.28 & \textbf{1.00} & 0.58 & 0.62 \\
\bottomrule
\end{tabular}%
}
\vspace{4pt}
\resizebox{\textwidth}{!}{%
\begin{tabular}{l cccc c cccccccccc c}
\toprule
& \multicolumn{4}{c}{\textbf{Seen (cont.)}} & & \multicolumn{10}{c}{\textbf{Unseen Categories (Test)}} & \\
\cmidrule(lr){2-5} \cmidrule(lr){7-16}
\textbf{Method}
& \includegraphics[width=0.04\linewidth]{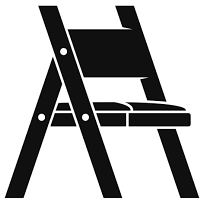}
& \includegraphics[width=0.04\linewidth]{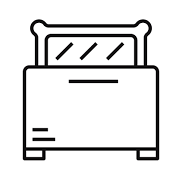}
& \includegraphics[width=0.04\linewidth]{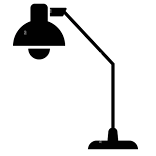}
& \includegraphics[width=0.04\linewidth]{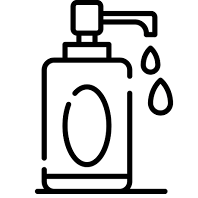}
& \textbf{Avg$_\text{seen}$}
& \includegraphics[width=0.04\linewidth]{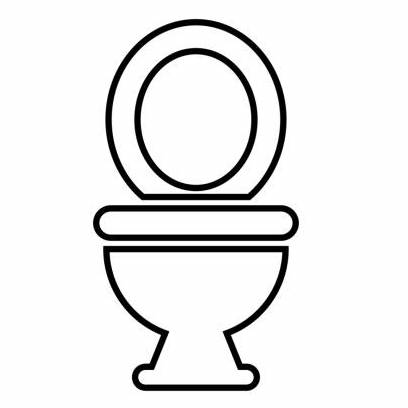}
& \includegraphics[width=0.04\linewidth]{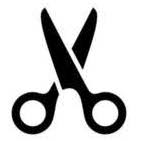}
& \includegraphics[width=0.04\linewidth]{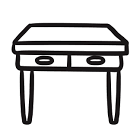}
& \includegraphics[width=0.04\linewidth]{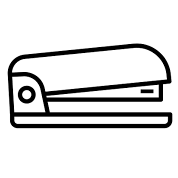}
& \includegraphics[width=0.04\linewidth]{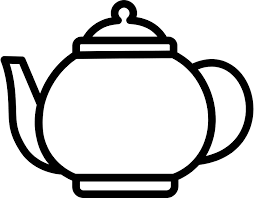}
& \includegraphics[width=0.04\linewidth]{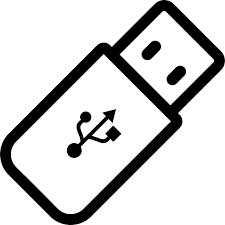}
& \includegraphics[width=0.04\linewidth]{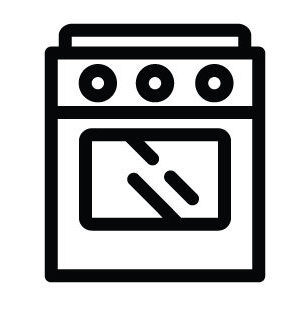}
& \includegraphics[width=0.04\linewidth]{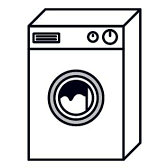}
& \includegraphics[width=0.04\linewidth]{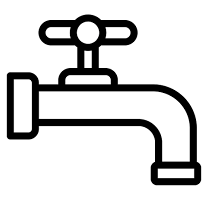}
& \includegraphics[width=0.04\linewidth]{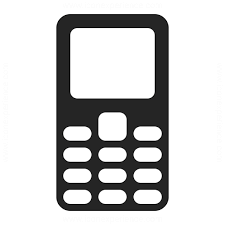}
& \textbf{Avg$_\text{unseen}$} \\
\midrule
Where2Act~\citep{mo2021where2act} & 0.13 & 0.18 & 0.13 & 0.40 & 0.26 & 0.18 & \textbf{0.35} & 0.38 & 0.28 & 0.05 & 0.21 & 0.17 & 0.20 & 0.15 & 0.15 & 0.21 \\
Implicit3D~\citep{zhong2023implicit3d} & 0.27 & 0.65 & 0.20 & 0.33 & 0.46 & 0.45 & 0.17 & 0.80 & 0.53 & 0.15 & 0.69 & 0.41 & 0.31 & 0.30 & 0.31 & 0.41 \\
ManipLLM~\citep{li2024manipllm} & 0.41 & \textbf{0.75} & 0.44 & 0.67 & 0.59 & 0.38 & 0.22 & \textbf{0.81} & \textbf{0.86} & 0.38 & \textbf{0.85} & 0.42 & 0.83 & 0.26 & 0.38 & 0.54 \\
\rowcolor{blue!6}
\textbf{Embodied-R1.5} & \textbf{0.82} & 0.68 & \textbf{0.47} & \textbf{1.00} & \textbf{0.76} & \textbf{0.67} & 0.25 & 0.73 & 0.76 & \textbf{0.94} & 0.22 & \textbf{0.92} & \textbf{0.90} & \textbf{0.69} & \textbf{0.53} & \textbf{0.66} \\
\bottomrule
\end{tabular}%
}
\vspace{-10pt}
\end{table*}

\begin{figure*}[t]
\centering
\includegraphics[width=\textwidth]{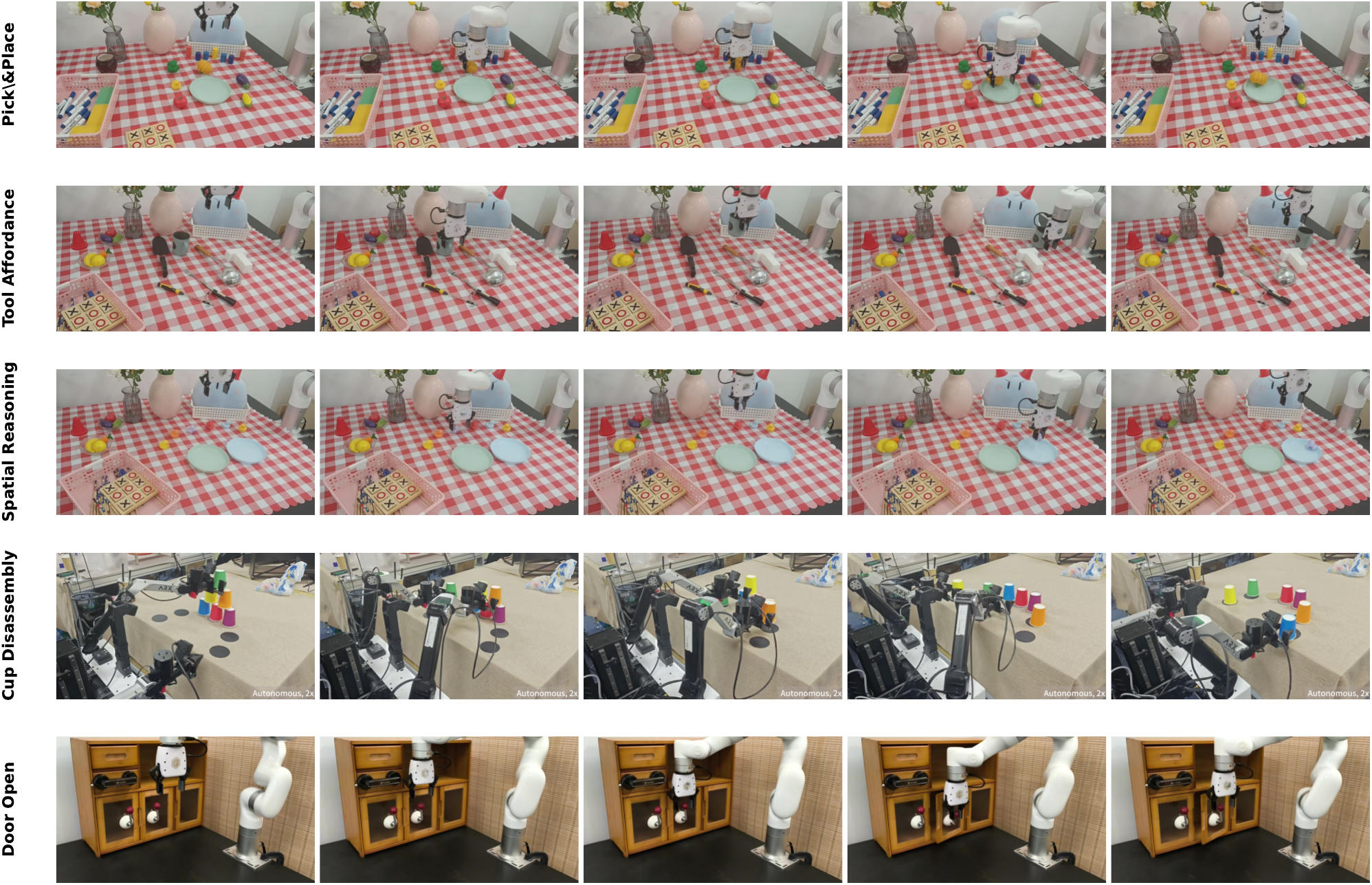}
\caption{\textbf{Zero-shot real-robot manipulation demonstrations.} Keyframe sequences from five task categories, each executed without any task-specific fine-tuning. From top to bottom: Pick\&Place, Tool Affordance, Spatial Reasoning, Cup Disassembly, and Door Open.}
\label{fig:realworld-demo}
\end{figure*}

\begin{table}[t]
\centering
\footnotesize
\caption{\textbf{Zero-shot real-robot manipulation results.} Success rate (\%) across 5 task categories on Xarm6 and ARX Lift2s platform ($n{=}6$ trials per task). Embodied-R1.5 uses the PGC closed-loop framework without any fine-tuning.}
\label{tab:realworld}
\vspace{-5pt}
\setlength{\tabcolsep}{3.5pt}
\renewcommand{\arraystretch}{1.05}
\begin{tabular}{l|ccccc}
\toprule
\textbf{Method} & \textbf{Pick\&Place} & \textbf{Tool Affordance} & \textbf{Spatial Reasoning} & \textbf{Cup Disassembly} & \textbf{Door Open} \\
\midrule
Embodied-R1 & 83.3 & 33.3 & 50.0 & 0.0 & 0.0 \\
RoboBrain 2.0 & \textbf{100.0} & 33.3 & 66.6 & 0.0 & 0.0 \\
\rowcolor{blue!6}
\textbf{Embodied-R1.5} & \textbf{100.0} & \textbf{100.0} & \textbf{83.3} & \textbf{66.6} & \textbf{16.6} \\
\bottomrule
\end{tabular}
\vspace{-10pt}
\end{table}

\textbf{ManiSkill-Affordance.} To evaluate affordance prediction for articulated object manipulation, we benchmark on ManiSkill-Affordance~\citep{mo2021where2act} (\cref{tab:maniskill}), which requires models to predict interaction points on 30 categories of articulated objects covering diverse manipulation primitives (opening, pressing, pulling, rotating, etc.). Following the ManipLLM~\citep{li2024manipllm} evaluation protocol, we use a Franka Panda arm equipped with a floating suction gripper in the SAPIEN~\citep{xiang2020sapien} simulator with PartNet-Mobility assets, and report manipulation success rate. Embodied-R1.5 achieves 0.76 average success rate on seen categories, surpassing the strongest baseline ManipLLM (0.59) by +17pp. On unseen categories that never appear during training, Embodied-R1.5 achieves 0.66 vs.\ ManipLLM's 0.54 (+12pp), demonstrating strong generalization to novel articulated structures. For unseen categories, particularly impressive performance is observed on Kettle (0.94), Oven (0.92), and Washing Machine (0.90), where the model must infer manipulation affordances purely from visual appearance without prior exposure. This validates that the OFG capability and affordance data pipeline enable robust generalization of physical manipulation reasoning beyond seen object categories.

\begin{table}[t]
\centering
\footnotesize
\caption{\textbf{RoboTwin2.0 benchmark results.} Success rate (\%) on 7 manipulation tasks. $\dagger$: fine-tuned with 400 demonstrations per task. Embodied-R1.5 is evaluated in a \textit{completely zero-shot} setting with no RoboTwin data used in any training stage.}
\label{tab:robotwin}
\vspace{2pt}
\renewcommand{\arraystretch}{1.1}
\setlength{\tabcolsep}{3pt}
\begin{tabular}{l|ccccccc|c}
\toprule
\textbf{Method} & \textbf{Click Bell} & \textbf{Click Clock} & \textbf{Move Stapler} & \textbf{Place Mouse} & \textbf{Shake Bottle} & \textbf{Move Card} & \textbf{Place Shoe} & \textbf{Avg} \\
\midrule
\multicolumn{9}{l}{\cellcolor{gray!8}\textit{Fine-tuned with 400 demos per task$\dagger$}} \\
RDT$\dagger$ & 9 & 12 & 0 & 0 & 45 & 11 & 7 & 12.0 \\
$\pi_0$$\dagger$ & 3 & 11 & 2 & 1 & 60 & 22 & 6 & 15.0 \\
$\pi_{0.5}$$\dagger$ & 66 & 89 & 42 & 39 & 97 & 84 & 93 & 72.9 \\
\midrule
\multicolumn{9}{l}{\cellcolor{gray!8}\textit{Zero-shot (no RoboTwin data in training)}} \\
\rowcolor{blue!6}
Embodied-R1.5 & 99 & 56 & 42 & 52 & 89 & 67 & 50 & 65.0 \\
\bottomrule
\end{tabular}
\vspace{-10pt}
\end{table}

\textbf{RoboTwin.} We further evaluate on RoboTwin2.0~\citep{chen2025robotwin} (\cref{tab:robotwin}), selecting 7 diverse manipulation tasks under the challenging \textit{demo\_randomized} configuration. Critically, Embodied-R1.5 is evaluated in a \textit{completely zero-shot} manner: \textbf{no RoboTwin data is used in any training stage}, and the model relies solely on pointing prediction combined with a unified motion logic for execution, similar to the approach in Embodied-R1 (see \cref{fig:robotwin-vis} for qualitative visualization). In contrast, all baselines ($\pi_0$, $\pi_{0.5}$, RDT) are fine-tuned with 400 demonstrations per task. Despite this significant disadvantage, Embodied-R1.5 achieves 65.0\% average success rate, approaching the fine-tuned $\pi_{0.5}$ (72.9\%) and substantially outperforming fine-tuned $\pi_0$ (15.0\%) and RDT (12.0\%). On Click Bell, the model reaches 99\% success rate, surpassing even the fine-tuned $\pi_{0.5}$ (66\%). This result demonstrates that strong affordance understanding and accurate pointing can serve as a powerful zero-shot manipulation primitive, achieving competitive performance with task-specific fine-tuned policies without requiring any in-domain demonstration data.

\textbf{Zero-Shot Real-Robot Experiments.} Following the evaluation protocol of Embodied-R1~\citep{yuan2025embodiedr}, we deploy Embodied-R1.5 on an Xarm6 manipulator and an ARX Lift2s arm using the PGC closed-loop framework, designing five task categories to probe different embodied capabilities (\cref{tab:realworld}). All evaluations require no fine-tuning of any kind, and each task is tested over 6 trials:
\begin{itemize}[leftmargin=*,nosep]
\item \textit{Pick\&Place} (general grasping): the robot is instructed to ``pick up [X] and put it on the plate,'' where [X] can be any object present on the cluttered tabletop. This tests object recognition and stable grasp planning among diverse distractors.
\item \textit{Tool Affordance} (functional part recognition): the robot must ``move [X] to the empty space on the right side of the table,'' where [X] is a tool (screwdriver, hammer, fork, etc.). Success requires grasping the tool by its functional handle rather than the working end, directly testing internalized OFG capability.
\item \textit{Spatial Reasoning} (spatial relation understanding): the robot is instructed to ``put the [X]-th duck toy from the left on the plate,'' where [X] varies across trials. This requires ordinal spatial reasoning over multiple visually similar objects before executing the grasp.
\item \textit{Cup Disassembly} (multi-step task decomposition): the robot must separate stacked cups one by one without toppling the remaining stack. This requires accurate task decomposition into sequential sub-steps with per-step pointing for each cup.
\item \textit{Door Open} (articulated object manipulation): the robot must open a cabinet door, requiring 3D trajectory reasoning to predict the arc-shaped motion path of the door handle and execute a smooth pulling motion.
\end{itemize}
Embodied-R1.5 achieves the best performance across all categories (\cref{fig:realworld-demo}). Most notably, on Cup Disassembly (66.6\% vs.\ 0\% for all baselines), Embodied-R1.5 successfully decomposes the task into sub-steps and provides accurate per-cup pointing, while Door Open (16.6\%) demonstrates emerging capability in 3D trace generation for articulated objects.

\subsection{Long-Horizon Closed-Loop Demonstrations}
\label{sec:exp:closedloop}

\begin{figure}[t]
\centering
\includegraphics[width=\textwidth]{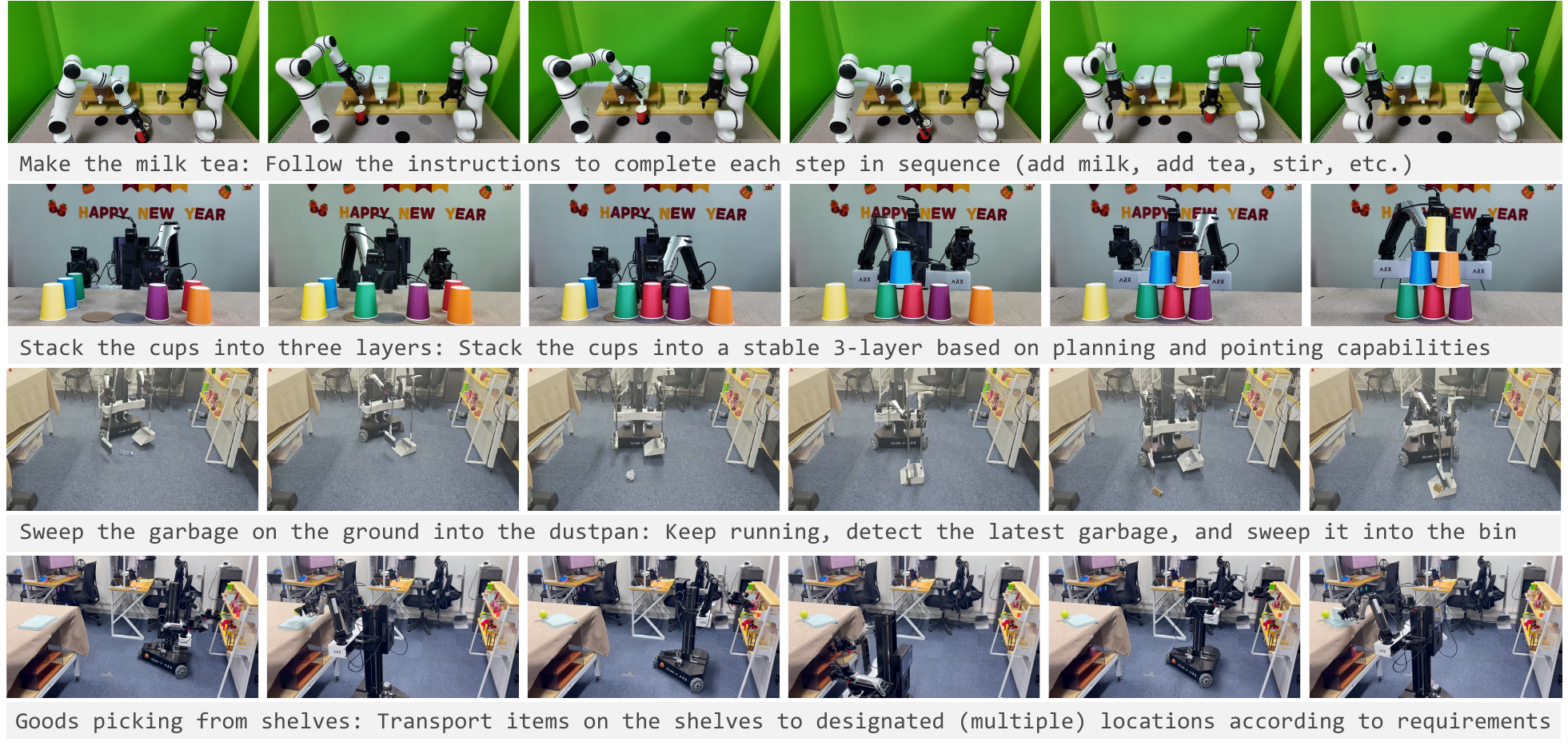}
\caption{\textbf{Qualitative demonstrations of the PGC closed-loop framework.} Keyframes from four representative long-horizon tasks (making milk tea, stacking cups, sweeping garbage, goods picking from shelves), each requiring 5--10 sequential sub-steps with autonomous error detection and recovery.}
\label{fig:pgc-demo}
\end{figure}

\begin{figure*}[t]
\centering
\includegraphics[width=\textwidth]{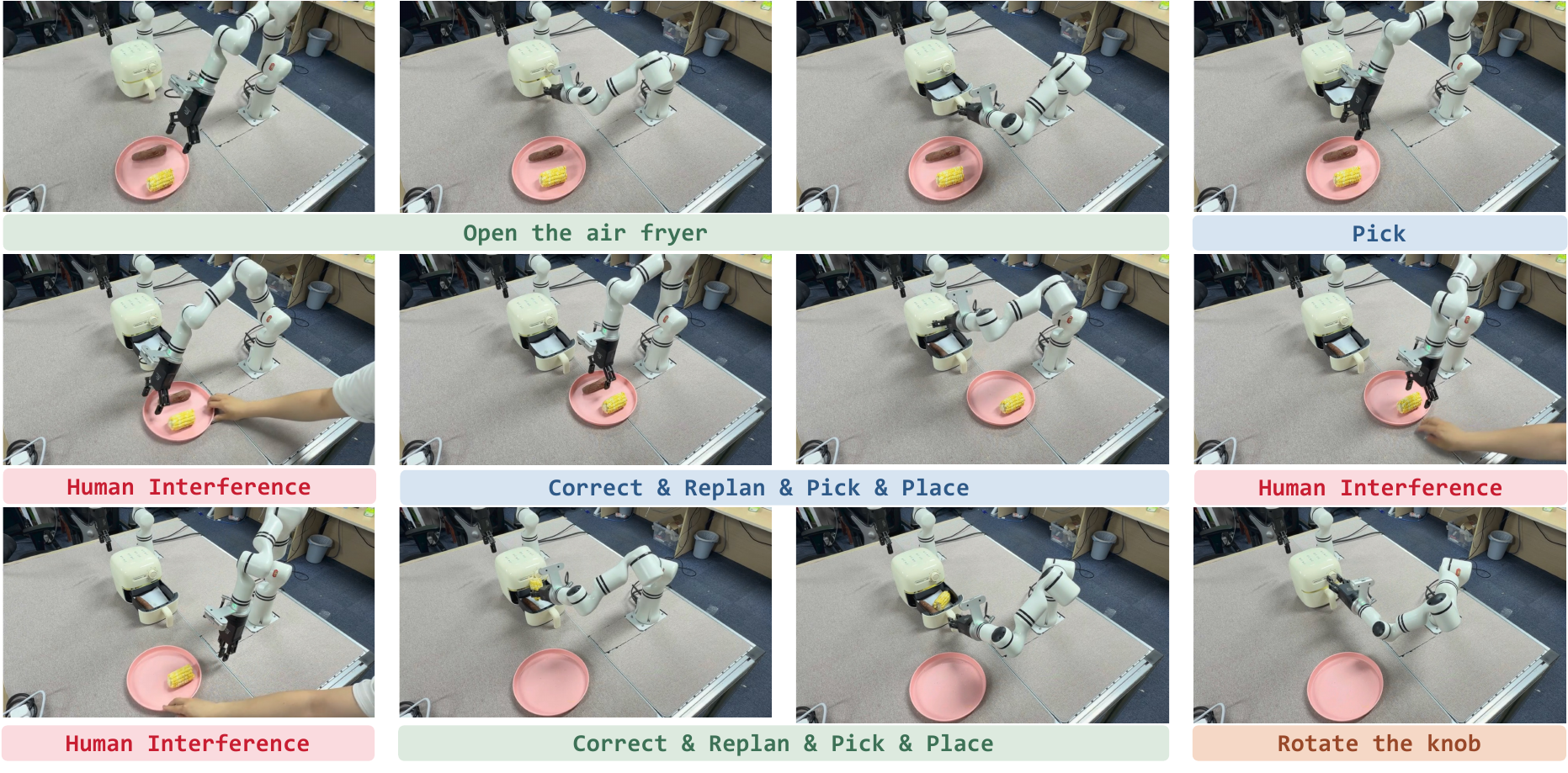}
\caption{\textbf{Robust correction under human perturbation.} The robot executes a pick-and-place task while a human operator actively disturbs the scene. Row 1: normal plan and grasp execution. Row 2: human moves the plate away; the Corrector detects the deviation and triggers re-localization. Row 3: human perturbs again; the system re-corrects and successfully completes the task.}
\label{fig:correction-demo}
\end{figure*}

We deploy the Planner-Grounder-Corrector (PGC) closed-loop framework on the bimanual ARX Lift2s and Realman RM75 platforms, designing four representative long-horizon tasks (\cref{fig:pgc-demo}). All demonstrations are performed zero-shot without any task-specific fine-tuning, with Embodied-R1.5 simultaneously serving as Planner, Grounder, and Corrector.

\textbf{Make milk tea} (10 steps): The robot receives a milk-tea instruction together with a detailed recipe. The Planner decomposes it into ten sequential steps (fetch cup, open lid, pour tea bag, add hot water, stir, etc.); the Grounder localizes required tools and containers at each step; the Corrector monitors execution and triggers re-execution upon detecting deviations (e.g., grasp slippage, misplacement). This scenario emphasizes instruction-following over extremely long horizons and the critical role of closed-loop correction under accumulated errors.

\textbf{Stack cups into three layers} (6 steps): Six color-distinct cups are stacked into a stable three-layer structure. The Planner decomposes this into six sequential pick-and-place sub-goals; the Grounder localizes target cups and placement positions; the Corrector verifies alignment after each placement, triggering re-grasp when cups tilt. This stresses precise spatial pointing and bimanual coordination.

\textbf{Sweep garbage} (cyclic): The robot continuously detects newly appeared garbage and sweeps it into a dustpan. The Planner maintains a cyclic detect-localize-sweep strategy; the Grounder rescans the scene each cycle to localize latest garbage; the Corrector triggers supplementary sweeps for missed items.

\textbf{Goods picking from shelves} (open-vocabulary): The robot picks user-specified items from a shelf and transports them to designated locations. Critically, all shelf items are absent from any training data and can be freely replaced with different object categories and arbitrary arrangements, testing completely zero-shot open-vocabulary recognition and localization. The Corrector verifies each pick-and-place and triggers re-execution upon failure.

\textbf{Robust correction under human perturbation.} We further demonstrate the correction capability of PGC under active human disturbances (\cref{fig:correction-demo}). During a ``pick up the corn and place it on the plate'' task, a human operator repeatedly intervenes by relocating the plate and displacing the target object mid-execution. After each perturbation, the Corrector detects the deviation between the current state and the expected plan in real time, triggering the Grounder to re-localize both the target object and the placement destination before resuming execution. Despite multiple consecutive disturbances, the system autonomously recovers and successfully completes the task without any manual reset. This demonstrates the critical role of closed-loop correction in scenarios where open-loop systems would inevitably fail.

All tasks achieve fully autonomous execution with no human intervention at any stage. Full video visualizations are available on our project page.

\subsection{Analysis}
\label{sec:exp:ablation}

We design a series of controlled experiments to validate key design choices, organized as research questions.

\begin{figure}[t]
\centering
\begin{minipage}[t]{0.48\textwidth}
\centering
\includegraphics[width=\textwidth,height=4.5cm,keepaspectratio]{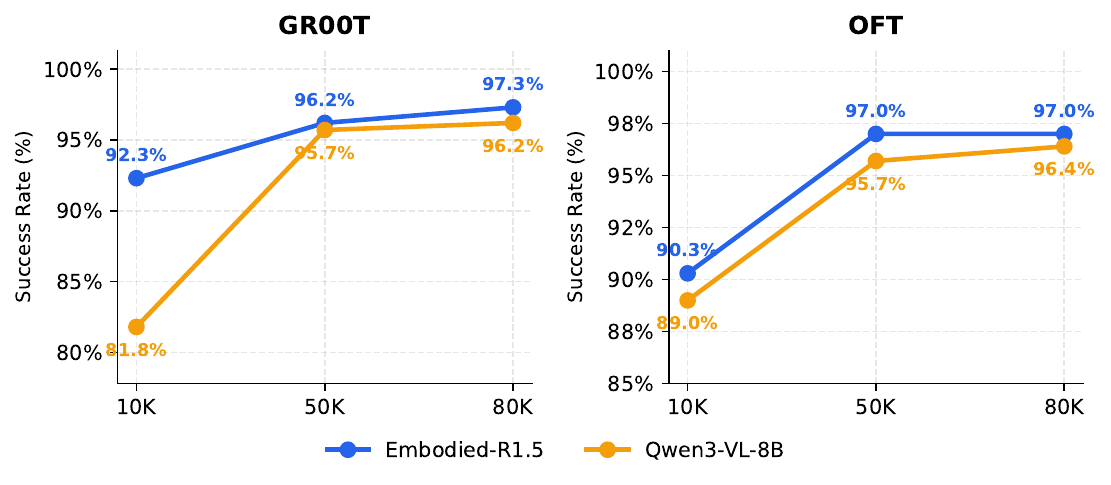}
\caption{\textbf{VLM backbone comparison on LIBERO.} Overall success rate (\%) of Embodied-R1.5 vs.\ Qwen3-VL-8B with two action experts at different training steps.}
\label{fig:ablation-backbone}
\end{minipage}%
\hfill
\begin{minipage}[t]{0.48\textwidth}
\centering
\includegraphics[width=\textwidth]{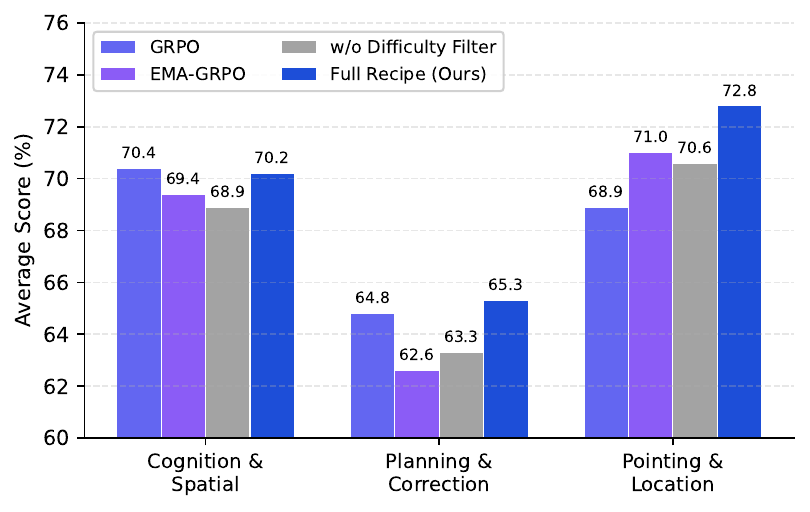}
\caption{\textbf{RFT algorithm ablation.} Average score across three capability dimensions for different RL configurations.}
\label{fig:ablation-rl}
\end{minipage}
\end{figure}

\textbf{Q1: Does RFT improve over pure SFT?}

\begin{wraptable}{r}{0.48\textwidth}
\vspace{-12pt}
\centering
\footnotesize
\caption{\textbf{SFT vs.\ RFT comparison.} Average score per capability dimension.}
\label{tab:sft-vs-rft}
\vspace{2pt}
\renewcommand{\arraystretch}{1.1}
\setlength{\tabcolsep}{4pt}
\begin{tabular}{l|ccc}
\toprule
\textbf{Model} & \textbf{Planning} & \textbf{Pointing} & \textbf{Cognition} \\
\midrule
Embodied-R1.5-SFT & 62.6 & 69.0 & 68.9 \\
\rowcolor{blue!6}
\textbf{Embodied-R1.5} & \textbf{65.3} & \textbf{72.8} & \textbf{70.2} \\
\midrule
$\Delta$ (RFT gain) & +2.7 & +3.8 & +1.3 \\
\bottomrule
\end{tabular}
\vspace{-10pt}
\end{wraptable}

We compare Embodied-R1.5 (SFT+RFT) with an SFT-only variant trained on the same data (\cref{tab:sft-vs-rft}). RFT yields consistent gains across all three capability dimensions: +3.8pp on Pointing, +2.7pp on Planning, and +1.3pp on Cognition. The largest improvement on Pointing aligns with our design expectation: point localization and trajectory prediction tasks admit verifiable geometric rewards (distance to ground-truth coordinates), which provide precise gradient signals that RL optimization can exploit most effectively. In contrast, Planning and Cognition tasks rely on LLM-as-judge rewards with inherently noisier supervision, yet still benefit meaningfully from the RFT stage. This confirms that reinforced fine-tuning is a necessary component that extracts additional performance beyond what supervised learning alone achieves.

\textbf{Q2: Does Embodied-R1.5 serve as a better VLM backbone for VLA?}
A central hypothesis of our work is that a stronger embodied VLM backbone should improve downstream VLA performance even without large-scale action pretraining. We test this by comparing Embodied-R1.5 and the base model Qwen3-VL-8B as backbones, paired with two representative action experts (GR00T and OpenVLA-OFT), on LIBERO across training (\cref{fig:ablation-backbone}). Key findings: (1)~Embodied-R1.5 provides dramatically faster convergence at early training (10K steps: 92.0\% vs.\ 82.0\% with GR00T, +10.0pp), indicating that internalized embodied knowledge substantially reduces the data and compute needed for VLA adaptation. (2)~The advantage persists at convergence (80K steps: 97.0\% vs.\ 96.0\%), confirming genuine capability transfer rather than merely a warm-start effect. (3)~The benefit is consistent across both action experts, ruling out architecture-specific confounds. These results validate our core design premise: investing in a strong embodied foundation model pays dividends downstream, enabling competitive VLA performance without expensive action pretraining.

\textbf{Q3: Which RL recipe works best?}
We ablate four RFT configurations (\cref{fig:ablation-rl}): (a)~vanilla GRPO, (b)~EMA-GRPO~\citep{feng2025onethinker}, (c)~without difficulty-aware data filtering, and (d)~the full recipe (difficulty-aware data filtering + dynamic filtering + global batch reward normalization). The full recipe achieves the best overall balance across all three dimensions (Cognition 70.2 / Planning 65.3 / Pointing 72.8). Specifically, difficulty-aware filtering contributes the most consistent gains by preventing the RL optimizer from wasting gradient updates on trivially easy or impossibly hard samples. Without filtering, the optimizer tends to over-optimize on already-solved tasks while under-training on challenging ones. EMA-GRPO shows marginal improvement over vanilla GRPO but still lags behind the full recipe, suggesting that global batch reward normalization is more effective than per-task moving averages for embodied reward distributions with heterogeneous task volumes.

\begin{figure}[t]
\centering
\includegraphics[width=\textwidth]{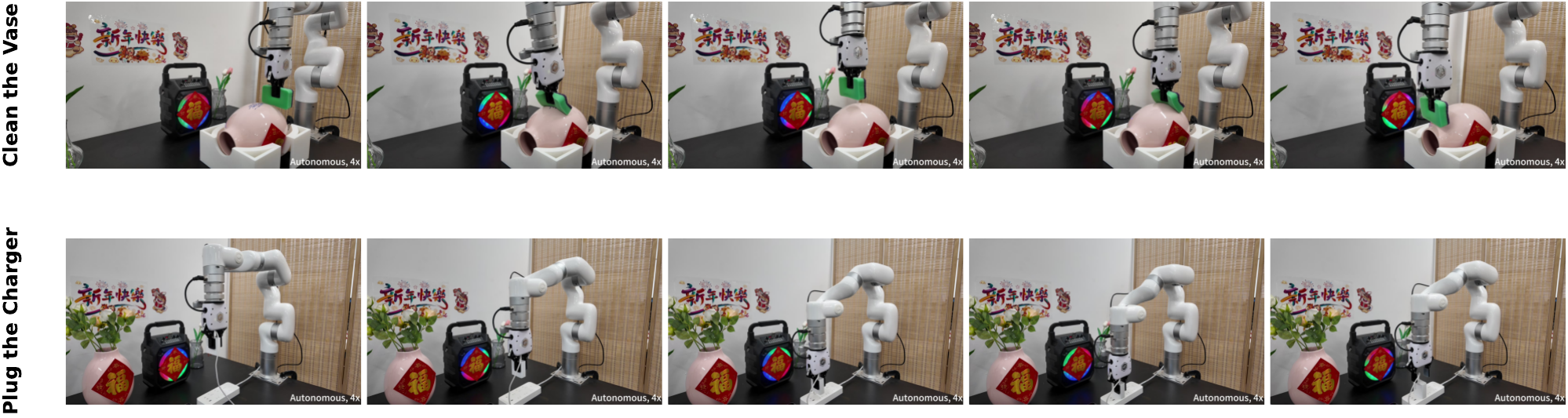}
\caption{\textbf{Contact-rich manipulation via vision-to-force handover.} Embodied-R1.5 provides visual generalization during approach; the force-aware policy achieves force generalization during contact. Top: cleaning a vase with appropriate contact pressure. Bottom: plugging a charger with sub-millimeter alignment precision.}
\label{fig:forceflow-demo}
\end{figure}

\textbf{Q4: Can Embodied-R1.5 enable contact-rich manipulation via force-aware policy?}
Standard VLA policies rely solely on visual feedback, which is insufficient for contact-rich tasks where interaction forces must be precisely regulated. We combine Embodied-R1.5 with a force-aware flow matching policy to form a vision-to-force handover mechanism: during the approach phase, Embodied-R1.5 provides semantic understanding and spatial localization to achieve \textit{visual generalization} across diverse objects and scenes; upon contact, the force-aware policy takes over and regulates interaction dynamics through force/torque feedback to achieve \textit{force generalization} across varying contact conditions. We validate on two representative tasks (\cref{fig:forceflow-demo}): \textit{Clean the Vase} requires maintaining appropriate pressure while conforming to curved geometry, and \textit{Plug the Charger} demands sub-millimeter alignment under tight contact constraints. Both tasks are executed fully autonomously, demonstrating that Embodied-R1.5's embodied reasoning naturally composes with force-aware policies to unlock contact-rich manipulation beyond what either modality alone can achieve. The complete technical report is available in ForceFlow~\citep{zhang2026forceflow}.

\section{Related Work}
\label{sec:related}

The pursuit of general physical intelligence has driven rapid development of Embodied Foundation Models (EFMs)~\cite{survey2025embodied}. Early efforts focus on individual capability dimensions, e.g., spatial reasoning~\cite{chen2024spatialvlm,ouyang2025spacer}, physical scene understanding~\cite{luo2025visual,azzolini2025cosmos}, planning~\cite{sermanet2024robovqa,mu2024embodiedgpt}, and embodied grounding~\cite{yuan2024robopoint,roboafford2025,yuan2025seeing,yuan2025embodiedr}. Recent works attempt broader unification: RoboBrain~\cite{team2025robobrain,tan2026robobrain} integrates spatial and temporal reasoning via chain-of-thought; MIMO-Embodied~\cite{hao2025mimo} and ACE-Brain-0~\cite{gong2026ace} fuse capabilities across driving and manipulation; Pelican-VL~\cite{zhang2025pelican} and RynnBrain~\cite{dang2026rynnbrain} target unified embodied reasoning with grounded pointing; and HY-Embodied~\cite{team2026hy} proposes iterative self-evolution. Embodied-R1.5 further unifies all three dimensions within a single model and proposes a multi-task balanced RL recipe to resolve heterogeneous training interference. EFMs can also serve as the foundation for VLA training. $\pi_{0.5}$~\cite{intelligence2025visionlanguageaction} co-trains with embodied reasoning data to mitigate forgetting, Gemini-Robotics-ER-1.5~\cite{team2025gemini} combines an embodied reasoning model with a VLA, and MolmoAct2~\cite{fang2025molmoact2} injects adaptive reasoning into action generation. Embodied-R1.5-VLA shows that a strong embodied backbone substantially reduces the data requirement for action learning.

\section{Conclusion}
\label{sec:conclusion}

We presented Embodied-R1.5, a unified Embodied Foundation Model that systematizes embodied reasoning into three core capability dimensions within a single 8B-parameter architecture. Around this unified capability system, we formalize the capability requirements of an EFM and integrate all three dimensions within a shared Transformer, replacing the fragmented multi-model paradigm characteristic of prior work; we further deliver a complete EFM recipe, including a 15B-token data corpus expanded by three automated data construction pipelines and a multi-task balanced RL recipe; we construct the Planner-Grounder-Corrector closed-loop execution framework, where a single model orchestrates the full autonomy stack and validates that internalized reasoning capabilities translate into long-horizon robotic autonomy with self-correction; and we fully open-source the model weights, training data, training code, and EmbodiedEvalKit evaluation framework as community infrastructure. Across 24 Embodied VLM benchmarks, Embodied-R1.5 attains SOTA on 16 of them with an average score of 70.4\% on the 21 main accuracy-based benchmarks, surpassing Gemini-Robotics-ER-1.5 and GPT-5.4 by 17.0\% and 21.7\% respectively; with only a small amount of action data, it can be adapted into Embodied-R1.5-VLA that consistently outperforms strong baselines such as $\pi_{0.5}$ across 4 popular manipulation benchmark suites (e.g., 92.4\% on SimplerEnv Google Robot Visual Matching); zero-shot real-robot experiments further cover instruction following, affordance grounding, articulated object manipulation, and long-horizon closed-loop execution, collectively corroborating the central thesis of this work that the internalization of embodied reasoning can partially substitute for large-scale action pretraining in the evaluated settings.

Despite these strong results, several directions remain open: the current model operates on 2D images, and incorporating native 3D perception such as point clouds and depth maps could further strengthen spatial reasoning in cluttered and occluded scenes; the VLA extension currently uses a lightweight action head, and exploring tighter coupling between reasoning tokens and action generation is a promising avenue; the PGC closed-loop framework has been validated on tabletop manipulation, and extending it to mobile manipulation and navigation with longer horizons and richer environments is an important next step; we hope that the open-source ecosystem released with this work will accelerate community progress on these fronts and bring general-purpose physical intelligence one step closer to practical deployment.

\section*{Contributions}

\textbf{Yifu Yuan} proposed the methodology and research direction, developed the complete training dataset, and was responsible for the entire pipeline, including algorithm implementation, model training, inference, and experimental analysis. Yifu Yuan also led the design of the EmbodiedEvalKit framework for evaluation and the drafting of the manuscript. \textbf{Yifu Yuan}, \textbf{Hongyao Tang}, and \textbf{Yi Ma} served as co-project leads~(\faGem), overseeing the overall execution of the project. The corresponding authors~(\faEnvelope) are \textbf{Shuyang Gu}, \textbf{Yi Ma}, \textbf{Hongyao Tang}, and \textbf{Jianye Hao}.

The success of the Embodied-R1.5 project is a collective effort of all contributors. \textbf{Yaoting Huang}, \textbf{Linqi Han}, and \textbf{Jiangeng Sun} contributed to model evaluation. Data construction and cleaning were performed by \textbf{Yaoting Huang}, \textbf{Linqi Han}, \textbf{Pengyi Li}, \textbf{Jiangeng Sun}, \textbf{Wenting Jia}, \textbf{Yucheng Hu}, \textbf{Zhao Zhang}, and \textbf{Yuxiao Li}. The real-world robotic platform setup and experiments were conducted by \textbf{Shuoheng Zhang}, \textbf{Xianze Yao}, \textbf{Pengyi Li}, \textbf{Yuhao Liu}, \textbf{Yutong Li}, \textbf{Ruihao Liao}, \textbf{Qiyu Wu}, and \textbf{Yuxiao Li}. Research guidance and supervision were provided by \textbf{Shuyang Gu}, \textbf{Zibin Dong}, \textbf{Fei Ni}, \textbf{Yan Zheng}, \textbf{Han Hu}, and \textbf{Jianye Hao}.

\bibliography{references}

\newpage
\appendix
\addtocontents{toc}{\protect\setcounter{tocdepth}{1}}
\section{Data Composition Details}
\label{app:data-details}

This appendix provides the full data composition details described in Section~\ref{sec:data}.

\subsection{Embodied Cognition \& Spatial Reasoning Data}

\textbf{Multi-view spatiotemporal reasoning.} Spatial reasoning and scene cognition are foundational for embodied VLMs to perceive the physical environment. Since inputs in embodied scenarios are often temporal and multi-view, we integrate VLM-3R~\citep{vlm3r}, Cambrian-S~\citep{cambrians}, and SAT~\citep{ray2024sat} datasets to strengthen the model's spatiotemporal reasoning under diverse viewpoints. For video inputs, we additionally incorporate multiple video spatial reasoning datasets~\citep{ouyang2025spacer,feng2025onethinker}. These datasets collectively cover object counting, relative distance, relative direction, spatial topological relations (above/inside/below/between, etc.), depth ordering, occlusion judgment, appearance order, object size, and absolute distance. They require the model to jointly reason about spatial information from different observation angles, establishing a foundation for 3D perception in embodied scenes.

\textbf{Depth estimation.} Depth perception is a critical channel for embodied VLMs to understand the 3D structure of the physical environment. We adopt indoor scene depth estimation data from DepthLM~\citep{cai2025depthlm}, which extracts sensor-level ground-truth depth values from high-quality indoor 3D datasets including ScanNet++~\citep{yeshwanth2023scannet}, Taskonomy~\citep{zamir2018taskonomy}, HM3D~\citep{ramakrishnan2021habitat}, and Matterport3D~\citep{chang2017matterportd}, constructing depth QA pairs based on specified image coordinates. The data encompasses both absolute metric depth and relative depth: the former provides precise distance values, while the latter enhances the model's understanding of scene depth hierarchies. During data construction, the point sampling strategy explicitly excludes pixels at object boundaries, at infinity, or in physically inconsistent regions, and normalizes camera focal lengths across all images for cross-source scale unification. We deliberately select only high-quality indoor datasets, as low-quality and synthetic data provide no positive contribution to VLM training without thorough cleaning.

\textbf{Embodied scene cognition.} To adapt the model to the distinctive viewpoints and semantic requirements of robotic manipulation scenarios, we curate approximately 106K samples from Robo2VLM~\citep{wang2025robo2vlm} and RoboBrain 2.0~\citep{team2025robobrain}, covering scene understanding question-answering from the robot's first-person perspective. This data helps the model establish environment cognition from the robot's manipulation viewpoint.

\textbf{Tabletop spatial reasoning.} To address the gap in fine-grained tabletop spatial reasoning, we construct the \textit{ER1.5-Spatial} dataset ($\sim$20K) from Fractal~\citep{brohan2022rt}, BridgeData~V2~\citep{walke2023bridgedata}, and DROID~\citep{khazatsky2024droid} datasets via a fully automated 3D scene annotation pipeline. The pipeline takes a single RGB image as input and produces a structured 3D semantic scene graph, from which spatial reasoning QA pairs are programmatically generated covering spatial relations, distance metrics, scene cognition, and appearance order. Full pipeline implementation details are provided in Appendix~\ref{app:spatial-pipeline}.

\subsection{Embodied Planning Data}

Embodied planning requires the model to assess the current execution state, comprehend the target task objective, and perform task decomposition and next-step planning in long-horizon tasks. We compile planning data from multiple sources, covering diverse planning demands across manipulation, navigation, and egocentric scenarios. We first integrate approximately 950K task planning QA pairs from open-source datasets including RoboVQA~\citep{sermanet2024robovqa}, EO-Data~\citep{qu2025eo}, AgiBot-World~\citep{bu2025agibot}, and Cosmos-Reason~\citep{azzolini2025cosmos}, encompassing long-horizon task planning capabilities in complex embodied scenarios and forming the large-scale robot visual QA planning data. In addition, we incorporate diverse samples from EgoPlan-IT~\citep{chen2024egoplan} and EgoRe~\citep{pei2026egothinker}, which are extracted from first-person videos and require the model to predict subsequent action sequences based on observed manipulation progress.

\subsection{Embodied Correction Data}

Error correction is critical for closed-loop autonomous execution. Existing robotic datasets predominantly contain successful demonstrations, while failure data with structured error annotations remains severely insufficient. We first incorporate 78K video question-answering pairs from the RoboFAC~\citep{ye2025robofac} dataset, which covers fault understanding and correction across different robots. To address comprehensive capability requirements, we draw upon the failure taxonomy established in prior work~\citep{ye2025robofac,pacaud2025guardian,liu2023reflect} and construct the \textit{ER1.5-Correction} dataset, a large-scale failure correction QA dataset covering the complete pipeline from task planning to execution. Through an automated perturbation generation pipeline that enables large-scale expansion, the dataset comprises 800K samples. The dataset is organized along two orthogonal dimensions: by \textit{capability level}, it spans failure detection (binary classification), failure localization (identifying the specific failure type and location), and failure correction (generating recovery plans from erroneous states); by \textit{stage}, it covers planning failures (step omission, redundancy, swap, object error) and execution failures (execution interruption, wrong manipulation target, incorrect action). The combination of these two dimensions yields six QA task types. Data sources encompass diverse real-robot datasets (BridgeData~V2~\citep{walke2023bridgedata}, RoboFail~\citep{liu2023reflect}) and simulation datasets (ManiSkill~\citep{tao2024maniskill3}, GEMBench~\citep{garcia2024gembench}).

Specifically, for planning failures, we start from correct sub-task plans and apply five structured perturbation operators to automatically generate erroneous plans. For each perturbed plan, we simultaneously generate QA pairs at multiple capability levels, while retaining the correct plan as a positive sample in the training data to prevent overfitting. For execution failures, we adopt three complementary strategies: truncating successful demonstration videos to simulate execution interruption, replacing sub-task descriptions to simulate object/action errors, and injecting physical perturbations in ManiSkill~\citep{tao2024maniskill3} simulation to simulate manipulation failures. For real failure records already present in RoboFail~\citep{liu2023reflect}, we directly extract and reformat them. All data types undergo human sampling verification, and the complete construction pipeline and templates are provided in Appendix~\ref{app:correction-pipeline}.

\subsection{Embodied Pointing Data}

Pointing is the signature capability of Embodied-R1.5. We organize pointing data according to the four pointing capabilities defined in Section~\ref{sec:cap:pointing}. Beyond extensively integrating existing open-source data, we also significantly expand data through automated pipelines in the directions of functional affordance and trajectory generation. 

\textbf{REG (Referring Expression Grounding).}
Referring expression grounding requires the model to precisely localize target objects based on natural language descriptions. We construct massive training data from multiple sources including RefCoco~\citep{kazemzadeh2014referitgame}, SAM2~\citep{ravi2025sam}, Pixmo-Points~\citep{deitke2025molmo}, CoSyn-Point~\citep{clark2026molmo2}, LVIS~\citep{gupta2019lvis}, RoboRefit~\citep{roborefit2023}, and Ref-L4~\citep{chen2025revisiting}.

\textbf{RRG (Region Referring \& Grounding).}
Region grounding is a core capability for object placement and spatial arrangement in embodied manipulation, requiring the model to identify suitable free regions for placement and assess placement feasibility. We integrate RoboPoint~\citep{yuan2024robopoint} covering diverse robotic manipulation pointing scenarios, RefSpatial~\citep{refspatial2025} providing region-level spatial referring and grounding, and ER1-Point inherited and extended from Embodied-R1. Additionally, we extract large-scale region grounding data from massive human interaction videos, providing diverse real-world placement scenarios. Notably, in Embodied-R1.5 we generate Regular Rearrangement samples via synthetic simulation data (RegularRearrangement). These tasks present scenes with blocks arranged in regular patterns (triangles, squares, etc.) with 1--2 blocks removed; the model must reason about where to point to complete the pattern, combining spatial reasoning with pointing capability.

\textbf{OFG (Object Functional Grounding).}
Functional grounding integrates visual localization with physical property understanding, requiring the model to localize functional parts of objects (e.g., cup handles, kettle spouts, tool grips), which is critical for correct grasping strategies in embodied manipulation. Since part-level annotation in the real world is extremely expensive, we construct OFG data from three directions: (1)~open-source affordance datasets: HandAL~\citep{guo2023handaldatasetrealworldmanipulable} provides human hand-functional part interaction annotations, PACO-LVIS~\citep{ramanathan2023paco} provides part-level affordance annotations, and InstructPart~\citep{wan2025instructpart} provides instruction-conditioned part grounding; (2)~simulation-based automatic extraction: PRISM~\citep{deshpande2025graspmolmo} renders diverse articulated objects and automatically generates functional grounding annotations, and PartNet-Maniskill extracts part-level data covering 17 categories of articulated objects from ManiSkill~\citep{tao2024maniskill3}; (3)~human interaction data: we extract human grasping patterns on functional parts from large-scale hand-object interaction datasets as supervision signals. See Appendix~\ref{app:affordance-trajectory-pipeline} for pipeline details.

\textbf{VTG (Visual Trace Generation).}
Trajectory prediction is crucial for internal planning in embodied tasks, depicting expected motion paths of objects or end-effectors as 2D or 3D feasibility curves. We automatically extract trajectory annotations from large-scale self-constructed simulation data and human interaction data, covering two trajectory types: robot-centric traces (describing end-effector motion paths) and object-centric traces (describing the motion trajectories of manipulated objects). See Appendix~\ref{app:affordance-trajectory-pipeline} for pipeline details.

\subsection{General Knowledge Data}

To maintain general vision-language capability and prevent catastrophic forgetting, we include general knowledge and reasoning samples from LLaVA~\citep{li2024llava}, HONEY~\citep{zhang2025bee}, MM-IF~\citep{zhao2025mmifengine}, and EUCLID~\citep{lian2025euclid}, covering diverse VQA, captioning, reasoning, and multi-modal instruction following tasks.


\section{Automatic Data Construction Pipeline}
\label{app:data-pipelines}

This appendix provides full details of the three automated data construction pipelines described in Section~\ref{sec:data}. Section~\ref{app:spatial-pipeline} describes the 3D scene annotation pipeline for spatial reasoning data (ER1.5-Spatial). Section~\ref{app:correction-pipeline} describes the failure-aware data construction pipeline for correction data (ER1.5-Correction). Section~\ref{app:affordance-trajectory-pipeline} describes the functional affordance and trajectory data construction pipelines.

\subsection{Pipeline 1: 3D Scene Annotation for Spatial Reasoning Data}
\label{app:spatial-pipeline}

This section details the automated 3D scene annotation pipeline used to construct the ER1.5-Spatial dataset ($\sim$20K samples). The pipeline takes single RGB images as input and produces structured 3D semantic scene graphs, from which tabletop-level spatial reasoning QA pairs are programmatically generated.

While several open-source 3D datasets provide geometric, semantic, and instance-level metadata --- including ScanNet~\citep{dai2017scannet}, ScanNet++~\citep{yeshwanth2023scannet}, and ARKitScenes~\citep{dehghan2021arkitscenes} --- these datasets predominantly operate at room-level granularity typically used for navigation scenarios, which is complementary to our data. Our goal is to construct spatial reasoning QA data at a finer \textit{tabletop manipulation} granularity: given an RGB image of a tabletop operation scene, automatically infer the complete 3D scene information (object categories, spatial positions, inter-object relations) and generate spatial reasoning QA training data. Early efforts create small-scale spatial QA datasets via manual or semi-automated annotation, an approach that does not scale. Our key insight is that once a detailed 3D semantic scene graph can be automatically reconstructed from a single RGB image, large-scale spatial reasoning QA data can be generated programmatically.

\paragraph{Data source.} Our input images are drawn from large-scale real-robot datasets in Open X-Embodiment~\citep{openxembodiment2023}, including BridgeData V2~\citep{walke2023bridgedata} and DROID~\citep{khazatsky2024droid}, which provide diverse tabletop operation RGB images spanning varied object compositions, table layouts, and lighting conditions. We sample representative frames from these datasets as pipeline input. Let the input image set be $\mathcal{I} = \{I_i\}_{i=1}^N$. The pipeline processes each image independently and outputs a structured annotation record $\mathcal{S}_i$ containing: semantic instance labels $\{c_k\}$, 2D instance masks $\{M_k\}$, a world-frame point cloud $\mathbf{P}_i^w$, per-instance 3D bounding boxes $\{B_k\}$, and the camera-to-world transformation $\mathbf{T}_i$. The pipeline proceeds through multiple stages: semantic understanding, geometry estimation with 2D segmentation, 3D lifting, and coordinate normalization. Quality control mechanisms are embedded within each stage.

\paragraph{Semantic understanding.} Before geometric processing, the pipeline extracts object category labels for each image. The input image is first resized so that both height and width are multiples of 14, satisfying the patch-based tokenizer constraint of downstream ViT models. We employ two complementary semantic annotation backends: Qwen3-VL leverages the open-domain understanding capability of a multimodal large language model to generate rich semantic descriptions, particularly effective for long-tail objects and functional characterizations; RAM++~\citep{huang2023openset} provides broader category coverage and higher recall at faster inference speed. The raw outputs of both backends are passed through a VLM-based \textit{semantic normalization module} that identifies and merges synonym groups (e.g., ``cup'' and ``mug''), removes redundant categories, and produces a normalized global label table $\mathcal{L} = \{(l_j, \text{id}_j)\}$ together with a text prompt $\mathcal{T}$ that conditions the subsequent detection and segmentation models. This normalization step is critical for reducing spurious detections in the downstream segmentation stage. As quality control at the semantic stage, we perform multi-frame consistency checks and filter out images with fewer than 2 valid detected objects (insufficient for generating spatial relation QA), as well as scenes where semantic ambiguity is detected or the robot arm itself is misidentified as a manipulable object.

\begin{figure}[t]
\centering
\includegraphics[width=\linewidth]{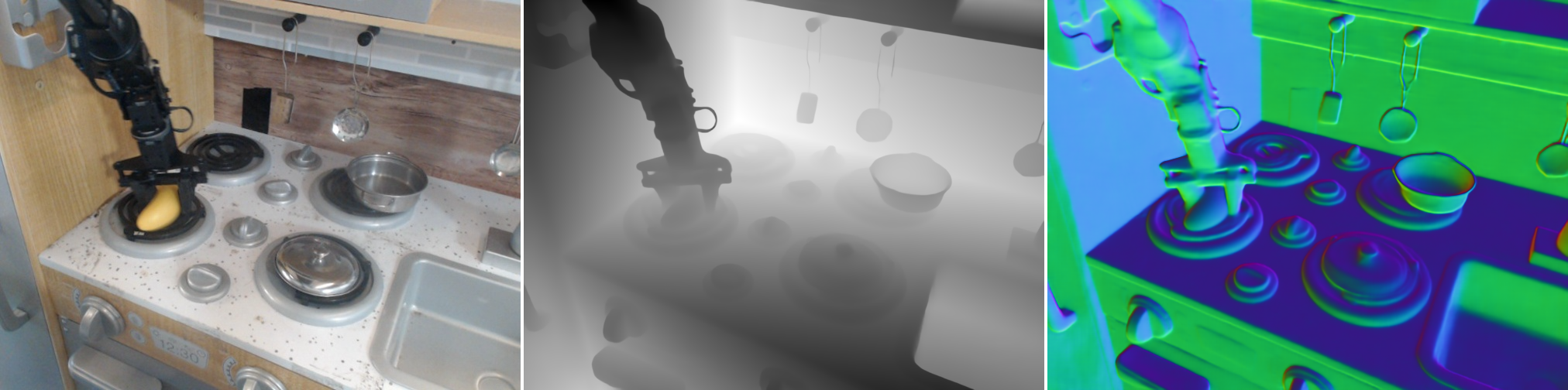}
\caption{Geometry estimation. MoGe-2 metric-scale depth and surface normal predictions on a BridgeData~V2 tabletop scene. From left to right: input RGB, predicted depth map, and predicted surface normals.}
\label{fig:spatial-vis-geo}
\end{figure}

\paragraph{Geometry estimation and 2D instance segmentation.} Building on the semantic labels, we perform geometry estimation and instance segmentation in parallel. For geometry estimation, we use the metric-scale variant of MoGe-2~\citep{wang2025moge} to jointly predict, from a single RGB image: a dense depth map $D_i \in \mathbb{R}^{H \times W}$ in absolute metric units (meters), a per-pixel surface normal map $\mathbf{N}_i \in \mathbb{R}^{H \times W \times 3}$, and camera intrinsics $\mathbf{K}_i \in \mathbb{R}^{3 \times 3}$. Since MoGe-2 directly outputs absolute-scale depth values, the subsequent 3D reconstruction and distance computation require no additional scale recovery step --- this is the key prerequisite enabling programmatic generation of distance estimation QA data. Through extensive experimentation, we found that MoGe-2 consistently outperforms Depth Anything V3 in reconstruction accuracy on tabletop-level scenes. For instance segmentation, we adopt Grounded-SAM-2~\citep{ren2024grounded} for open-vocabulary segmentation: GroundingDINO generates candidate detection boxes $\{b_k\}$ conditioned on the text prompt $\mathcal{T}$, and SAM2 refines each detection box into a precise binary mask $M_k$, yielding the complete per-image instance annotation $\{(M_k, c_k, \text{conf}_k)\}$. Compared to using a single global prompt, conditioning on per-image labels significantly reduces false positives from categories absent in the scene. At this stage, we filter out instances with detection confidence below 0.3, as well as instances whose mask area is less than 0.5\% or greater than 50\% of the image area (the former are typically noise detections; the latter are typically background misdetections such as the table surface itself). Figure~\ref{fig:spatial-vis-geo} illustrates the geometry estimation and instance segmentation outputs on two example scenes.

\paragraph{3D lifting (inverse perspective projection).} Using the depth map $D_i$, intrinsic matrix $\mathbf{K}_i$, and instance masks $\{M_k\}$, each foreground pixel $(u, v)$ belonging to instance $k$ is lifted to a 3D point in the camera coordinate frame via inverse perspective projection:
\begin{equation}
    \mathbf{p}^c = D_i(u,v) \cdot \mathbf{K}_i^{-1} [u,\; v,\; 1]^\top.
\end{equation}
To suppress depth discontinuity artifacts at object boundaries, we apply edge-aware filtering that removes pixels whose depth gradient exceeds 3 standard deviations of the local neighborhood mean, and perform voxel downsampling (voxel size 2\,mm) to control point cloud size. Additionally, we apply Statistical Outlier Removal (SOR) to each instance's 3D point cloud, ensuring reliable downstream bounding box estimation.

\begin{figure}[t]
\centering
\begin{subfigure}{0.24\linewidth}
    \includegraphics[width=\linewidth]{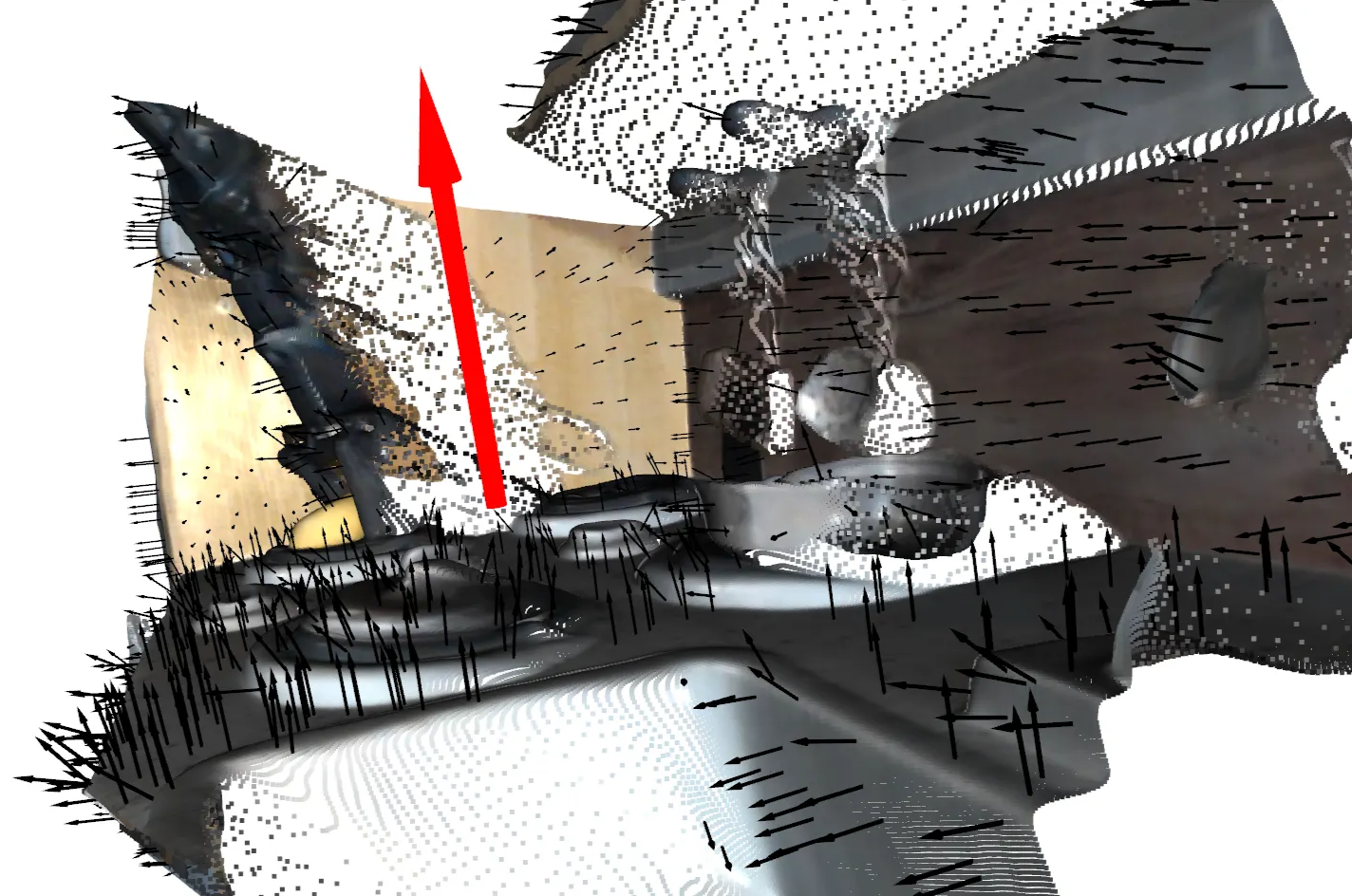}
    \caption{RANSAC normals}
\end{subfigure}
\hfill
\begin{subfigure}{0.24\linewidth}
    \includegraphics[width=\linewidth]{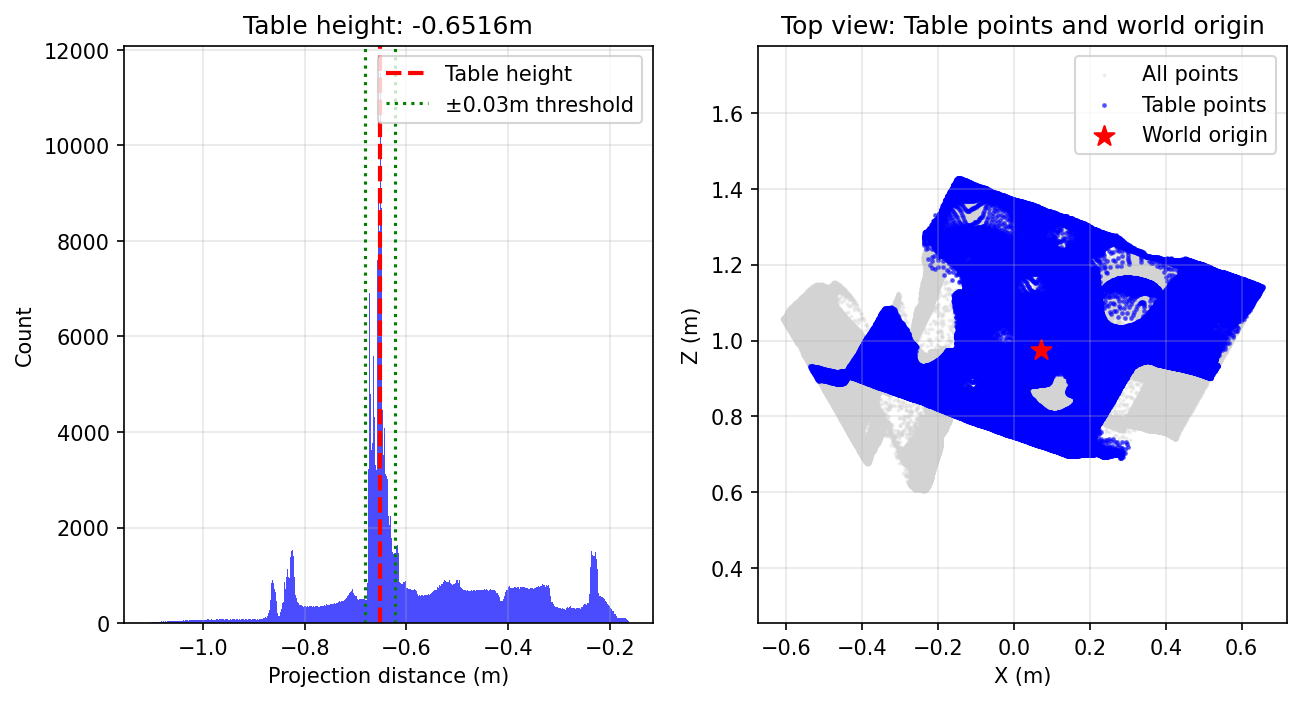}
    \caption{Height histogram}
\end{subfigure}
\hfill
\begin{subfigure}{0.24\linewidth}
    \includegraphics[width=\linewidth]{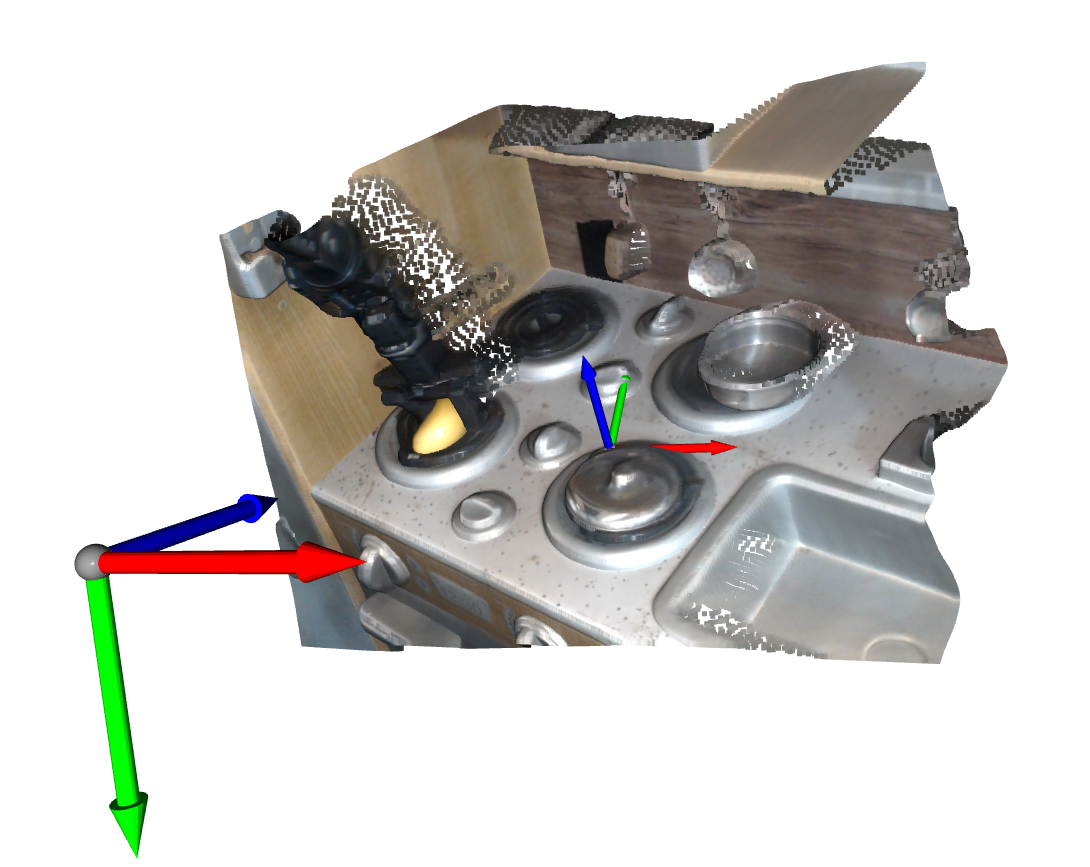}
    \caption{World frame axes}
\end{subfigure}
\hfill
\begin{subfigure}{0.24\linewidth}
    \includegraphics[width=\linewidth]{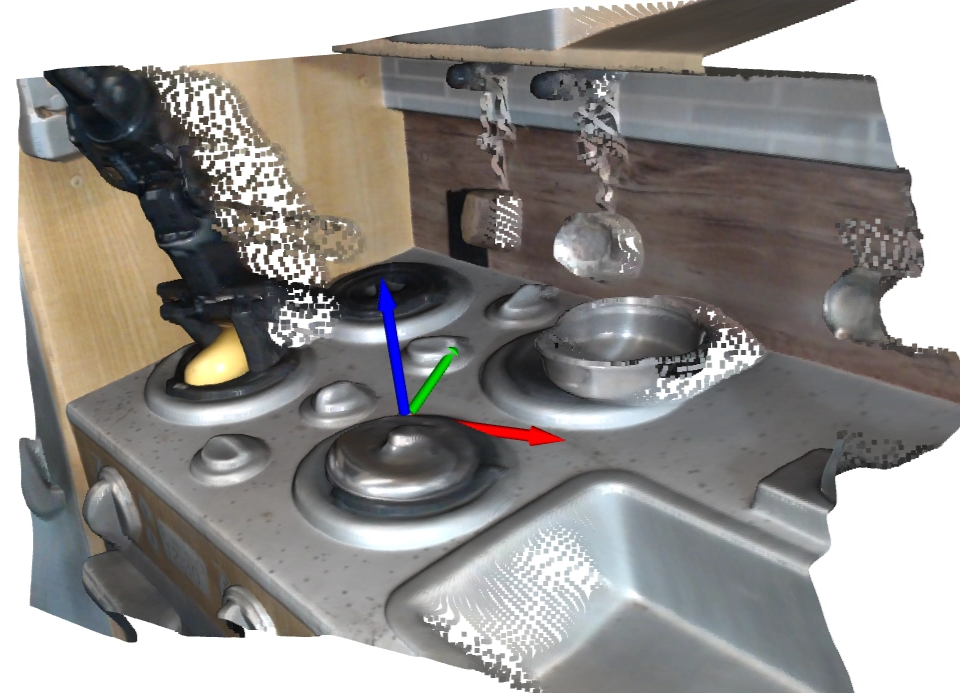}
    \caption{Aligned point cloud}
\end{subfigure}
\caption{Horizontal plane alignment. (a)~RANSAC identifies the dominant plane normal from the predicted normal map. (b)~The projection histogram determines the table surface height $d_\text{peak}$. (c)~The constructed world coordinate frame with the $z$-axis aligned to gravity. (d)~The resulting point cloud after transformation, with the table surface at $z = 0$.}
\label{fig:spatial-vis-align}
\end{figure}

\paragraph{Horizontal plane alignment.} In tabletop manipulation scenarios, the dominant plane is the table surface. We transform the camera coordinate frame to a world coordinate frame in which the table surface lies at $z = 0$ and the $z$-axis points vertically upward. First, we estimate the dominant plane normal $\hat{\mathbf{n}}_\text{best}$ via RANSAC on the per-pixel normal map $\mathbf{N}_i$: a seed normal $\vec{n}_\text{seed}$ is randomly sampled, and all normals with dot product $S_j = \vec{n}_\text{seed} \cdot \vec{n}_j > 0.996$ (angular deviation $< 5^\circ$) are counted as inliers; this is repeated for $K = 100$ iterations, selecting the estimate with the largest inlier count. As a quality check, if the best inlier ratio is below 30\%, we deem the scene to lack a clear dominant plane (e.g., a cluttered non-tabletop scene) and discard it. Second, we determine the table height via a projection histogram: all camera-frame points are projected onto $\hat{\mathbf{n}}_\text{best}$ to obtain $d_j = \hat{\mathbf{n}}_\text{best} \cdot \mathbf{p}_j^c$, and the histogram peak $d_\text{peak}$ identifies the table surface height (Figure~\ref{fig:spatial-vis-align}a--b). Third, we construct the $4 \times 4$ camera-to-world transformation: setting $\mathbf{z}_\text{new} = \hat{\mathbf{n}}_\text{best}$, we build the remaining axes via Gram-Schmidt orthogonalization to obtain the rotation matrix $\mathbf{R} = [\mathbf{x}_\text{new},\; \mathbf{y}_\text{new},\; \mathbf{z}_\text{new}]^\top$, set the translation $\mathbf{t} = [0,\; 0,\; -d_\text{peak}]^\top$, and form:
\begin{equation}
    \mathbf{T}_{c \to w} = \begin{bmatrix} \mathbf{R} & \mathbf{t} \\ \mathbf{0}^\top & 1 \end{bmatrix}.
\end{equation}
The transformation is applied to all point clouds ($\mathbf{P}_i^w = \mathbf{T}_{c \to w} \cdot \mathbf{P}_i^c$) and camera poses, ensuring that the $z$-axis of all processed images is consistently aligned with the scene's gravity direction (Figure~\ref{fig:spatial-vis-align}c--d). After transformation, we further verify that the $z$-coordinates of most object bounding box bottoms lie within 7\,cm of $z = 0$; scenes that fail this sanity check are flagged as low-quality and removed.

\begin{figure}[t]
\centering
\includegraphics[width=\linewidth]{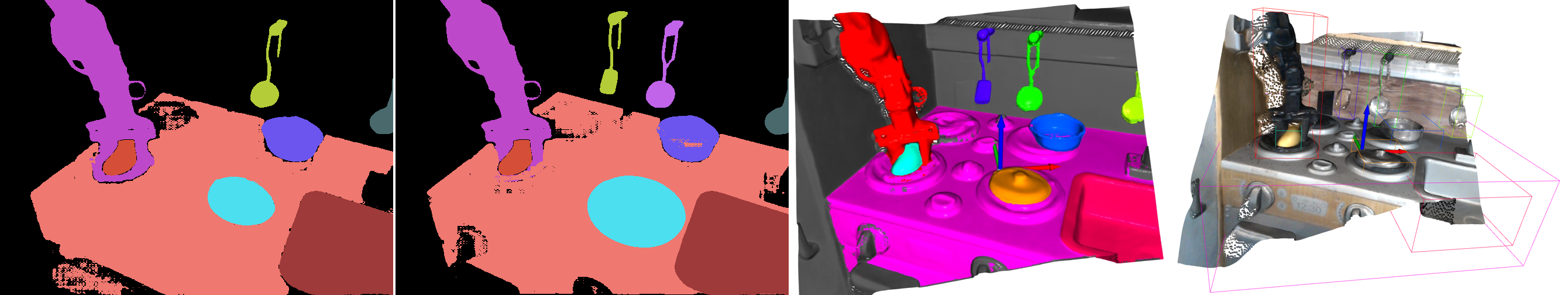}
\caption{Instance segmentation and 3D scene reconstruction. From left to right: raw instance segmentation masks, fused instance masks, per-instance 3D point clouds with semantic coloring, and fitted 3D bounding boxes in the gravity-aligned world frame.}
\label{fig:spatial-vis-seg}
\end{figure}

\paragraph{Spatial reasoning QA generation.} All intermediate outputs are aggregated into a unified per-scene metadata record $\mathcal{S}_i$ containing: image, normalized semantic label table, per-instance 2D masks and 3D point cloud segments, estimated intrinsics, per-instance 3D bounding boxes, and the world-coordinate transformation matrix. Leveraging this dense spatial annotation, we programmatically generate QA pairs covering key task types including spatial relations, distance metrics, scene cognition, and appearance order, constituting the tabletop-level spatial reasoning QA dataset ER1.5-Spatial ($\sim$20K samples) that directly feeds into the training data system described in Section~\ref{sec:data}.

\subsection{Pipeline 2: Failure-Aware Data Construction for Correction Data}
\label{app:correction-pipeline}

This section details the construction of the ER1.5-Correction dataset ($\sim$800K+ samples), a large-scale failure-aware QA dataset designed to endow the model with failure perception and autonomous correction capabilities. Motivated by the failure taxonomy frameworks of RoboFAC~\citep{ye2025robofac} and Guardian~\citep{pacaud2025guardian}, the dataset covers both simulation and real-world scenarios, organizing failures along two orthogonal dimensions.

\paragraph{Failure taxonomy.} We categorize failures by \textit{stage} and \textit{cognitive level}. By stage: (1)~\textit{Planning failures} arise from inaccurate task decomposition, including step omission, step redundancy, step swap, object error, and action/location error. (2)~\textit{Execution failures} arise from physical execution imprecision, including execution interruption, wrong manipulation object, wrong action, and operation failure. By cognitive level, we construct QA at three progressive tiers: failure detection (binary yes/no), failure localization (identifying the failure type and which subtask), and failure correction (generating a recovery plan). The combination yields six QA task types.

\paragraph{Data sources.} We collect raw data from multiple complementary sources. For simulation: RoboFail~\citep{liu2023reflect} provides failure execution videos across 10 tasks with structured annotations; ManiSkill~\citep{tao2024maniskill3} contributes $\sim$3.5K samples via 26 perturbation schemes across 11 tasks; GEMBench~\citep{garcia2024gembench} covers 31 tasks with keyframe images and subtask lists. For real-world: RoboFail additionally provides 30 failure episodes across 10 real-world tasks; BridgeData~V2~\citep{walke2023bridgedata} contains over 60K episodes with complete image sequences, language instructions, and step-by-step reasoning annotations; RoboFAC~\citep{ye2025robofac} provides failure videos across 6 tasks with multi-level QA annotations.

\paragraph{Planning failure construction.} The core idea is applying structured perturbations to correct subtask plans. Given a correct plan $\mathcal{P} = \{s_1, \dots, s_n\}$, we apply five perturbation operators: step deletion ($\mathcal{D}_\text{del}$), step duplication ($\mathcal{D}_\text{dup}$), step swap ($\mathcal{D}_\text{swap}$), object replacement ($\mathcal{D}_\text{obj}$, substituting the manipulation target with another scene object), and action/location replacement ($\mathcal{D}_\text{act}$, replacing the action verb with a semantically similar but executionally inequivalent alternative). Each perturbation simultaneously generates all three cognitive levels of QA. We retain original correct plans as positive samples to prevent overfitting. For BridgeData, we leverage its step-by-step reasoning annotations for fine-grained perturbation; for ManiSkill and GEMBench, perturbations are constructed from task structure and keyframe information.

\paragraph{Execution failure construction.} We exploit inconsistencies between video content and instructions via three strategies: (1)~\textit{Truncation}: successful videos are truncated at subtask boundaries to simulate execution interruption. (2)~\textit{Description replacement}: the manipulation object or action verb in subtask descriptions is modified to create mismatches with actual video content. (3)~\textit{Perturbation injection}: in ManiSkill, physical perturbations are injected during task execution (e.g., opposing forces during grasping) to produce realistic failure videos. For RoboFail and RoboFAC, which natively contain real failure records, we directly extract and reformat into unified QA.

\paragraph{Quality control and statistics.} All QA templates undergo human sampling review. We verify that perturbed plans genuinely cannot complete the task, ensure truncation points fall at subtask boundaries, confirm that replaced objects/actions are semantically plausible but executionally incorrect, and enforce positive/negative sample balancing for detection QA. The final dataset comprises $\sim$800K+ samples: BridgeData contributes $\sim$802K (19 QA subtypes), ManiSkill $\sim$3.5K (11 tasks), with additional contributions from RoboFail, RoboFAC, and GEMBench. The three cognitive tiers are approximately balanced.

\subsection{Pipeline 3: Functional Affordance \& Trajectory Data Construction}
\label{app:affordance-trajectory-pipeline}

This section describes the two automated pipelines that generate functional affordance data (for OFG) and trajectory data (for VTG).

\subsubsection{Object Functional Affordance Data}

Object Functional Grounding (OFG) requires the model to localize functional parts of objects (handles, spouts, buttons, etc.) and understand their action affordances. Part-level annotation in the real world is extremely expensive, creating a bottleneck for scaling OFG data. We construct OFG data automatically from two complementary sources:

\textbf{Simulation-based extraction.} We leverage the advantage that part-level semantic annotations in simulation environments are essentially free. We render diverse articulated objects from multiple viewpoints and lighting conditions, and automatically generate functional grounding annotations by mapping semantic part labels to image coordinates.

\textbf{Human interaction data extraction.} We extract functional part grounding from large-scale hand-object interaction datasets, where human grasping patterns naturally indicate functional affordance locations. By tracking contact regions between human hands and objects, we obtain supervision signals for where and how objects should be grasped for different manipulation intents.

\subsubsection{Trajectory Data}

Visual Trace Generation (VTG) generates ordered point sequences that describe complete manipulation trajectories, supporting both 2D and 3D trace formats. While many robot demonstration datasets contain action sequences, they lack the explicit 2D/3D visual trajectory annotations required for VTG training. We design automated trajectory extraction pipelines to bridge this gap, constructing data from two complementary trajectory types:

\textbf{Robot-centric traces (end-effector flow).} These traces describe the expected motion path of the robot end-effector. For robot demonstration datasets that record end-effector poses, we project the 3D end-effector positions onto the 2D image plane using camera intrinsics and extrinsics, producing pixel-level end-effector motion traces. For 3D traces, we directly extract end-effector position sequences in world coordinates and normalize them relative to the workspace. The resulting traces encode the spatial trajectory the robot arm should follow to complete a manipulation task, providing direct motion planning supervision.

\textbf{Object-centric traces (object flow).} These traces describe the expected motion path of the target object from its starting position to the goal position. For hand-object interaction datasets, we track the motion of manipulated objects across frames using object tracking and segmentation, generating object-centric motion flow trajectories. For simulation data, we directly read object pose sequences from the physics engine. Object flow traces are particularly important for tasks like pushing, pouring, or sweeping, where the desired outcome is defined by the object's motion rather than the end-effector's path.

Both trace types undergo quality filtering (removing trajectories with excessive noise, insufficient motion, or physically implausible paths), coordinate normalization, and are paired with natural-language instructions describing the intended action. Data sources span large-scale robot demonstration datasets (e.g., DROID, GenManip~\citep{gao2025genmanip}, InternData-A1~\citep{tian2026interndata}) and human interaction video datasets, collectively providing diverse motion patterns across manipulation primitives.

\section{Qualitative Visualizations}
\label{app:visualizations}

\begin{figure*}[t]
\centering
\includegraphics[width=\textwidth]{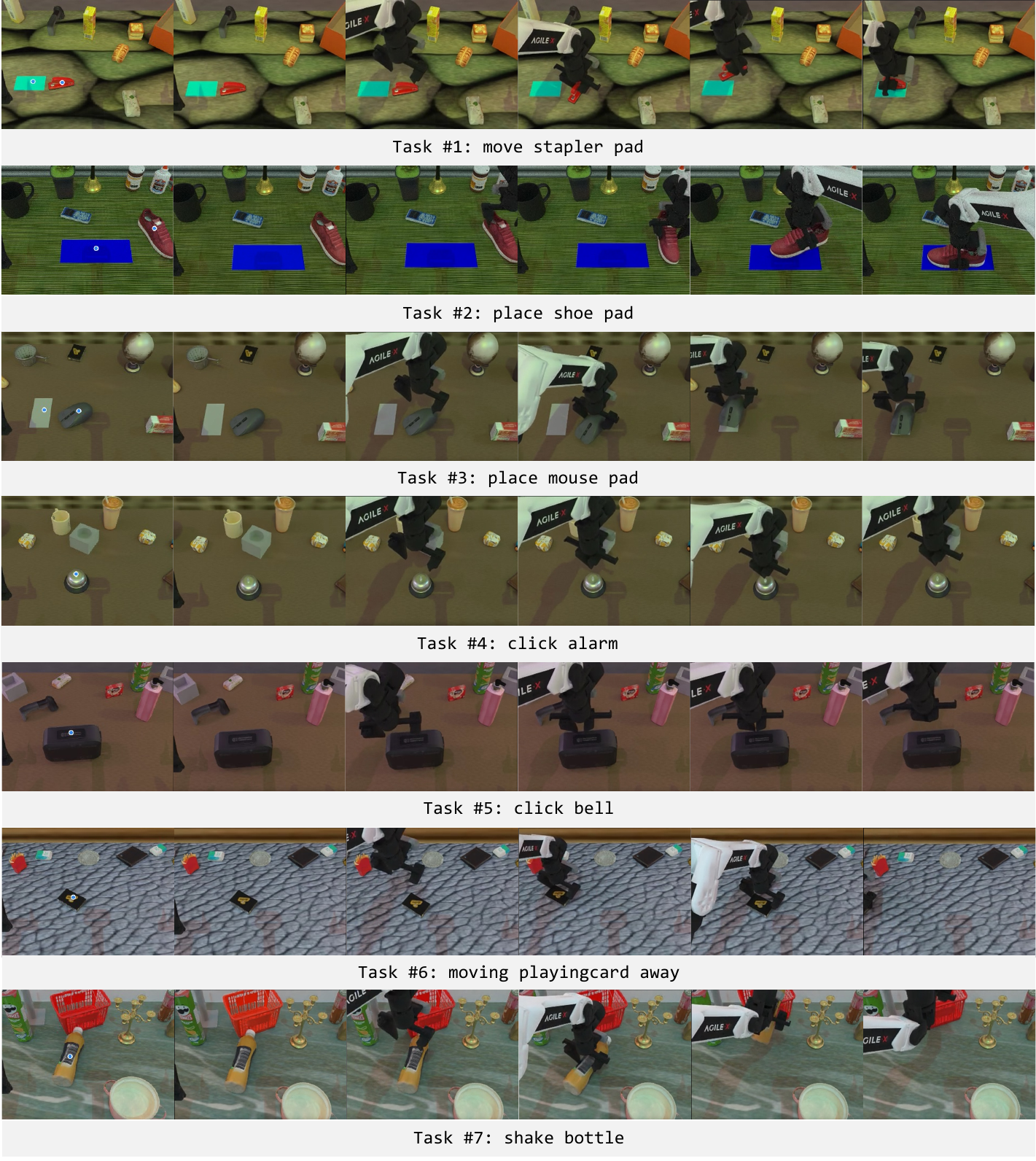}
\caption{\textbf{RoboTwin zero-shot manipulation visualization.} Embodied-R1.5 predicts accurate pointing locations (shown as colored dots) for each task, which are then converted to robot actions via a unified motion logic. The model successfully identifies grasp points, place targets, and functional affordances across diverse manipulation tasks without any RoboTwin training data.}
\label{fig:robotwin-vis}
\end{figure*}

\begin{figure*}[t]
\centering
\begin{minipage}{1\textwidth}
    \centering
    \includegraphics[width=\textwidth]{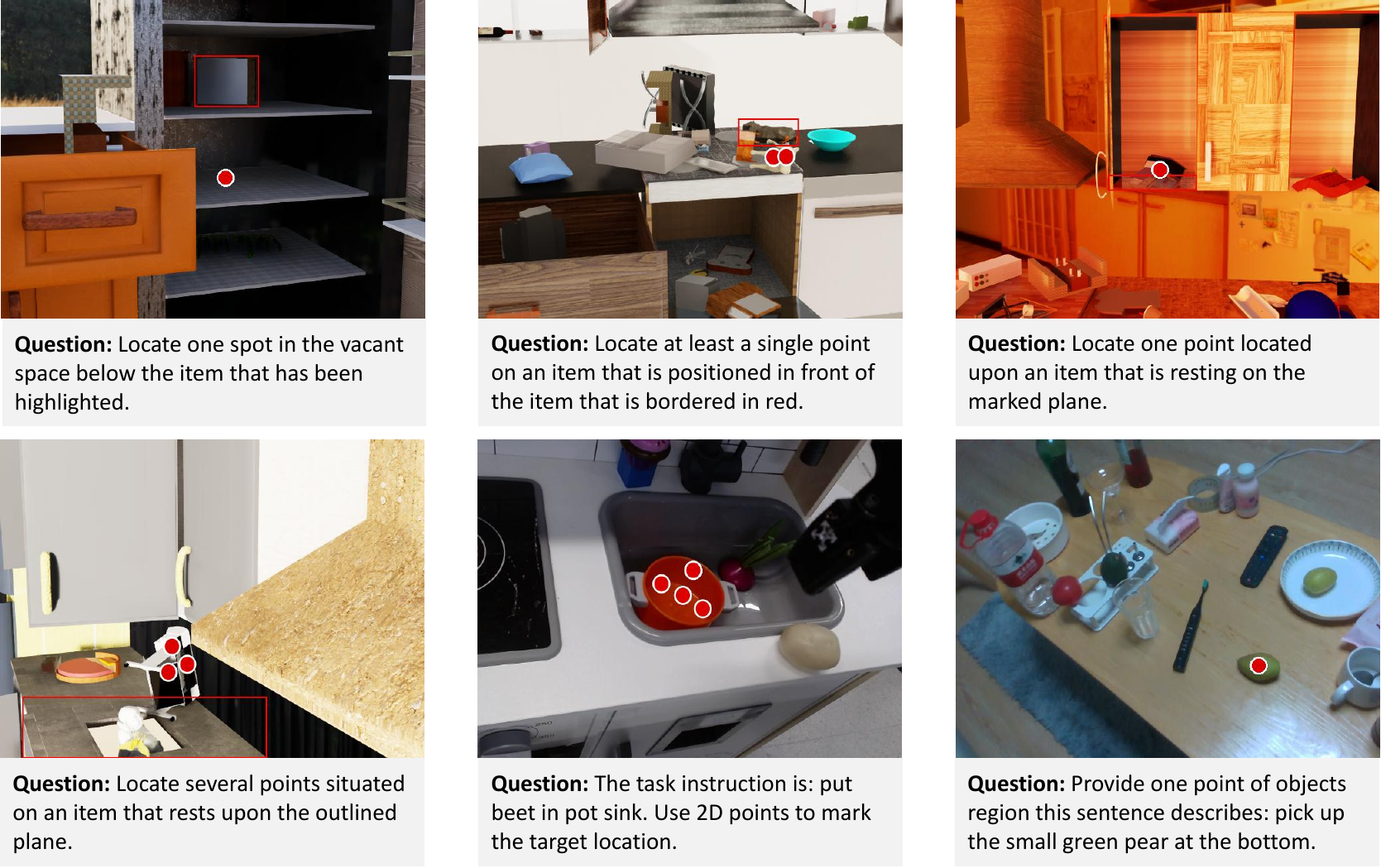}
    \includegraphics[width=\textwidth]{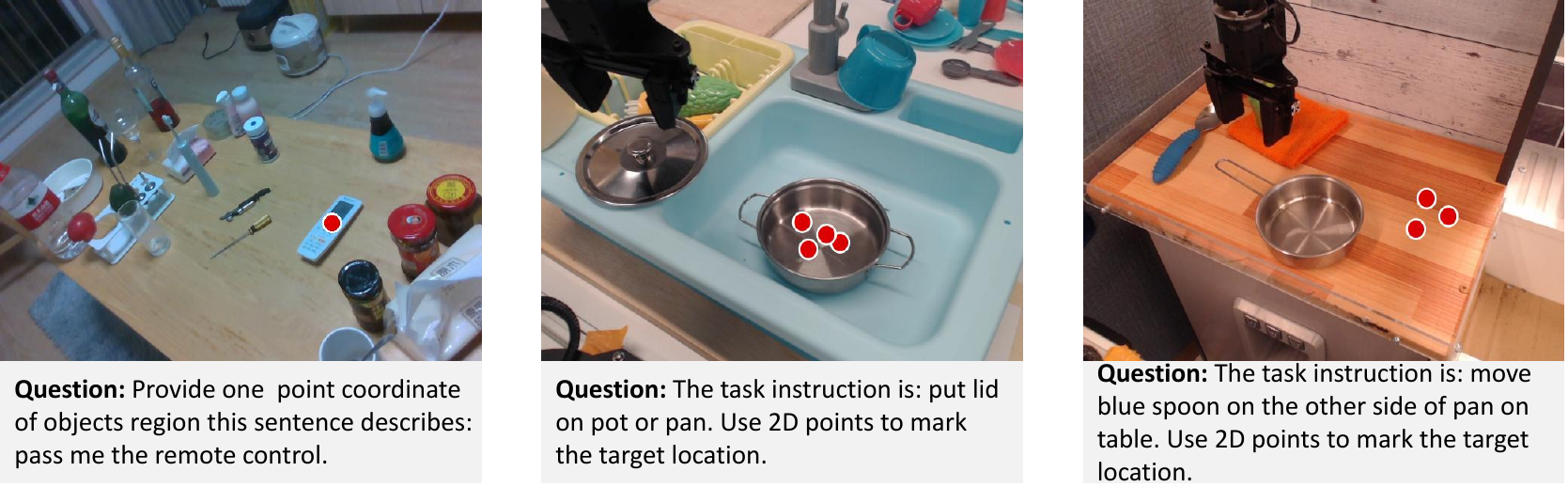}
\end{minipage}
\caption{\textbf{Pointing examples of Embodied-R1.5.} Qualitative visualizations covering referring expression grounding, region-level free-space identification, and functional part localization across diverse manipulation scenes.}
\label{fig:point-examples-vis}
\end{figure*}

\begin{figure*}[t]
\centering
\begin{minipage}{1\textwidth}
    \centering
    \includegraphics[width=\textwidth]{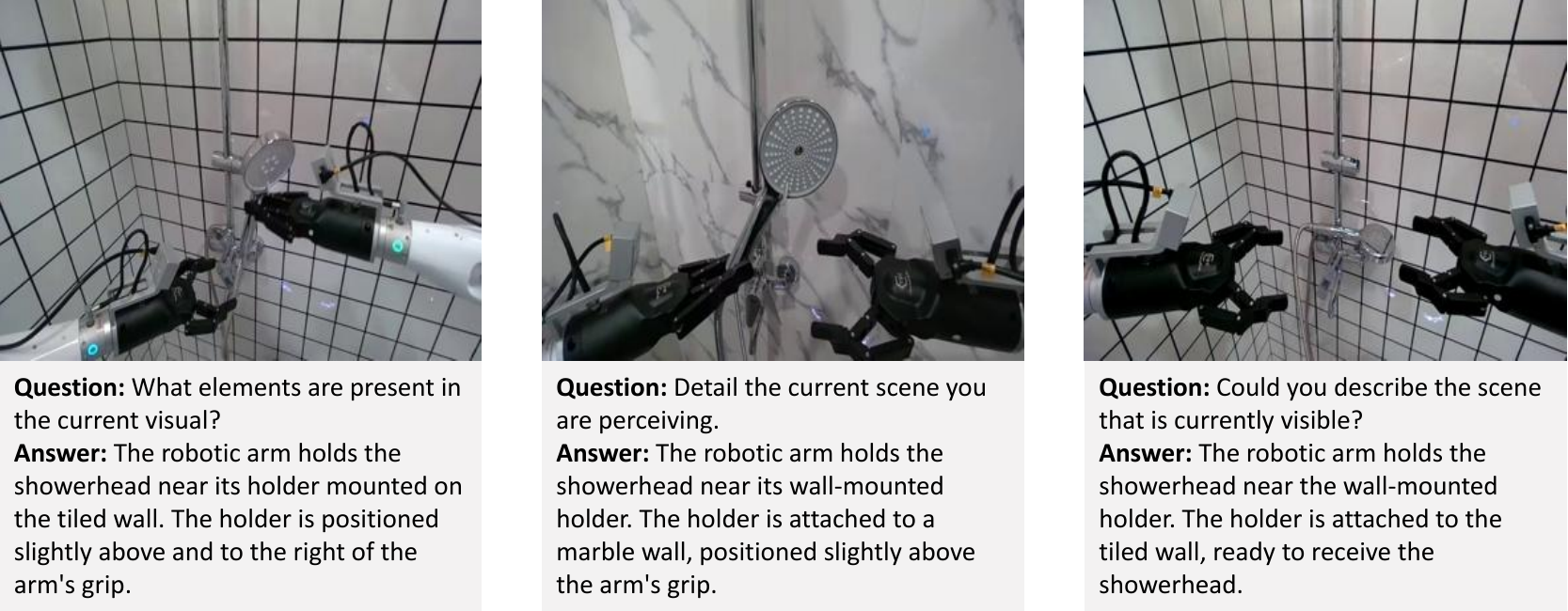}
    \includegraphics[width=\textwidth]{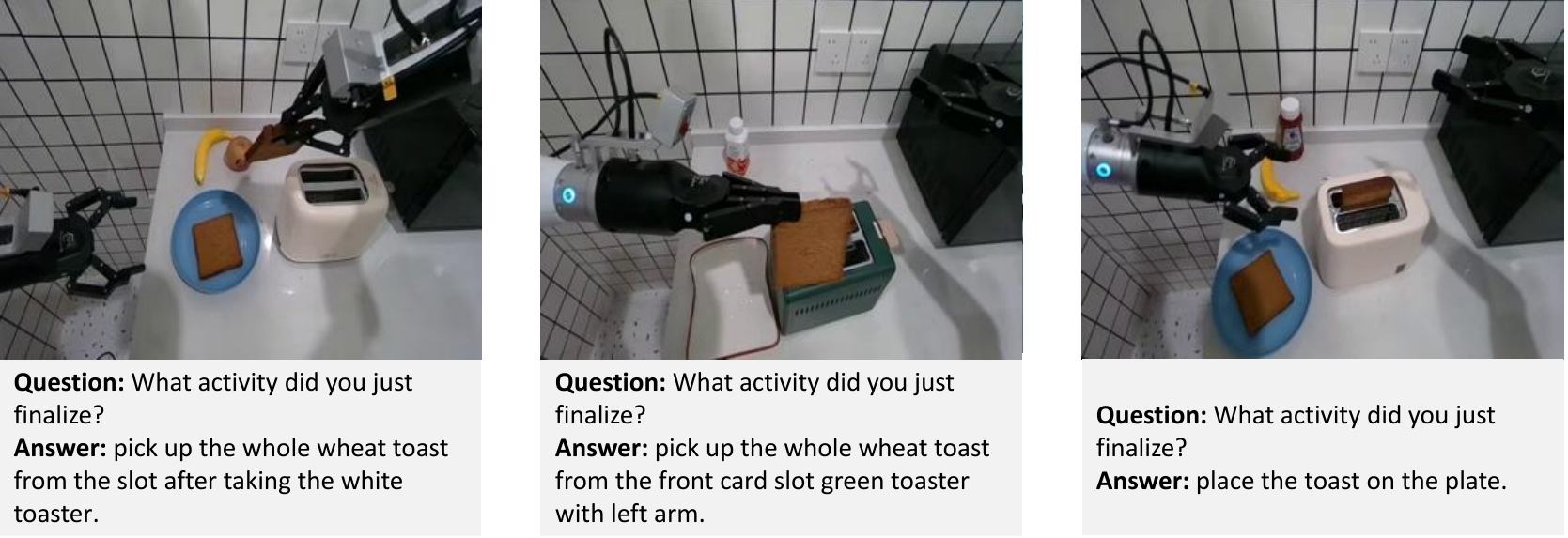}

    \includegraphics[width=\textwidth]{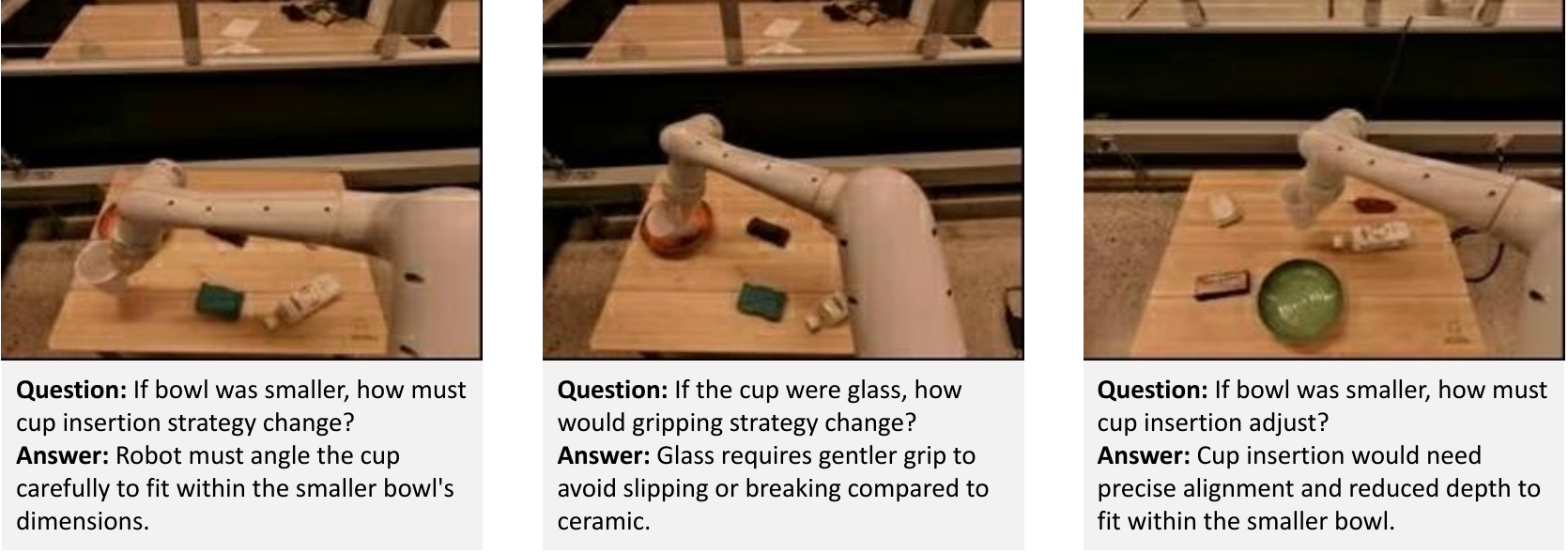}
\end{minipage}
\caption{\textbf{Spatial reasoning examples of Embodied-R1.5.} Qualitative visualizations of multi-view spatiotemporal reasoning, depth and metric understanding, and object-level spatial relation reasoning under diverse embodied scenes.}
\label{fig:spatial-examples-vis}
\end{figure*}

\begin{figure*}[t]
\centering
\begin{minipage}{1\textwidth}
    \centering
    \includegraphics[width=\textwidth]{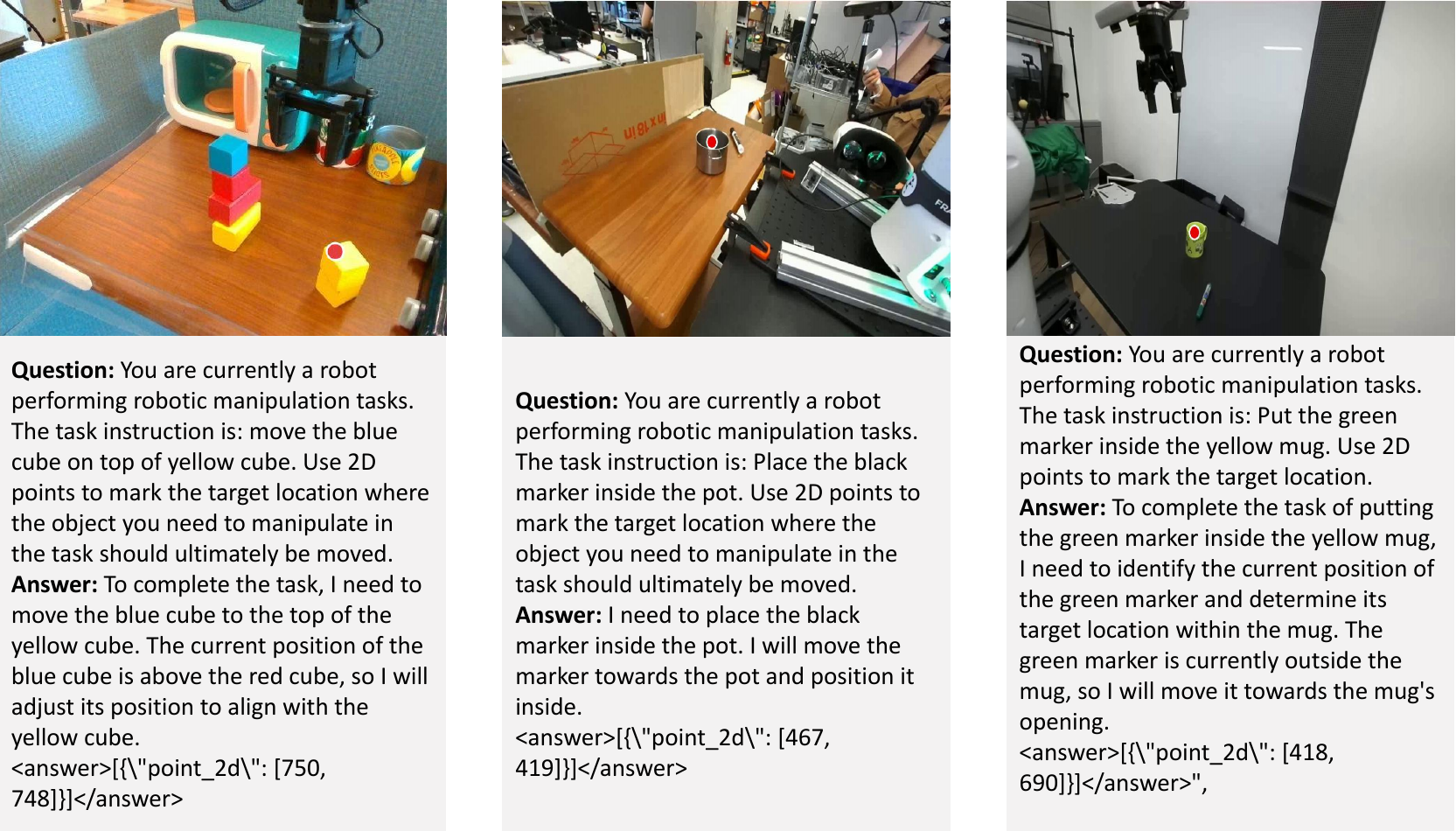}
    \includegraphics[width=\textwidth]{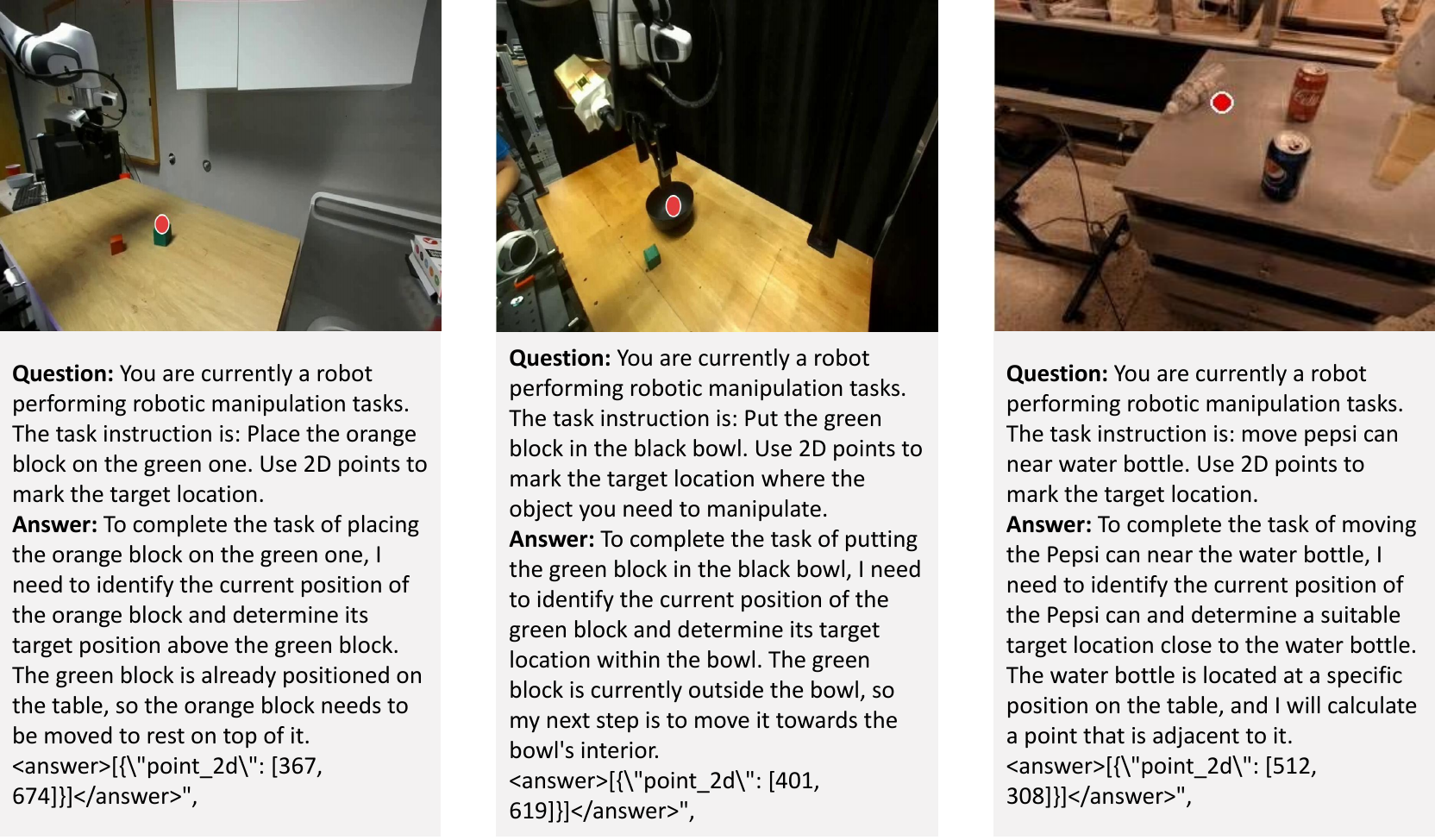}
    \includegraphics[width=\textwidth]{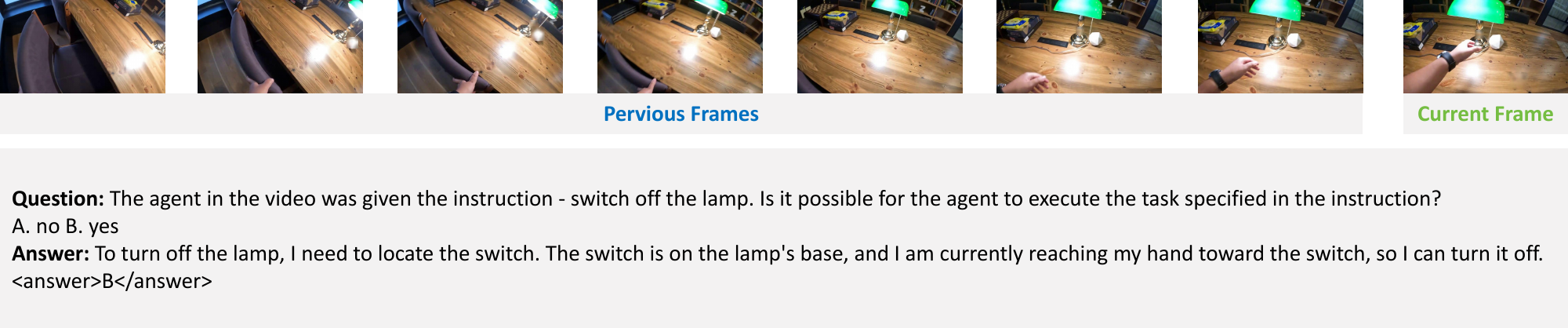}
\end{minipage}
\caption{\textbf{Chain-of-thought reasoning examples of Embodied-R1.5 (1/2).} Pointing over single-image manipulation scenes and feasibility planning over egocentric video, with the final decision emitted within an \texttt{<answer>} tag.}
\label{fig:thinking-examples-vis}
\end{figure*}

\begin{figure*}[t]
\centering
\begin{minipage}{1\textwidth}
    \centering
    \includegraphics[width=\textwidth]{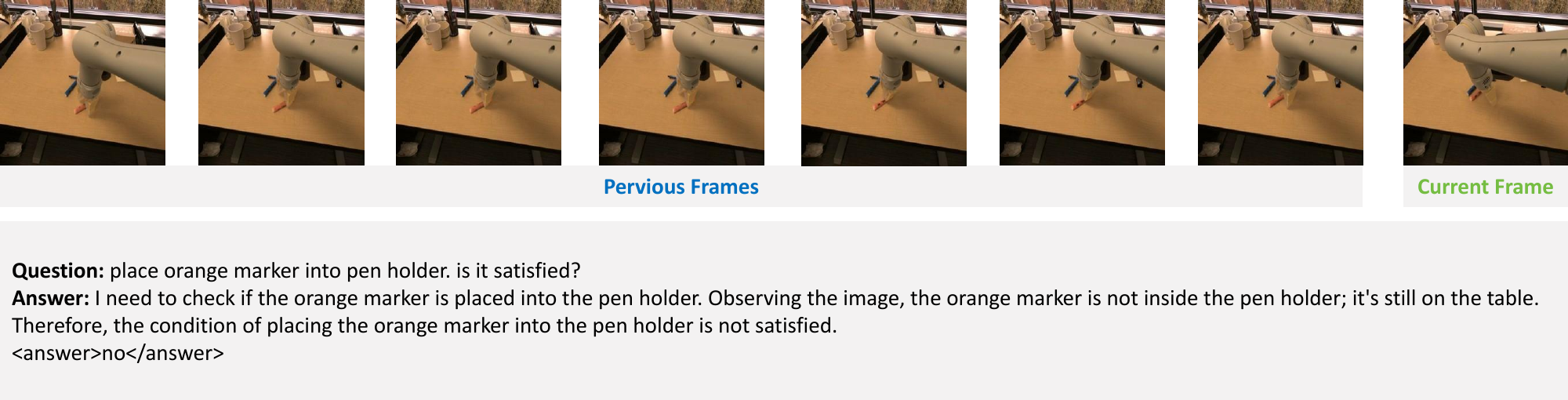}
    \includegraphics[width=\textwidth]{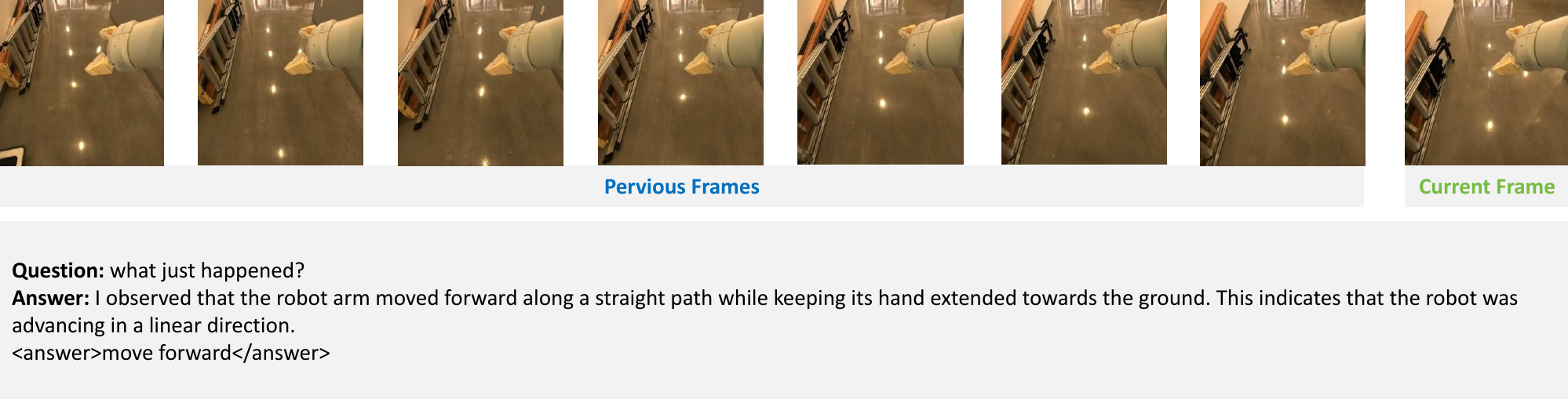}
\end{minipage}
\caption{\textbf{Chain-of-thought reasoning examples of Embodied-R1.5 (2/2).} State verification and correction and action understanding over observation frame sequences, with the final decision emitted within an \texttt{<answer>} tag.}
\label{fig:thinking-examples-vis-2}
\end{figure*}

This appendix provides qualitative visualizations of Embodied-R1.5 across four complementary aspects: zero-shot pointing on RoboTwin manipulation tasks (\cref{fig:robotwin-vis}), embodied pointing across diverse scenes (\cref{fig:point-examples-vis}), embodied spatial reasoning (\cref{fig:spatial-examples-vis}), and explicit chain-of-thought reasoning across pointing, planning, correction, and action understanding (\cref{fig:thinking-examples-vis,fig:thinking-examples-vis-2}). Together, these visualizations illustrate how a unified embodied foundation model handles a wide spectrum of grounding and reasoning queries within a single architecture.

\subsection{RoboTwin Zero-Shot Manipulation}
\label{app:robotwin-vis}

\subsection{Pointing Examples}
\label{app:point-examples}

\subsection{Spatial Reasoning Examples}
\label{app:spatial-examples}

\subsection{Chain-of-Thought Reasoning Examples}
\label{app:thinking-examples}

\end{document}